\documentclass[onecolumn,10pt]{article}         
\usepackage[top=3.6cm, bottom=3.2cm, left=2.5cm, right=2.5cm]{geometry}

\usepackage{graphicx}
\usepackage{booktabs}
\usepackage{subfigure}
\usepackage{tikz}
\usepackage{amsmath,amssymb} 
\usepackage{bm}
\usepackage{todonotes}
\usepackage{multirow}
\usepackage{diagbox}
\usepackage{xcolor}
\usepackage{algorithm}
\usepackage{algpseudocode}
\usepackage{array}
\usepackage{color}
\usepackage{graphicx}
\usepackage{setspace}
\usepackage{lipsum}
\newtheorem{definition}{Definition}

\newcommand{\NFA}{{\rm NFA}}

\usepackage[nocompress]{cite}
\usepackage{url}

\makeatletter
\def\url@leostyle{%
	\@ifundefined{selectfont}{\def\UrlFont{\sf}}{\def\UrlFont{\small\ttfamily}}}
\makeatother
\begin{document}
\title{Anisotropic-Scale Junction Detection and \\ Matching for Indoor Images}
\author{Nan~Xue$^1$, Gui-Song~Xia$^{1}$, Xiang~Bai$^{2}$, Liangpei~Zhang$^{1}$, Weiming~Shen$^1$, 
	\\
	$^1${\em State Key Lab. LIESMARS, Wuhan University, Wuhan, China.}\\
	$^2${\em Electronic Information School, Huazhong University of Science and Technology, China.}
}

\maketitle

\begin{abstract}
	Junctions play an important role in characterizing local geometrical structures of images, and the detection of which is a longstanding but challenging task.
	Existing junction detectors usually focus on identifying the location and orientations of junction branches while ignoring their scales which however contain rich geometries of images.
	This paper presents a novel approach for junction detection and characterization, which especially exploits the locally anisotropic geometries of a junction and estimates its scales by relying on an \emph{a-contrario} model. The output junctions are with anisotropic scales, saying that a scale parameter is associated with each branch of a  junction, and are thus named as {\em anisotropic-scale junctions} (ASJs). We then apply the new detected ASJs for matching indoor images, where there are dramatic changes of viewpoints and the detected local visual features, e.g. key-points, are usually insufficient and lack distinctive ability.
	We propose to use the anisotropic geometries of our junctions to improve the matching precision of indoor images.
	The matching results on sets of indoor images demonstrate that our approach achieves the state-of-the-art performance on indoor image matching.
\end{abstract}

\vspace{2cm}
\section{Introduction}\label{sec:introduction}

%
%
%
%

\begin{figure}
	\centering
	\includegraphics[width=0.4\textwidth]{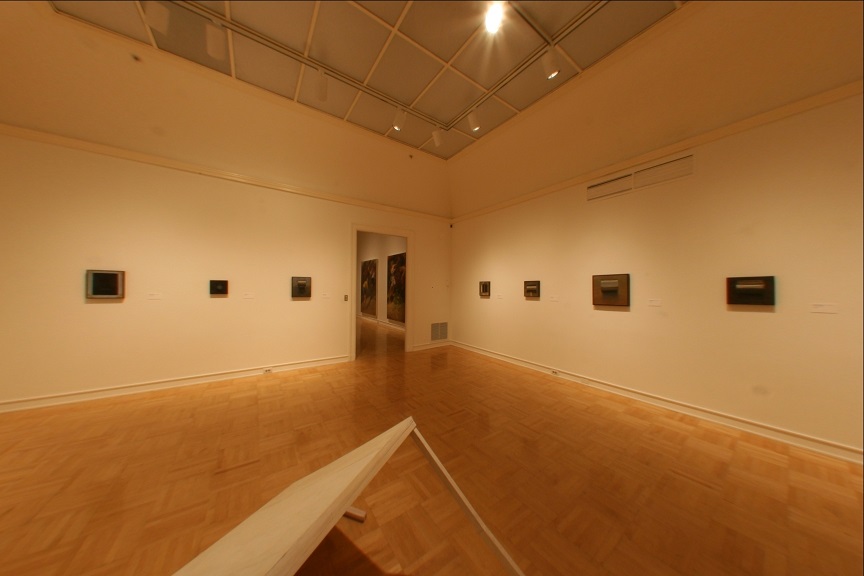}
	\includegraphics[width=0.4\textwidth]{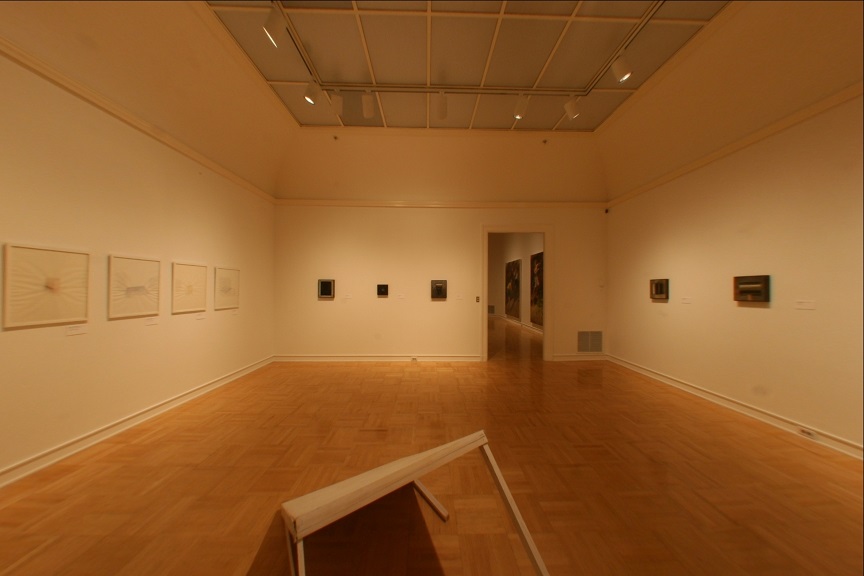}
	\caption{A pair of indoor images. It can be seen that these images are dominated by
		geometrical structures. e.g. the edges of the door, and low-textured wall.}
	\label{fig:example-indoor-images}
\end{figure}

Image correspondence is a key problem for many computer vision tasks, such as structure-from-motion~\cite{Wu13,CrandallOSH13,FuhrmannLG14,openMVG}, object recognition~\cite{WangBWLT10,ChiaRLR12} and many others~\cite{YanWZYC15,ShenLYXWW15}.
The past decades have witnessed the big successes on that problem achieved by detecting and matching local visual features~\cite{MikolajczykS02,Lowe04,MatasCUP02,TuytelaarsG04,YuM11}. 
Although most of existing image matching algorithms relying on such local visual features perform well for images containing rich photometric information, e.g. outdoor images, they usually lose their efficiency on images that are less photometric and dominated by geometrical structures such as indoor images displayed in Fig.~\ref{fig:example-indoor-images}. 
In the indoor scenario, images are often dominated by low-texture parts and are with severe viewpoint changes, in which case it is reported to be more effective to make the correspondence of geometrical structures~\cite{FanWH12B,LiYLLZ16} such as line segments~\cite{GioiJMR10,GioiJMR12} and junctions~\cite{ShenP00}.

The line segment matching problem has been studied in recent years since it can represent more structural information than key-points.
Many algorithms match line segments by using either photometric descriptors with individual line segments~\cite{WangWH09} or the initial geometric relation~\cite{FanWH12B,LiYLLZ16} to assist line segment matching. 
The approaches using pre-estimated epipolar geometry usually perform better than those of using photometric descriptors~\cite{WangWH09}, but the epipolar geometry estimation still needs key-point correspondences in many situations. In the indoor scenes, due to the fact that descriptors for low-textured regions are not distinctive enough, it is very likely to produce unstable epipolar geometry for inferring the line segment matching.
It is thus of great interest to develop elegant ways to make the correspondences of geometrical structures of images while get rid of the errors raised by the key-point correspondences, for finally achieving better matching of indoor images.

Alternatively, as a kind of basic structural visual features, junctions, have been studied as the primary importance for perception and scene understanding in recent years~\cite{Marr82,Adelson00,GuoZW07}.
Being a combination of points and ray segments, junctions contain richer information than line segments, {\em i.e.} including a location and at least two ray segments (known as {\em branches}). 
Ideally, the information contained by a pair of junctions enables us to recover the correspondences between images up to affine transformations.
However, due to the difficulties in the estimation of the endpoints of junction branches, most of junction detection algorithms~\cite{WuXZ07,MaireAFM08,Sinzinger08,PuspokiU15,PuspokiUVU16,XiaDG14} concentrate on identifying the locations and orientation of branches while ignoring their length. 
This actually simplifies junctions as key-points and does not fully exploit their capabilities for image correspondences. 
To characterize the structure of junctions better, the detector ACJ \cite{XiaDG14} estimates scale invariant junction and it can be represented isotropically as a circle region with two or more dominant orientations. Every  orientation represents a branch of junction and the radius of circle is equal to the shortest length among these branches.
Although the orientation of branches is invariant with respect to viewpoint, it is not enough for estimating the affine transformation. Fortunately, if we can estimate the length of every branch, the affine transformation will be determined by a pair of junction correspondence.
 
Motivated by this, we are going to study for exploiting the invariance of junctions through estimating scale (length) of branches.
For indoor images, the inherent scale for junctions usually are the length of some (straightforward) boundary for salient objects in images, which contains rich structure information and beyond local features. More precisely, we proposed an \emph{a-contrario} approach that models the endpoints of a ray segment starting at given location with initial orientations, which check the proposed point if it should be a part of the ray segment according to \emph{number-of-false-alarms} (NFA). When the points that belong to the ray segment occurs continuously until the continuity broken, the inherent scale for the ray segment is determined. In reality, the initial orientations are noised, we also optimize them with the junction-ness based on the \emph{a-contrario} theory. 
Once the anisotropic scale is estimated for each branch (ray segment), the local homography can be estimated from any pair of junctions extracting from two different images. Theoretically, the correct correspondence produce reasonable local affine homography while incorrect correspondences generate local homographies in their own way. Considering the certainty of junction locations, the regions around location can be mapped by correct or incorrect affine homography. Correct homographies will map one image to another with minimal patch distortion. Comparing the regions with induced affine homography for a pair of junctions can check if the pair are correspondence. When the corresponding junctions are identified in image pairs, the results will produce more structure information. Our contributions in this paper are
\begin{itemize}
	\item We extent the junction detector in~\cite{XiaDG14} to anisotropic-scale geometrical structures, which can better depict the geometric aspect of indoor images.
	\item We developed an efficient scheme for making the correspondence of \emph{anisotropic-scale junctions}. More precisely, as a detected \emph{anisotropic-scale junction} provides at least three points, each pair of junctions in images can induce an affine homography. We finally present a strategy by  induced homographies to generate accurate and reliable correspondences for the location and anisotropic branches of junctions simultaneously.
	\item We evaluate our method on challenging indoor image pairs, e.g. some of images are from the indoor image datasets 
	used in \cite{SrajerSPP14,FurukawaCSS09} and our results demonstrate that it can achieve state-of-the-art
	performance on matching indoor images.
\end{itemize}

The rest of this paper is organized as follows. First, the existing research related to our work is given in Sec.~\ref{sec:related-work}. In Sec.~\ref{sec:problems}, the problem of detecting and matching junctions for indoor scene is discussed. Next to this section, an \emph{a-contrario} approach for detecting \emph{anisotropic-scale junction} is described. As for the junction matching, we design a dissimilarity in Sec.~\ref{sec:matching} to find the correspondences. The experimental results and analysis for our approach are given in Sec~.\ref{sec:exp}. Finally, we conclude our paper in Sec.~\ref{sec:con}.
\section{Related works}
\label{sec:related-work}
In this section, we briefly review the existing approaches for junction detection and matching as well as geometrical structure matching for indoor images. 
\subsection{Junction detection}
Detecting junction structure in images has been studied for years\cite{Forstner86,HarrisS88,MikolajczykS04,MaireAFM08,ForstnerDS09,PuspokiU15,PuspokiUVU16}.
In the early stage, junction was studied as corner points~\cite{Forstner86,HarrisS88}. 
For the sake of recognition, the scale of junctions or other key-points also have been studied~\cite{Lowe04,MikolajczykS04,XiaDG14}. These approaches estimate the scale around junction locations by using scale space theories~\cite{AlvarezM97,Lowe04,MikolajczykS02,MikolajczykS04} to handle the viewpoint changes across different images. 
Since these approaches determine the scale of interested points in very local area, their precision and discriminability will be lost quickly. Besides, these methods mainly focus on the localizations and scales of corner points while ignoring the differences between different type of junctions. 

To overcome these shortcomings, the ACJ detector~\cite{XiaDG14} was proposed to detect and characterize junctions with non-linear scale space. In this work, an \emph{a-contrario} approach is proposed for determining the location and branches of junctions with interpretable isotropic scales, which characterizes the ray segments as junction branches and locations explicitly.
The scales for detected junctions correspond to the optimal size at which one can observe the junction in the image. 

Similar to junction detection, there is an elegant detector named edge based region (EBR) detector~\cite{TuytelaarsG04}  for detecting affine invariant regions by estimating relative speed for two points that move away from a corner in both directions along the curve edges. This work can be regarded as a kind of junction detector in curve dominated images. The straight edges which are common in indoor scenario cannot be tackled in this way.

Although above mentioned approaches can extract junctions, their geometric representation is not exploited sufficiently. The scales estimated by these methods are local and insufficient for characterizing indoor scenes. 

\subsection{Junction matching}
Junction matching has been attended since early years and shown promising  matching accuracy\cite{ShenP00,VincentL04}. 

In \cite{ShenP00}, 
a model for estimating endpoints of junction branches is proposed which is very close to our work that estimating anisotropic scales for each branch. 
Differently, their approach~\cite{ShenP00} requires a roughly estimated fundamental matrix while our proposed method 
estimating anisotropic scales for each branch directly without fundamental matrix. For known fundamental matrix, the local homography between a pair of junctions can be estimated to produce more accurate results and refining epipolar geometry meanwhile~\cite{VincentL04}. These results are very related to recent approach for the hierarchical line segment matching approach LJL~\cite{LiYLLZ16}. In this work, detected line segments are used to generate junctions with virtual intersections in the first stage. After that, junctions are regarded as key-points for matching initially. Finally the epipolar geometry induced from initial matching is used to estimate line segment correspondences with local junctions. The matching accuracy in fact relies on the descriptors of virtual intersections.
Although their matching results are promising, 
the problem of estimating epipolar geometry need to other ways.

\subsection{Indoor image matching with geometric structure}
Most of indoor scenes can be described by using simple geometric elements such as points and line segments. As a combination of points and lines, junction is also a sort of useful geometrical structure for indoor scene. 
There has been many approaches such as Canny edge detector~\cite{Ruff87} and line segment detector (LSD)~\cite{GioiJMR10,GioiJMR12} to extract line segments. LSD, which can produce more complete line segments than canny edge without any parameter tuning procedure, has been applied in many tasks such as line-segments matching~\cite{LiYLLZ16} and 3D reconstruction~\cite{RamalingamASLP15}. Compared with key-points, line-segments can produce more complete result that contain the primary sketch for the scene.

Most of algorithms for line-segments matching rely on key-point correspondences. More precisely, key-points for an input image pair are firstly detected by using SIFT~\cite{Lowe04} or other detectors while estimating the epipolar geometry between the image pair by using RANSAC~\cite{FischlerB81} and its variants. Based on the fundamental matrix $F$ induced by key-points matching, many approaches such as line-point-invariant (LPI)~\cite{FanWH12B} and line-junction-line (LJL)~\cite{LiYLLZ16} can match line segments correctly. LPI has ability to handle the relation between line-segments and matched key-points with viewpoint changes. LJL\cite{LiYLLZ16} method matches image pairs in multiple stages. In the first, detected line-segments are intersected with appropriate threshold to produce junctions and matching these intersections in the same way with key-points matching. Then, local homography are estimated for these junctions with the estimated fundamental matrices from key-points matching results.
Although these approaches produce good performance in many cases, the matching results are in favor of matching lines instead of line-segments. Their results show that lines are matched while the endpoints of line segments are not matched very well. Except for the reason that the estimated epipolar geometry is sometime erroneous, there is a important reason for failure of line-segments matching that line segment detectors can not guarantee that the line-segments are consistent across imaging condition varying. In many situations, a line segment $l_A$ detected in image $I_A$ might be decomposed to two or more collinear line segments $l_B^1,\ldots,l_B^k$ in another image $I_B$. In this case, the results of line matching can be regarded as correct if the line in $l_A$ is corresponding with $l_B^1,\ldots,l_B^k$. However, in the aspect of line-segments matching, there exists no correct corresponding line-segments for $l_A$ in image $I_B$. On the other hand, the existing line segment matchers rely on the results of key-point matching. Once the key-points matching failed or inaccurate, the induced result of line segment matching will be affected in some extent. 
\section{Problem Statement}
\label{sec:problems}
\subsection{Junction Model}
\begin{figure}
	\centering
	\includegraphics[width=0.98\linewidth]{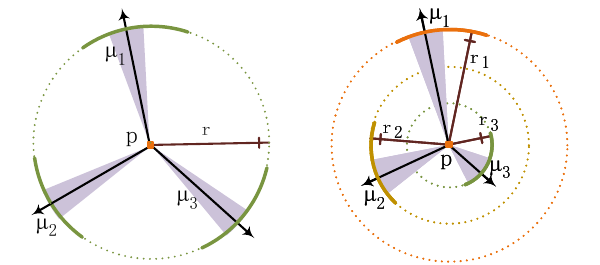}
	\caption{Template of isotropic-scale junction (left) defined in ACJ~\cite{XiaDG14} and anisotropic-scale 		junction (ASJ) (right) proposed in our work.}
	\label{fig:junction-model}
\end{figure}
The early researches for junction detection usually focus on the orientations of branches and the locations while ignoring the length or scale of each branch. 
Even though the junction locations and orientation of branches are important to depict geometric structure for images, lacking scale of branches limits their performance for image matching. Motivated by this, we want to propose a new junction model for characterizing junction better. We define the our junction model by considering the endpoint of each branch. Since the length of every branch is possible to be different, we call our model as \emph{anisotropic-scale junction}. As a special case, junction model with isometric branches is called \emph{isotropic-scale junction}.
\begin{definition}[Anisotropic-scale junction]
	\label{def:aniso-junction}
	An anisotrpic-scale junction with $M$ branches starting at the same location $\bm{p}$ is denote as
	\begin{equation}
	\jmath = \left\{\bm{p}, \left\{r_i,\theta_i \right\}_{i=1}^M\right\}
	\end{equation}
	where $r_i$ and $\theta_i$ are the scale and orientation for $i$-th branch, $M$ is the number of branches.
\end{definition}

Fig.~\ref{fig:junction-model} provides an example for the difference between anisotropic-scale (left) and isotropic-scale (right) junctions is shown. The \emph{isotropic-scale junction} is actually a special case for the anisotropic model when the length of all branches are identical. 

\subsection{Detecting Junction Locations and Isotropic Branches}
Since junction is formed by several intersected line segments, the problem of localizing the intersection and identifying the normal angle of these line segments is easier to be focused. Once the isotropic junction model is defined, this problem becomes a template matching problem. Based on this idea, Xia~\emph{et al.} exploited the junction-ness for branches with given scale $r$ and orientations $\theta_i$ and then an \emph{a-contrario} approach is derived to determine meaningful junctions for input images~\cite{XiaDG14}. The junction-ness for given scale $r$ and orientation $\theta_i$ actually contains the neighbor information of normalized gradient. Different from points, the neighborhood for a given scale $r$ and $\theta_i$ is a sector. As shown in the left of Fig.~\ref{fig:junction-model}, the dark area with $\theta_1$ represent the sector neighbor of the branch. The sector neighbor for given location $\bm{p}$, scale $r$ and orientation $\theta$ can be  denoted mathematically as 
\begin{equation}
\label{eq:sector-def}
\begin{split}
S_{\bm{p}}(r,\theta):= \left\{
\bm{q}\in\Omega; \right.
\bm{q}\neq\bm{p},
\left\|\vec{\bm{pq}}\right\|\leq r,
\\
\left.
d_{2\pi}(\alpha(\vec{\bm{pq}}),\theta)\leq \Delta(r)
\right\}
\end{split}.
\end{equation}
where the $\Delta(r)$ is defined as $\frac{\tau}{r}$ with some predefined parameter $\tau$, $\Omega$ is the domain of input image, $d_{2\pi}$ is the distance along the unit circle, defined as $d_{2\pi}(\alpha,\beta) = \min\left(|\alpha-\beta|,2\pi-|\alpha-\beta|\right)$ and $\alpha(\vec{\bm{pq}})$ is the angle of the vector $\vec{\bm{pq}}$ in $[0,2\pi]$.

Since a junction is formed by edges and corner points, the normal angle for gradient should be consistent with the orientation of branches. Followed with this idea, if most of points $\bm{q}\in S_{\bm{p}}(r,\theta)$ have close normal angles with orientation $\theta$, the corresponding scale $r$ and orientation $\theta$ should be meaningful to be a branch of the junction. For a given sector $S_{\bm{p}}(r,\theta)$, the junction-ness can be measured by 
\begin{equation}
\label{eq:junction-ness-branch}
\omega_{\bm{p}}(r,\theta) = \sum_{\bm{q}\in S_{\bm{p}}(r,\theta)}\gamma_{\bm{p}}(\bm{q}),
\end{equation}
and $\gamma_{\bm{p}}(\bm{q})$ is the pairwise junction-ness with
\begin{equation}
\begin{split}
\label{eq:junction-ness-pairwise}
\gamma_{\bm{p}}(\bm{q}) = \left\|\nabla\tilde{I}(\bm{q})\right\|
\cdot \max(\left|\cos(\phi(\bm{q})-\alpha(\vec{\bm{pq}}))\right|
\\ -\left|\sin(\phi({\bm{q}})-\alpha(\vec{\bm{pq}}))\right|,0)
\end{split},
\end{equation}
where the $\left\|\nabla\tilde{I}(\bm{q})\right\|$ is the norm of normalized gradient at point $\bm{p}$, $\phi(\bm{q})$ for pixel $\bm{q}$ is defined as $\phi(\bm{q}) = (\arctan\frac{I_y(\bm{q})}{I_x(\bm{q})}+\pi/2)~modulo~(2\pi)$, $I_x,I_y$ are the partial derivative of input image in $x$ and $y$ direction.

For the isotropic scale junction with two or more branches, the minimal junction-ness for one of the branches is used to describe the junction-ness for the entire junction with the equation \eqref{eq:junction-ness-entire}
\begin{equation}
\label{eq:junction-ness-entire}
t(\jmath):= \min_{m=1,\ldots,M}\omega_{\bm{p}}(r,\theta_m),
\end{equation}
where the number $M$ and $m$ represent the total number of branches and branch index for the junction $\jmath$.

\subsection{Analysis for Estimating Anisotropic-scale Branches}
Although the equation~\eqref{eq:junction-ness-entire} measures junction-ness for a given junction, it does not contain any anisotropic scale for branches. Such definition of junction-ness only keeps the information that each branch's scale $r_i$ is larger than $r$ and it cannot be used for handling more sophisticated transformations such as affine transform and projective transform. To overcome this problem, we define the anisotropic-scale junctions with independent scales in Def.~\ref{def:aniso-junction}. The difference between isotropic-scale junction and the anisotropic-scale one can be observed in Fig.~\ref{fig:junction-model}.
It is easy to see that the junction-ness for entire junction defined in Eq.~\eqref{eq:junction-ness-entire} cannot be used to exploit independent scales $r_i$. Fortunately, the isotropic-scale junctions detected by ACJ~\cite{XiaDG14} is meaningful and the problem of estimating scale $r_i$ and orientation $\theta_i$ can be simplified to estimating only scale $r_i$ with given location $\bm{p}$ and orientation $\theta_i$. In other words, for the detected isotropic junctions, we need to exploit a robust method to estimate the length of corresponding ray segment with specific orientation $\theta$.

One plausible way to model the unknown scale with respect to given location $\bm{p}$ and orientation $r$ is that simply modify the junction-ness defined in Eq~\eqref{eq:junction-ness-branch} to $\omega_{\bm{p}}(r)$ with specific $\theta$. Then, the \emph{a-contrario} approach in~\cite{XiaDG14} seems to be feasible to check whether the scale $r$ is $\varepsilon$-meaningful.
The corresponding cumulative distribution function (CDF) used to get $\varepsilon$-meaningful scale $r$ can be formulate to 
\begin{equation}
\label{eq:acj-distribution-conv}
F(t;J(r,\theta)) = \mathbb{P}\{\omega_{\bm{p}}\geq t\} = \int_{t}^{+\infty}
d\left(\mathop{\star}_{j=1}^{J(r,\theta)} p\right)
\end{equation}
where the $p$ represents the distribution of random variable $\omega_{\bm{p}}(r,\theta)$ with
\begin{equation}
\label{eq:acj-distribution}
p(z) = \frac{1}{2}(\delta_0(z) + \frac{2}{\sqrt{\pi}}e^{-\frac{z^2}{4}} {\rm erfc}(\frac{z}{2}))\mathbf{1}_{z\geq 0}.
\end{equation}
$J(r,\theta)$ is the number of pixels in corresponding sector neighbor and the operator $\mathop{\star}_{j=1}^{J(r,\theta)}$ produces the convolutional probability density function (PDF) with $J(r,\theta)$ times, which actually represents the random variable of $\omega_{\bm{p}}(r,\theta)$.
The $\varepsilon$-meaningful scale for given orientation and location can be determined by the inequality
\begin{equation}
\label{eq:NFA-xia}
\NFA(\jmath) := \#\mathcal{J}(1)\cdot F(t;J(r,\theta))\leq \varepsilon,
\end{equation}
where $\#\mathcal{J}(1)$ is the number of test for junctions with $1$ branch.

\begin{figure}
	\centering
	\includegraphics[width = 0.4\linewidth]{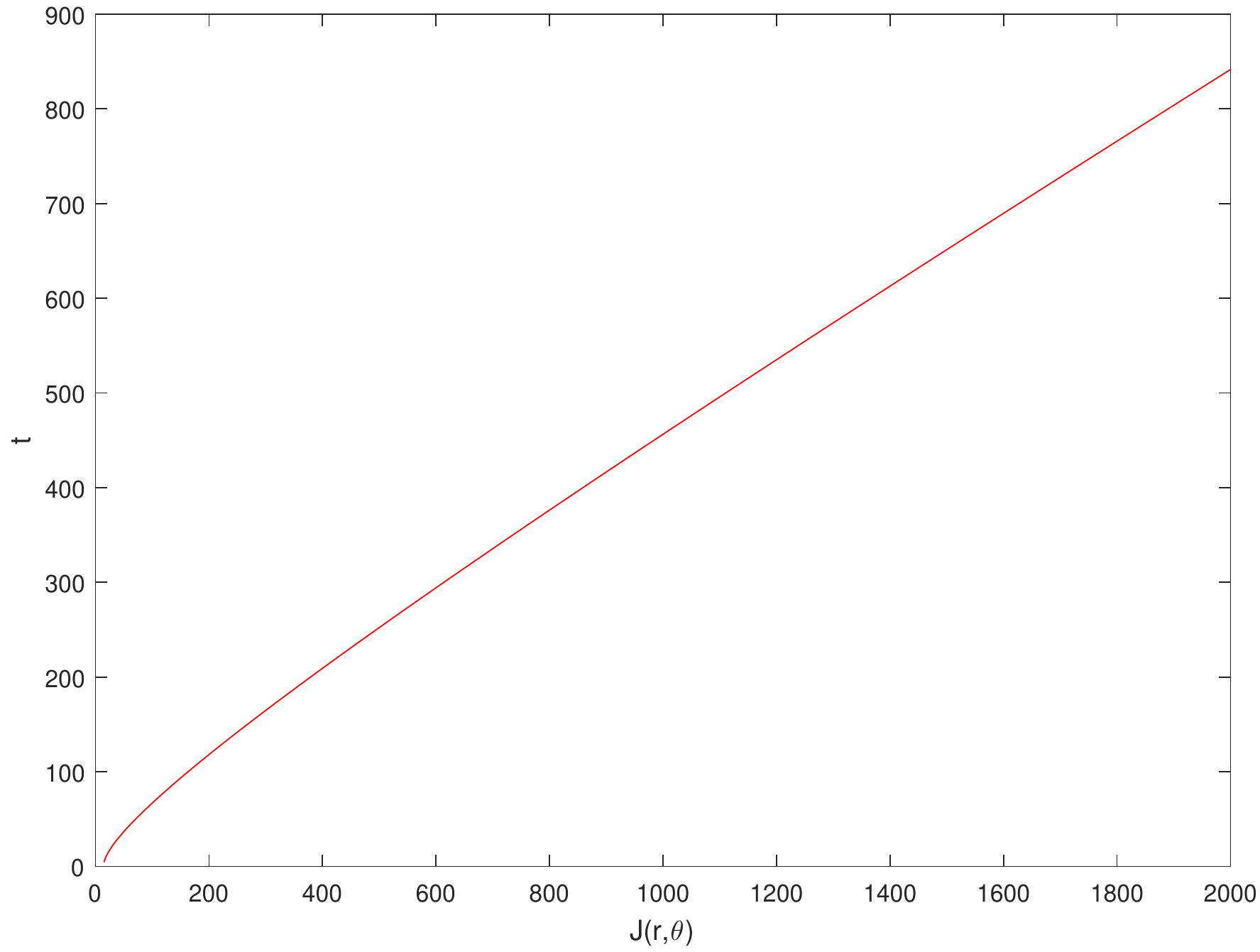}
	\caption{The relationship between the convolution times $J(r,\theta)$ and corresponding minimal value with $F(t;J(r,\theta)) = 0$}
	\label{fig:plot-min-zero}
\end{figure}

However, the NFA defined in Eq.~\eqref{eq:NFA-xia} has to face the fact that there exist several junctions in indoor images which have extremely large scale branches. This fact would lead to the above inequality disabled. To illustrate this problem, we studied the relationship between convolution times $J(r,\theta)$ with the minimal junction-ness that can make the probability $F(t;J(r,\theta)) = 0$. As shown in Fig.~\ref{fig:plot-min-zero}, if the value of $\omega_{\bm{p}}(r,\theta)$ is greater than $\frac{J(r,\theta)}{2}$, the probability of $F(t;J(r,\theta))$ will be equal to $0$ constantly, which may cause the inequality degenerated to $0\leq \varepsilon$. In fact, the pairwise junction-ness defined in Eq~\eqref{eq:junction-ness-pairwise} can reach to $1$ and then the $\omega_{\bm{p}}(r,\theta)$ will be equal to $J(r,\theta)$. Therefore, the junction-ness in~\cite{XiaDG14} is infeasible to model the unknown scale.
\section{An a-contrario model for anisotropic-scale junction detection}
\label{sec:detection}
To solve the problems addressed in Sec.~\ref{sec:problems}, we derive a differential junction-ness model for depicting scale with given location and orientation. Since the scale $r_i$ for each branch of junction $\jmath$ is irrelevant, we just model the endpoint of each branch independently. 

\subsection{Differential Junction-ness Model}
Suppose the isotropic junctions have been detected in a small scale $r_0$, the inherent scales of branches will be greater than $r_0$. If we increase the scale $r_0$ to larger $r_1$, though the junction-ness is still larger, the error $\omega_{err} = \omega_{\bm{p}}(r_1,\theta) - \omega_{\bm{p}}(r_0,\theta)$ will not be increased significant. 
A reasonable way to recognize the un-significant variation is to study the variation of $\omega_{\bm{p}}(r,\theta)$ with respect to $r$ increased. 
Here, we first reformulate the junction-ness for a branch \eqref{eq:junction-ness-branch} in continuous form. The junction-ness for position $\bm{p}$, scale $R$ and orientation $\theta$ is
\begin{equation}
\label{eq:strength-continous}
\omega_{\bm{p}}(r,\theta)
= \int_0^r dr\int_{\theta-\delta(r)}^{\theta+\delta(r)}
\gamma_{\bm{p}}\left(\bm{p}+r\left(\cos\psi,\sin\psi\right)\right)d\psi,
\end{equation}
where the $\delta(r)$ is the angle width for given scale, here, we select $\delta(r) = \frac{\tau}{r}$. The descrete partial derivative $\frac{\partial \omega_{\bm{p}}(r,\theta)}{\partial r}$ is given by

\begin{equation}
\begin{split}
\frac{\partial \omega_{\bm{p}}(r,\theta)}{\partial r}
= \sum_{i} \gamma_{\bm{p}}\left(\bm{p}+r\begin{bmatrix}
\cos\psi_i\\\sin\psi_i
\end{bmatrix}\right)\\
+ \delta'(r)\sum_{i} \gamma_{\bm{p}}\left(\bm{p}+r_i\begin{bmatrix}
\cos(\theta-\delta(r))\\
\sin(\theta-\delta(r))
\end{bmatrix}\right)\\\
+ \delta'(r)\sum_{i} \gamma_{\bm{p}}\left(\bm{p}+r_i\begin{bmatrix}
\cos(\theta+\delta(r))\\
\sin(\theta+\delta(r))
\end{bmatrix}\right)
\end{split},
\end{equation}
where $\psi_i$ is the $i$-th sample angle in the range $\left[\theta-\delta(r),\theta+\delta(r)\right]$ and $r_i$ is the $i$-th sample point in the range $[0,r]$.
\subsection{Null Hypothesis and Distribution}
\label{sec:Null-Hypothesis-1}
After the differential junction-ness model built, we need to find a robust way to check if the value of ${\partial\omega_{\bm{p}(r,\theta)}}/{\partial r}$ for specific $r$ is significant enough. One way to achieve this goal is developing an \emph{a-contrario} approach to control the threshold automatically. Since our work is an extension of ACJ~\cite{XiaDG14}, the null hypothesis here should be same, we say the variables $\left\|\nabla\tilde{I}(\bm{q})\right\|$ and $\phi(\bm{q})$ follow the null hypothesis $\mathcal{H}_0$ if
\begin{enumerate}
	\item $\forall\bm{q}\in\Omega$, $\left\|\nabla\tilde{I}(\bm{q})\right\|$ follows a Rayleigh distribution with parameter 1;
	\item $\forall\bm{q}\in\Omega$, $\phi(\bm{q})$ follows a uniform distribution over $[0,2\pi]$;
	\item All of the random variables $\left\|\nabla\tilde{I}(\bm{q})\right\|, \phi(\bm{q})\}_{\bm{q}\in\Omega} $ are independent each other.
\end{enumerate}
According to the dicussion in \cite{XiaDG14}, every $\gamma_{\bm{p}}(\bm{q})$ follows the distribution \eqref{eq:acj-distribution} independently. The random variable 
$\frac{\partial \omega_{\bm{p}(r,\theta)}}{\partial r}$ follows the distribution of the random variable
\begin{equation}
S_r = \sum_{i=1}^{m} X_i
+ \delta'(r)\sum_{i=1}^{k} (Y_i + Z_i),
\end{equation}
where the random variable $X_i,Y_i,Z_i$ follow the distribution in equation \eqref{eq:acj-distribution} , $m$ is the number of sampling points for $\psi_i$ and $k$ is the number of sampling points for $r_i$. The function $\delta'(r)$ will be very small for reasonable $r$ (for example, $r\geq4$ induced $|\delta'(r)| = \frac{\tau}{r^2} \leq \frac{\tau}{16}$) since the parameter $\tau$ should have small values. Hence, the random variable could be approximated with
$S_r \approx \sum_{i=1}^{m} X_i$
for computational simplicity.
In practice, $k$ is larger than 10 and therefore the PDF of $\sum_{i=1}^{k} (Y_i+Z_i)$ can be apprixmated accurately by using the Central Limit Theorem as
\begin{equation}
f(t) = \frac{1}{\sqrt{4k\sigma^2\pi}}
\exp\left(-\frac{(t-2k\mu)^2}{4k\sigma^2}\right)
\end{equation}
where $\mu$ and $\sigma^2$ are the expectation and variance of \eqref{eq:acj-distribution}. The PDF of $\delta'(R)\left(\sum_{i=1}^{k}Y_i+Z_i\right)$ is
\begin{equation}
\label{eq:distribution-approximate}
\tilde{f}(t)
= \frac{R^2}{\tau}f(-\frac{R^2}{\tau}t)
= \frac{1}{\sqrt{2\pi\cdot  2k\sigma^2\frac{\tau^2}{R^4}}}
\exp\left(-\frac{t+\frac{2\tau}{R^2}k\mu}{2\cdot2k\sigma^2\frac{\tau^2}{R^4}}\right),
\end{equation}
which is the Gaussian distribution with mean $-\frac{2\tau}{R^2}k\mu$ and variance $2k\sigma^2\tau^2/R^4$. Meanwhile, the random variable $\sum_{i=1}^{m} X_i$ follows $\mathcal{N}(m\mu,m\sigma^2)$. Therefore, the random variable $S$ follows the distribution $\mathcal{N}(m\mu - 2k\mu\frac{\tau}{R^2},m\sigma^2 + 2k\sigma^2\tau^2/R^4)$ approximately.

The probability $\mathbb{P} \left( S_r \geq \frac{\partial \omega_{\bm{p}}(r,\theta)}{\partial r} \right) $ for given $r$ and $\theta$ follows the distribution $f(t)$
\begin{equation}
\mathbb{P}_{\bm{p}}(r,\theta):=\mathbb{P} \left(S_r \geq \frac{\partial \omega_{\bm{p}}}{\partial r}\right) = \int_{{\partial \omega_{\bm{p}}}/{\partial r}}^{\infty} d\left(\mathop{\star}_{i=1}^M p\right),
\end{equation}
describes the fact that scale cannot be increased with a sufficient small incremental at $R$ along orientation $\theta$ under the hypothesis $\mathcal{H}_0$. The smaller probability $\mathbb{P}_{\bm{p}}(r,\theta)$ is, the more confident that scale $r$ is a reasonable scale. The small probability $\mathbb{P}(r,\theta)$ means that the point $\bm{p} + r\cdot[\cos\theta,\sin\theta] $ belongs to the branch with high possibility. Ideally, the existed branch should produce a series small probability in a interval $[r_0,r_1]$. Then, the (maximum) scale of the branch should be defined as $r_1$. We use the probability $\mathbb{P}_{\bm{p}}(r,\theta)$ to check if the point $\bm{p}_{r}^{\theta} = \bm{p} + r\cdot[\cos\theta,\sin\theta]^T$ belongs to the branch.

\subsection{Number of Test and Number of False Alarms}
\label{sec:model1}
In last subsection, we conclude that \emph{sufficient} small probability of $\mathbb{P}_{\bm{p}}(r,\theta)$ indicates that the point with certain direction $\theta$ and radius $R$ belongs to the branch more probably. The definition of \emph{sufficient} probability need to be cleared. According to the Helmholtz principle, we bound the \emph{sufficient} probability with the expectation of the number of occurrences of this event is less than $\varepsilon$ under the \emph{a-contrario} random assumption~\cite{DesolneuxMM07} with
$$
\NFA(r,\bm{p},\theta) = N_s\cdot \mathbb{P}_{\bm{p}}(r,\theta) \leq \varepsilon,
$$
where the $N_s$ denotes the number of occurrences of the point occurs along the given location and orientation. Since the location and orientation of the branch are known, expected number of false alarms  should be smaller than $\sqrt{NM}$ where $N$ and $M$ are the number of rows and columns of the corresponding image. When the point $\bm{p}_r^{\theta}$ $\forall r\in (0,r_1]$ rejects the hypothesis $H_0$, the scale of the branch should be $r_1$. The scale $R$ is called as the maximum (meaningful) scale of the branch if the scale $R$ is the maximum scale that satisfies inequality
$$
\NFA(r,\bm{p},\theta) = \sqrt{NM}\cdot \mathbb{P}_{\bm{p}}(r,\theta) \leq \varepsilon, \forall r\in (0,R].
$$
Usually, the $\varepsilon$ is defined as $1$, which means the expected Number of False Alarm is not larger than 1.

\subsection{Scale Ambiguity for Branches}
\begin{figure}
	\centering
	\begin{tikzpicture}[scale=0.8]
	\node[anchor=south west,inner sep=0] at (0,0)
	{\includegraphics[width=0.26\textwidth]{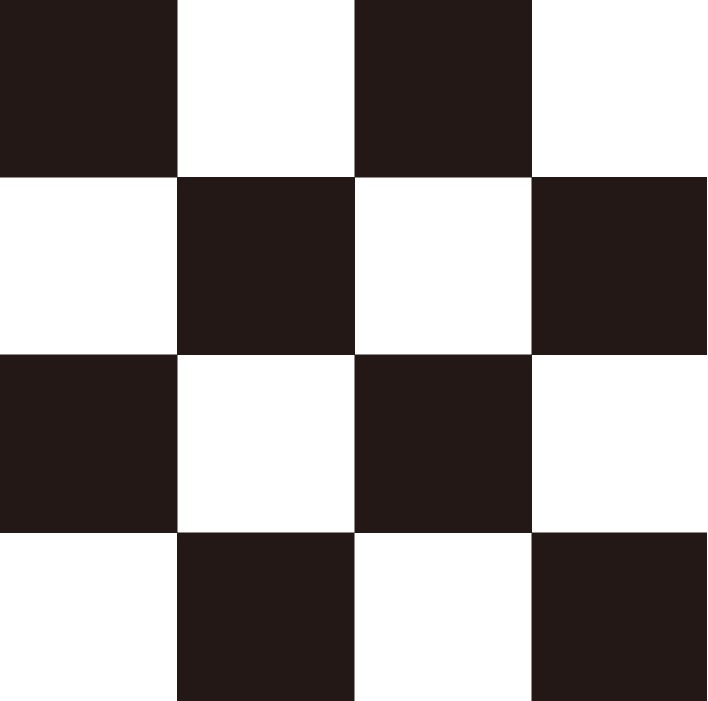}};
	\coordinate (p0) at (0,1.33);
	\coordinate (p1) at (1.37,1.33);
	\coordinate (p2) at (2.75,1.33);
	\coordinate (p3) at (4.13,1.33);	
	\coordinate (p4) at (5.47,1.33);	
	\coordinate (q0) at (1.37,0.00);
	\coordinate (q1) at (p1);
	\coordinate (q2) at (1.37,2.68);
	\coordinate (q3) at (1.37,4.06);
	\coordinate (q4) at (1.37,5.43);	
	\coordinate (o0) at (2.75,0.00);
	\coordinate (o1) at (p2);
	\coordinate (o2) at (2.75,2.68);
	\coordinate (o3) at (2.75,4.06);
	\coordinate (o4) at (2.75,5.43);	
	\draw[red,thick=8pt] (p0) -- (p1) -- (p2) -- (p3) -- (p4);
	\draw[green,thick=8pt] (q0) -- (q1) -- (q2) -- (q3) -- (q4);
	\draw[orange,thick=8pt] (o0) -- (o1) -- (o2) -- (o3) -- (o4);
	\draw[blue] (p1) circle [radius = 0.6];
	\draw[cyan] (p2) circle [radius = 0.6];
	\filldraw[teal] (p0) circle [radius = 0.05];	
	\filldraw[teal] (p1) circle [radius = 0.05];
	\filldraw[teal] (p2) circle [radius = 0.05];	
	\filldraw[teal] (p3) circle [radius = 0.05];	
	\filldraw[teal] (p4) circle [radius = 0.05];
	\filldraw[teal] (q0) circle [radius = 0.05];
	\filldraw[teal] (q2) circle [radius = 0.05];	
	\filldraw[teal] (q3) circle [radius = 0.05];	
	\filldraw[teal] (q4) circle [radius = 0.05];
	\filldraw[teal] (o0) circle [radius = 0.05];
	\filldraw[teal] (o2) circle [radius = 0.05];
	\filldraw[teal] (o3) circle [radius = 0.05];
	\filldraw[teal] (o4) circle [radius = 0.05];	
	\node[below left] at (p0) {$\bm{p}_0$};
	\node[below left] at (p1) {$\bm{p}_1$};
	\node[below right] at (p2) {$\bm{p}_2$};
	\node[below left] at (p3) {$\bm{p}_3$};
	\node[below right] at (p4) {$\bm{p}_4$};
	\node[below left] at (q0) {$\bm{q}_0$};
	\node[below right] at (q2) {$\bm{q}_2$};
	\node[below left] at (q3) {$\bm{q}_3$};
	\node[below right] at (q4) {$\bm{q}_4$};
	\node[below right] at (o0) {$\bm{o}_0$};
	\node[below left] at (o2) {$\bm{o}_2$};
	\node[below right] at (o3) {$\bm{o}_3$};
	\node[below left] at (o4) {$\bm{o}_4$};	
	\end{tikzpicture}
	\centering
	\caption{Scale ambiguity for branches. The junction $\jmath_1$ and $\jmath_2$ located at $\bm{p}_1$ and $\bm{p}_2$ have more than one scales respectively.}
	\label{fig:scale-ambiguity}
\end{figure}
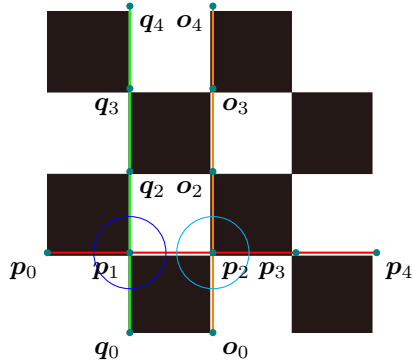

Junctions are located at the intersections of line segments. Suppose there exist two junctions $$\jmath_1 = \{\bm{p}_1,\{r_1^1,\theta_1^1\},\{r_1^2,\theta_1^2\}\},$$ $$\jmath_2 = \{\bm{p}_2,\{r_2^1,\theta_2^1\},\{r_2^2,\theta_2^2\}\}$$ where the two-tuples $\{\{r_i^j,\theta_i^j\}\}$ denotes the scale and orientation for the $j$-th branch of the $i$-th junction and $\bm{p}_i$ is location of the $i$-th junction. In the case that the junction $\jmath_2$ is located at $\bm{p}_2 = \bm{p}_1 + r_1^1\left[\cos\theta_1^1, \sin\theta_1^1 \right]^T$ and $\theta_1^1 = \theta_2^1$, the scale ambiguity occurs since the line segment $\bm{p}_1\bm{p}_2$ and the branch $\{r_2^1,\theta_2^1\}$ are co-linear. The scale of the first branch of $\jmath_1$ can be regarded as either $r_1^1$ or $r_1^1+r_2^1$. For example, there are two junctions $\jmath_1$ and $\jmath_2$ located at $\bm{p}_1$ and $\bm{p}_2$ respectively in the Fig.~\ref{fig:scale-ambiguity}. The branch along the direction of ${\bm{p}_1\bm{p}_2}$ for $\jmath_1$ and $\jmath_2$ are co-linear with the line segment marked as red. For the branch of $\jmath_1$, its scales are $\left\|\bm{p}_2-\bm{p}_1\right\|$, $\left\|\bm{p}_3-\bm{p}_1\right\|$ and $\left\|\bm{p}_4-\bm{p}_1\right\|$ while the scales of the branch of $\jmath_2$ are $\left\|\bm{p}_3-\bm{p}_2\right\|$ or $\left\|\bm{p}_4-\bm{p}_2\right\|$. To eliminate the ambiguity, we define the scale for a branch as follow
\begin{definition}[Scale of a branch]
	\label{def:scale}
	Suppose there exist a branch starting at point $\bm{p}$ in the direction $\theta$,  the possible salient scales are $r_1,r_2,\ldots,r_m$, we define the scale of this branch $r^*$as
	$$
	r^* = \max_{i} r_i.
	$$
\end{definition}
The branch with such scale is more stable and more global than other features. However, there exist some challenges to estimate such scales from images. Most existing approaches and the model proposed in Sec.~\ref{sec:model1} estimate the line segment or branches based on orientations of level-lines extracted from the  gradient of image~\cite{GioiJMR10}.
The line segment detected from the image in Fig.~\ref{fig:scale-ambiguity} could be either $\bm{p}_1\bm{p}_2$ or $\bm{p}_1\bm{p}_3$ since the level-line around the points $\bm{p}_2$  have probability to aligned with the orientation of vector $\bm{o}_2-\bm{p}_2$, which will lead to the line segment that are co-linear with the branch of $\jmath_1$ across the point $\bm{p}_2$ to $\bm{p}_3$ or $\bm{p}_4$. When the viewpoint changed, illumination varied or noise increased, the orientations of level-lines around $\bm{p_2}$, $\bm{p}_3$ and $\bm{p}_4$ will be changed with uncertainty. Then, the scale cannot be estimated robust for different imaging conditions.

Fortunately, the inherent property for location of junctions is stable whatever the imaging condition is. Although the orientations of level-lines around the locations of junctions will change with uncertainty, most of them are still aligned to one of the lines that intersects the junction. Motivated by this, we use the very local isotropic-scale junctions in a small neighbor(e.g. $5\times 5$ or $7\times7$ window size) instead of gradient field and level-lines. For a pixel in an image, we calculate the junction-ness for different orientations in a small neighbor according to ACJ~\cite{XiaDG14} algorithm as
\begin{equation}
\label{eq:junction-ness-branch-fix-scale}
\omega_{\bm{p}}(\theta) = \frac{1}{\sigma}\left(\sum_{\bm{q}\in S_{\bm{p}}(\theta)}\frac{\gamma_{\bm{p}}(\bm{q})}{\sqrt{n}}
-\sqrt{n}\mu\right)
,
\end{equation}
where $S_{\bm{p}}(\theta)$ is defined in \eqref{eq:sector-def} with fixed radius (eg. $r=3,5,7$), $n$ is the cardinal number of set $S_{\bm{p}}(\theta)$. $\mu$ and $\sigma^2$ are the mean and variance defined in \eqref{eq:distribution-approximate}. Then, we leverage the non-maximal-suppression (NMS)~\cite{NeubeckG06} to obtain the very local junctions and filter out branches for these junctions with non-meaningful NFA values according to \eqref{eq:NFA-xia}. These very local junctions are denoted as $\jmath_{\bm{p}} = \{\theta_{\bm{p}}^i,\omega_{\bm{p}}^i,\NFA_{\bm{p}}^i \}_{i=1}^K$, where the $\omega_{\bm{p}}^i$ and the $\NFA_{\bm{p}}^i$ is the strength and corresponding NFA value for branch with $\theta_i$ orientation. In the case that pixel $\bm{p}$ is on (around) an edge, there will be two $\theta_i$ that align to the orientation of this edge up to $\pm\pi$. If the pixel $\bm{p}$ is around another junction, there will be multiple orientations aligned with different branches of this junction. Meanwhile, we incorporate the strength $\omega_{\bm{p}}^i$ instead of the norm of (normalized) gradient with into the \emph{a-contrario} model proposed in Sec.~\ref{sec:model1} with modified probabilistic distribution.
\subsection{Modified Probabilistic Distribution}
For the sake of estimating scale for a branch with definition \ref{def:scale}, the functions $\gamma_{\bm{p}}$ and $\omega_{\bm{p}}(r,\theta)$ measuring the junction-ness should be changed to
\begin{equation}
\label{eq:junction-ness-pairwise-modified}
\begin{split}
\tilde{\gamma}_{\bm{p}}(\bm{q}) = \omega_{\bm{q}}^i
\cdot \max(\left|\cos(\theta_{\bm{q}}^i-\alpha(\vec{\bm{pq}}))\right|-
\\ \left|\sin(\theta_{\bm{q}}^i-\alpha(\vec{\bm{pq}}))\right|,0)
\end{split}
\end{equation}
and
\begin{equation}
\label{eq:junction-ness-modified}
\tilde{\omega}_{\bm{p}}(r,\theta) = \sum_{\bm{q}\in S_{\bm{p}}(r,\theta)}
\tilde{\gamma}_{\bm{p}}(\bm{q}),
\end{equation}
where the index $i$ in Eq.~\eqref{eq:junction-ness-pairwise-modified} is the orientation $\theta_{\bm{q}}^i$ that is most close to $\theta$.

According to the Central Limit Theorem(CLT), the random variable $\omega_{\bm{p}}^i$ follows the Gaussian distribution with mean $0$ and variance $1$, the distribution for $\tilde{\gamma}_{\bm{p}}(\bm{q})$ is
\begin{equation}
\label{eq:distribution-updated}
\tilde{p}(z) = \frac{1}{2}\delta_0(z) + \mathbf{1}_{z\neq 0}\frac{\sqrt{2}}{\sqrt{\pi^3}}\int_{0}^{1}\frac{1}{y}e^{-\frac{z^2}{2y^2}}\frac{1}{\sqrt{2-y^2}}dy,
\end{equation}

Then,  the null Hypothesis discussed in Sec.~\ref{sec:Null-Hypothesis-1} is updated to
\begin{equation}
\label{eq:probability-updated}
\mathbb{P}_{\bm{p}}(r,\theta):=\mathbb{P} \left(S_r \geq \frac{\partial \tilde{\omega}_{\bm{p}}}{\partial r}\right) = \int_{{\partial \tilde{\omega}_{\bm{p}}}/{\partial r}}^{\infty} d\left(\mathop{\star}_{i=1}^M \tilde{p}\right).
\end{equation}
\subsection{Junction Detection}
So far, the \emph{a-contrario} approach for anisotropic scale estimation is derived. For an input image, isotropic junctions and local junctions for each pixel are firstly detected by ACJ~\cite{XiaDG14} for initialization. The results for junctions are denoted as $\{\jmath_i\}_{i=1}^N$ and local junctions at fixed small scale (usually $r=3,5,7$) for every pixels are $\jmath_{\bm{p}} = \{\theta_{\bm{p}}^i,\omega_{\bm{p}}^i,\NFA_{\bm{p}}^i \}_{i=1}^K$ where $\bm{p}$ is the coordinate of a pixel.

We estimate the scale $r_i^j$ for branch $\theta_i^j$ according to the Number of False Alarm
$$
\NFA(r_i^j,\bm{p}_i,\theta_i^j) = \sqrt{NM}\cdot\mathbb{P}_{\bm{p}}(r_i^j,\theta_i^j)\leq \varepsilon,
$$
where the probability $\mathbb{P}_{\bm{p}}(r,\theta)$ is the updated version in Eq.~\eqref{eq:probability-updated}. The scale $r_i^j$ is searched starting at $r_i$ until the NFA is larger than $\varepsilon$.

The accuracy for orientations of branches detected by ACJ~\cite{XiaDG14} is depend on the scale $r_i$ which is bounded by a predefined parameter and hence noised. The scales for ASJ is more sensitive to the noise which should be refined. A branch with the most accurate orientation $\theta$ should have the maximum junction-ness with the scale $r_{\theta}$, we optimize the objective function
\begin{equation}
	\hat{\theta} =\arg\max_{\theta} \sum_{\bm{q}\in S_{\bm{p}}(r_{\theta},\theta)}\gamma_{\bm{p}}(\bm{q}),
\end{equation}
to refine the orientation for $\theta$ and check the branch with orientation $\hat{\theta}$ and scale $r_{\hat{\theta}}$ is $\varepsilon$-meaningful branch.

\section{ASJ Matching for Indoor Images}
\label{sec:matching}
Since the ASJs contain rich geometric structure informations represented by the anisotropic scales, we are going to study the matching method taken full advantage of ASJs. For a pair of junction $\jmath^P$ and $\jmath^Q$ detected from images $I^P$ and $I^Q$, the homography can be estimated by the points set that contain their locations and endpoints for branches, which can be used to compare junctions for correct correspondences. Since there exist $L$-junctions, $Y$-junctions and $X$-junctions in an image and the type of a junction might be different across images because of occlusion, the homography estimated from a pair of junctions might be invalid. Fortunately, whatever the type of junction is, the location can be intersected from any two of branches that are not parallel each other, which is saying that a junction with more than two branches can be decomposed two several $L$-junctions. Without saying, the $L$-junction with two branches that their orientation $\theta^1$ and $\theta^2$ are equal up to $\pi$ should be filtered out. After decomposing and filtering, the detected in an image are all $L$-junctions. 

The perspective effects are typically small on a local patch~\cite{MikolajczykS05}, which can be approximated by affine homography. We use a pair of $L$-junctions to estimate such homographies. Suppose there are $N^P$ and $N^Q$ decomposed $L$-junctions in image $I^P$ and $I^Q$, denoted as $\{\jmath_i^P\}_{i=1}^{N^P}$ and $\{\jmath_i^Q\}_{i=1}^{N^Q}$ respectively. If a pair of junctions $(\jmath_n^{P},\jmath_m^Q)$ are matched, an affine homography would be induced once the orientations are determined. In order to derive a unique affine homography, we define the partial order for two branches $(r^1,\theta^1)$ and $(r^2,\theta^2)$ of a $L$-junction as
\begin{equation}
\left\{
\begin{split}
(r^1,\theta^1) < (r^2,\theta^2), ~\mathit{if} \langle \theta^1,\theta^2\rangle <\pi\\
(r^1,\theta^1) < (r^2,\theta^2),
~\mathit{if} \langle \theta^1,\theta^2\rangle >\pi
\end{split}
\right. .
\end{equation}
Every junction need to be sorted by the order defined above. The affine homography for a pair of junction $\jmath_n^P$ and $\jmath_m^Q$ are estimated by using DLT (Direct Linear Transform) with their locations and endpoints for the branches. More precisely, we solve the equations
\begin{equation}
\begin{split}
\bm{q}_i = H\bm{p}_i, ~ i=0,1,2\\
\end{split},
\end{equation}
where $\bm{p}_i$ and $\bm{q}_i$ are  the homogeneous representation of locations and two branches for $\jmath_n^P$ and $\jmath_m^Q$ respectively. The matrix $H$ is
$$
H = \begin{bmatrix}
h_1 & h_2 &h_3\\h_4 & h_5 & h_6\\0&0&1
\end{bmatrix}
$$
represents the affine transform induced by $\bm{p}_i$ and $\bm{q}_i$ for $i=0,1,2$.

From the image pair $(I^P,I^Q)$, there can be $N^P\times N^Q$ affine homographies, denoted by $H_{n,m}$, which maps the $n$-th junction in $I^P$ to $m$-th junction in $I^Q$. For correct correspondence , the matrix $H_{n,m}$ will map the image $I^P$ to $I^Q$ accurate around the location of junctions while the mismatch will map the image $I^P$ only correct at the endpoints and locations but erroneous at other positions. For the sake of saving computational resource, we just map a patch $\mathcal{P}(\jmath_n^P)$ around $\jmath_n^P$ to $\mathcal{P}(\jmath_n^Q)$ in $I^Q$ and map $\mathcal{P}(\jmath_m^Q)$ to $\mathcal{P}(\jmath_m^P)$ in $I^P$ by using matrix $H_{n,m}$ and its inverse $H_{n,m}^{-1}$. Then, the distance between two features $\jmath_n^P$ and $\jmath_m^Q$ are measured by
\begin{equation}
\mathcal{D}(\jmath_n^P, \jmath_m^Q)
=
\tilde{D}(\mathcal{P}(\jmath_n^P),
\mathcal{P}(\jmath_n^Q))
+
\tilde{D}(\mathcal{P}(\jmath_m^Q),
\mathcal{P}(\jmath_m^P))
\end{equation}
where the distance $\tilde{D}$ are the distance between two patches calculated by raw patches, SIFT descriptor or other descriptors.

Benefiting with the homographies induced by ASJ, the distance between original patches and mapped patches for correct correspondence is usually very small while larger for incorrect correspondence, we can use ratio test proposed in \cite{Lowe04} to filter out the incorrect correspondence.

\section{Experimental Analysis}
\label{sec:exp}
This section illustrates the results and analysis for ASJ detection and matching routines with comparison to existing approaches for junction detection, junction matching, key-points matching and line segment corresponding. In our experiments, we first detect anisotropic-scaled junctions by relying on
the procedures presented in Section \ref{sec:detection}, and then make the correspondence of
junctions with the affine homography induced by these semi-local geometrical structures.
\vspace{-1mm}
\subsection{Stability and Control of the Number of False Detection}
The \emph{a-contrario} approaches detect meaningful events controlled by the threshold $\epsilon$: it bounds the average number of false detections in an image following null hypothesis. In this subsection, we check the average number of false detections in Gaussian noise image and illustrate the results of detected ASJs with fixed threshold $\epsilon$.

Experimentally, we generate $1000$ random images with $256\times 256$ pixels which follow standard Gaussian distribution independently pixel-wised. For each pixel, we generate an orientation randomly from the uniform distribution in the interval $[0,2\pi)$ and estimate scale at this pixel with the orientation. Ideally, there is no meaningful line-segment structure appeared in random images but might be detected mistakenly, which are counted in number of false detection averagely. If the number of false detection can be controlled by the \emph{NFA} proposed \emph{a-contrario approach}, the approach would be identified as correct \emph{a-contrario approach}.
\begin{figure*}[htb!]
	\centering
	\subfigure[original images]{
		\includegraphics[height=0.2\linewidth]{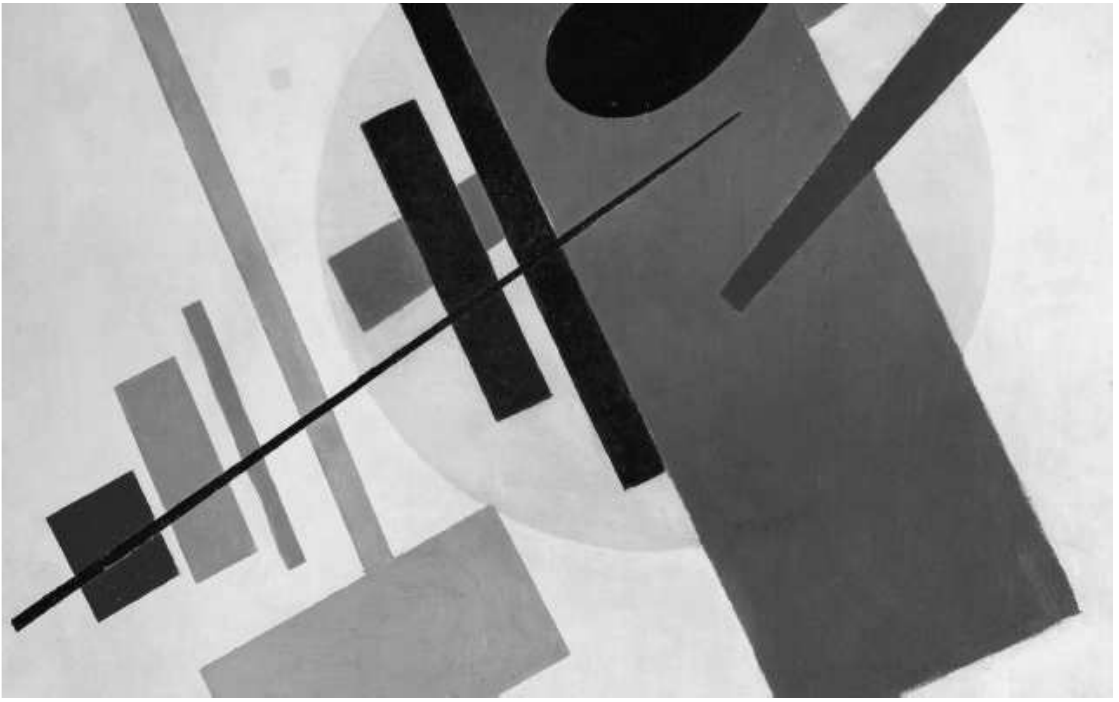}
		\includegraphics[height=0.2\linewidth]{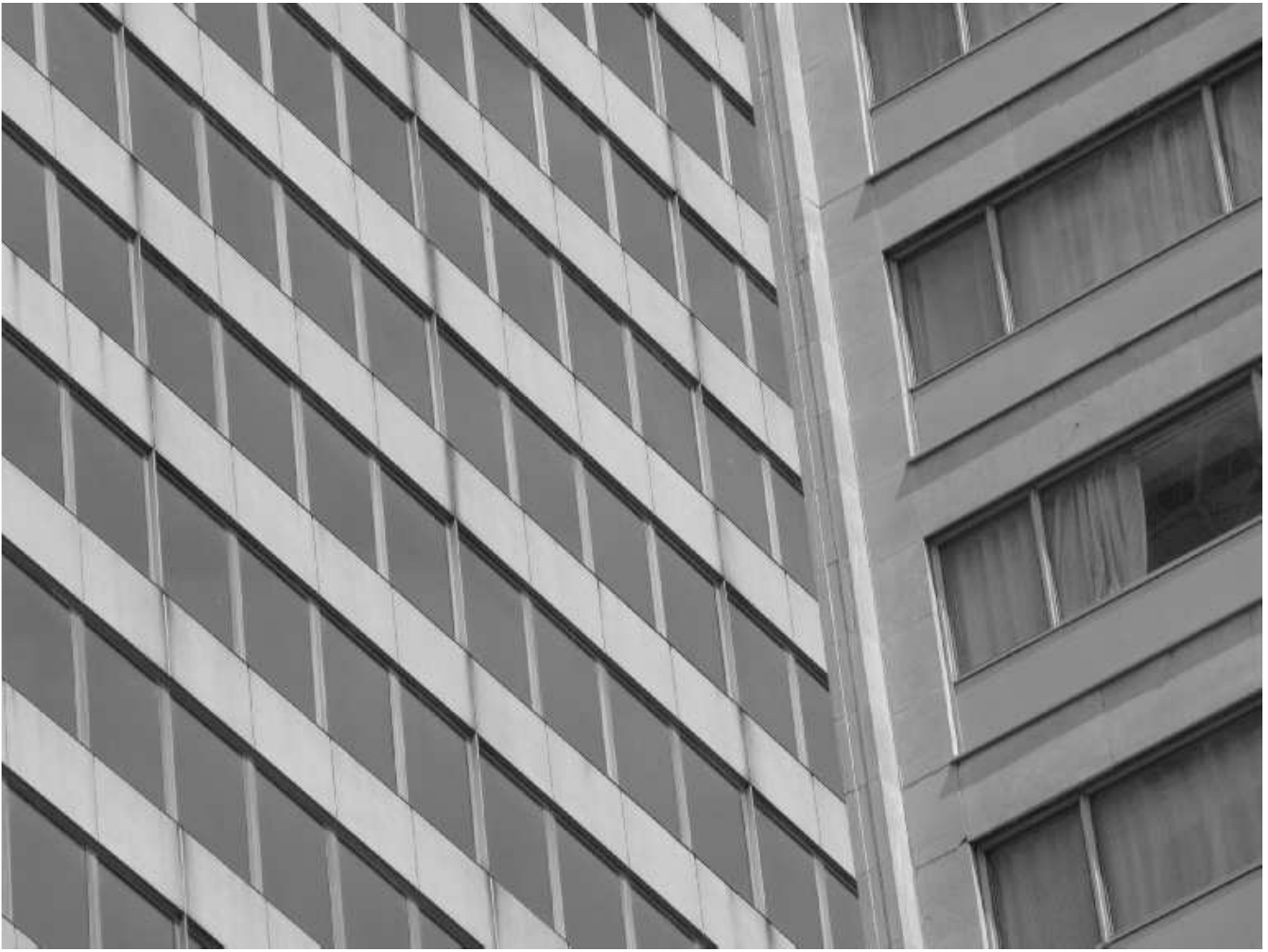}
		\includegraphics[height=0.2\linewidth]{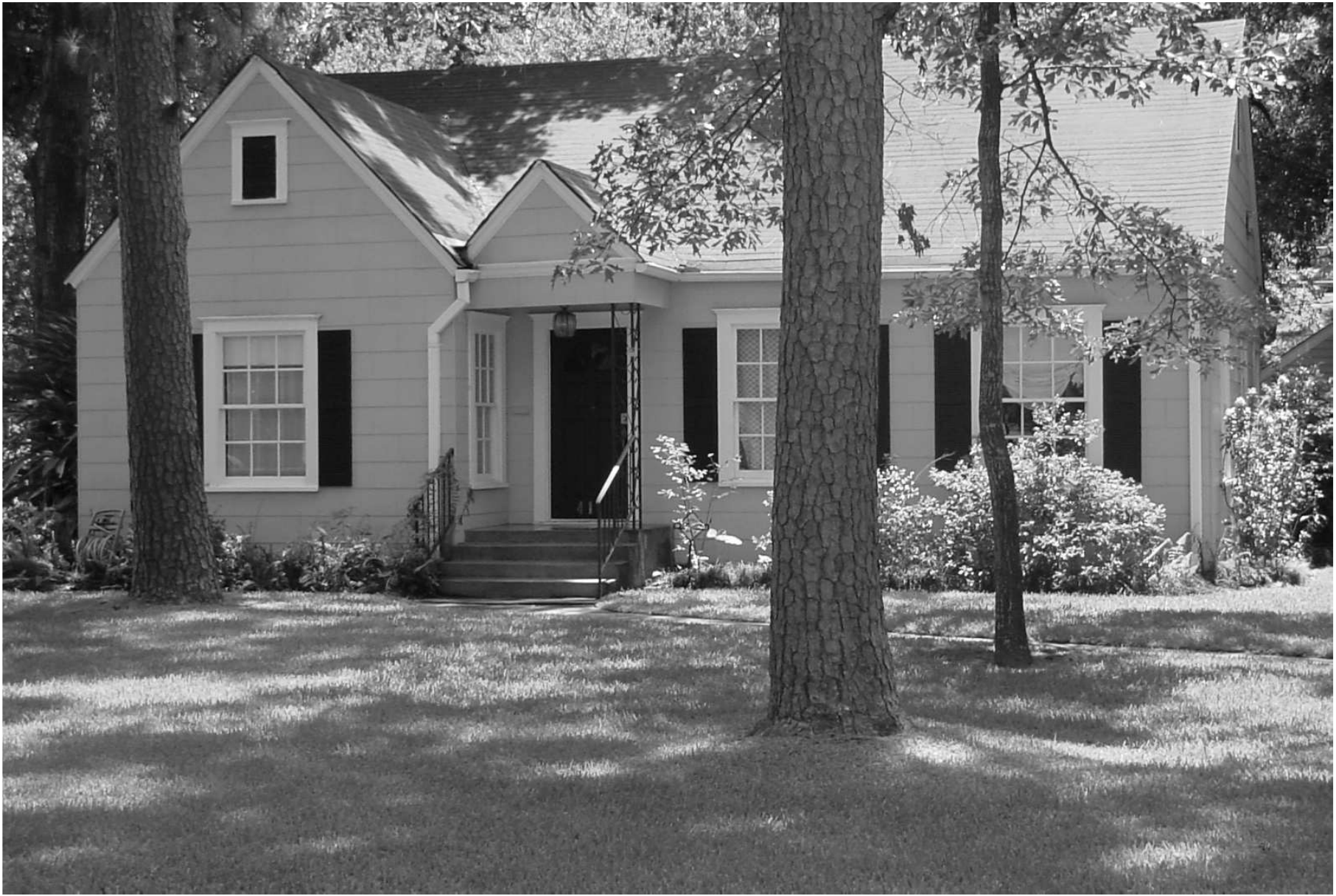}
	}
	\subfigure[repeatability rate with respect to scale changes]{
		\includegraphics[height=0.24\linewidth]{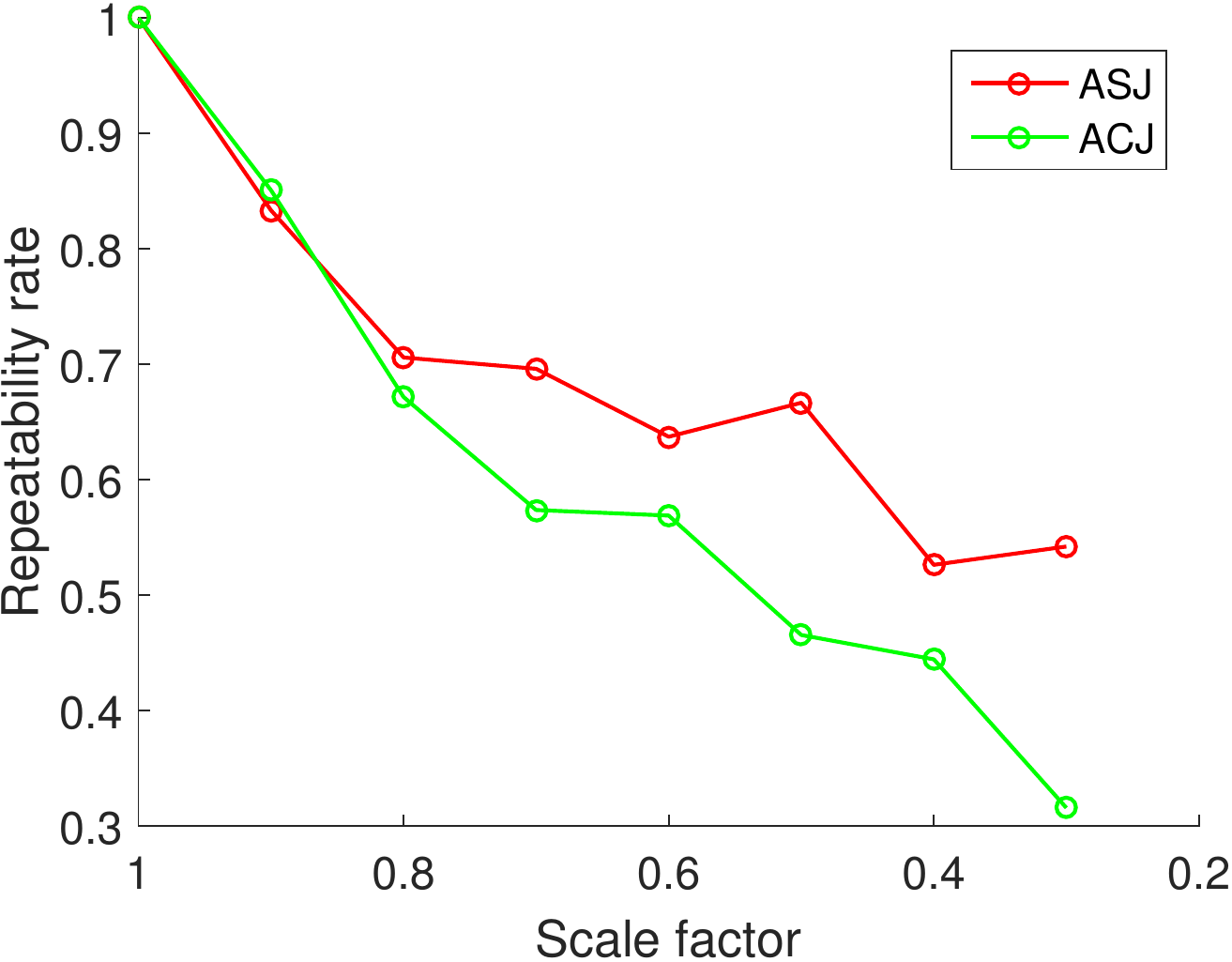}
		\includegraphics[height=0.24\linewidth]{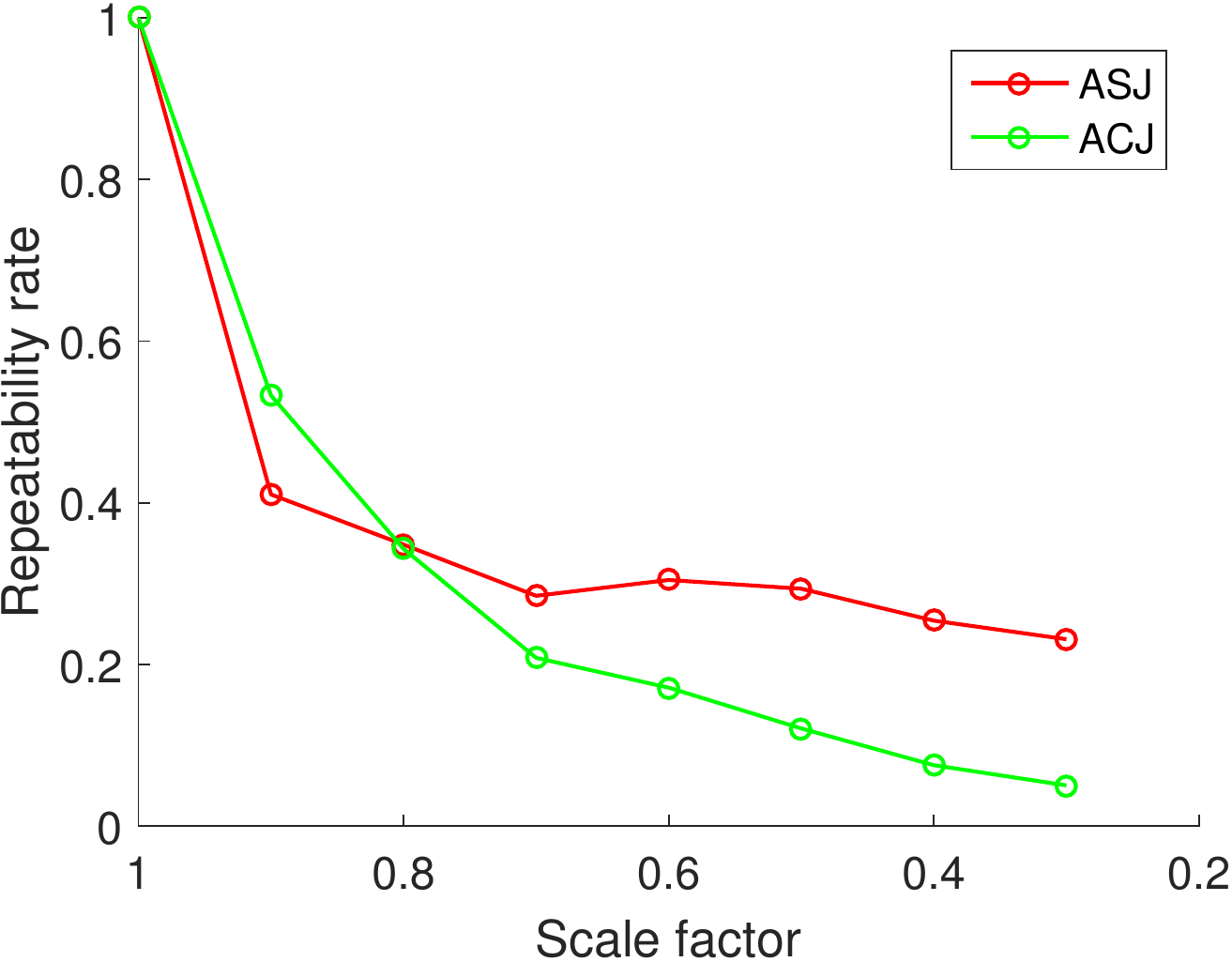}
		\includegraphics[height=0.24\linewidth]{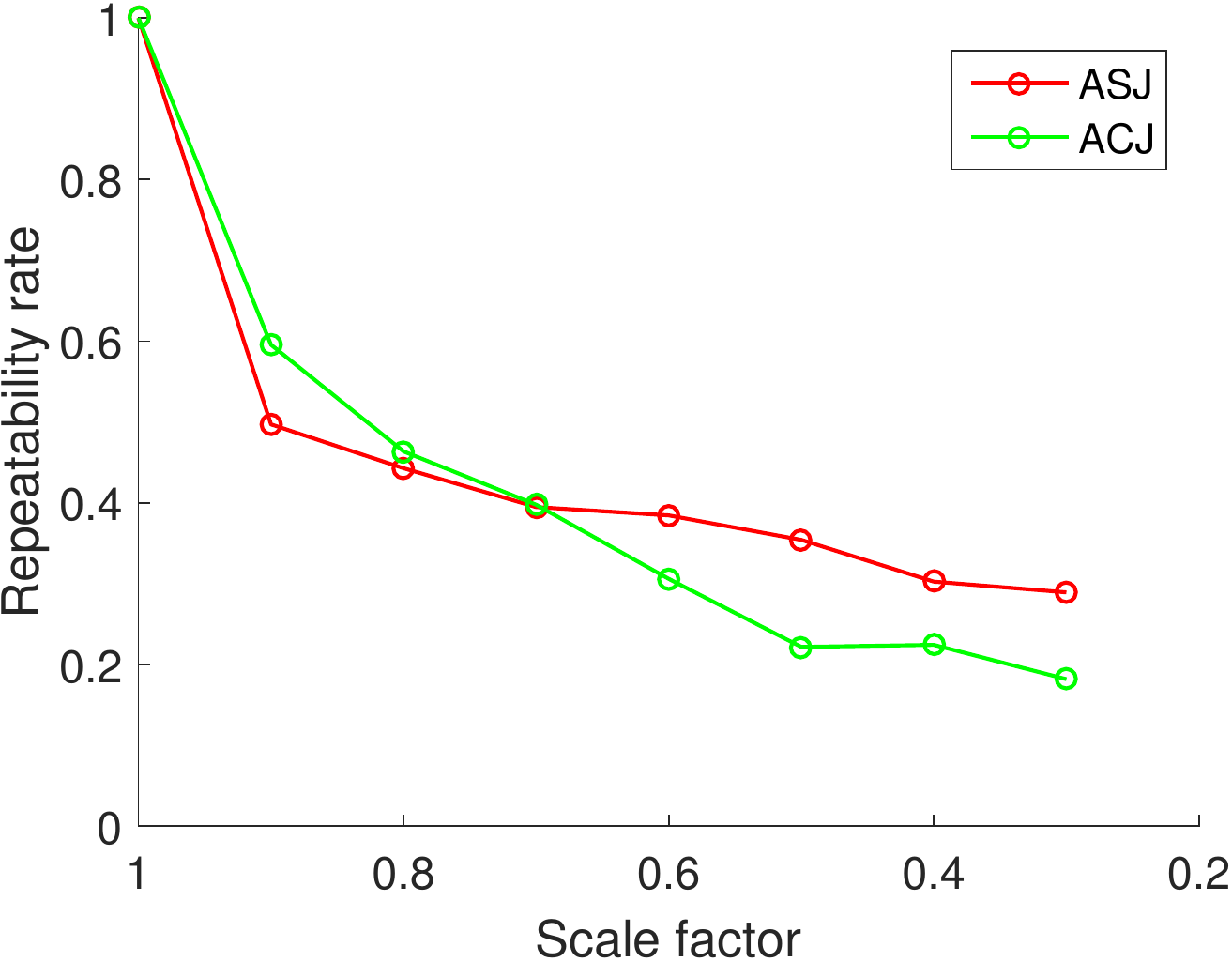}
	}
	\caption{Repeatability rate with respect to scale change. Original images to generate image sequences are shown in  the first row. In the second row, the repeatability is shown as a function of scale factors.}
	\label{fig:rep-exp}
\end{figure*}
\begin{table}[htb!]
	\centering
	\caption{Average number of false detections in $1000$ images generated by Gaussian white noises}
	\scriptsize
	\begin{tabular}{c|cccccc}
	\hline
	$\epsilon$ & 0.01 & 0.1 & 1 & 10 & 100 & 200\\
	\hline
	\hline
	Avg.
	False  & 0.002 &	0.006 &	0.198 &	5.923 &	66.472 & 132.676
	\\
	\hline
	\end{tabular}
	\label{tab:nfa-gaussian}
\end{table}
The average number of false detections in $1000$ Gaussian noise images are reported in the Tab.~\ref{tab:nfa-gaussian}. The value of NFA are varied in our experiments from $10^{-2}$ to $10^2$ and the corresponding average number of false detections are upper bounded by the NFA.

\subsection{Comparison with ACJ}
It is necessary to compare the repeatability for our proposed ASJ with ACJ since we extend the \emph{a-contrario} model for scale estimation to discuss their difference. 
Following with the baseline experiments proposed in~\cite{XiaDG14}, these images are firstly zoomed with different factors to form the image sequences with scale change. Then, the ASJ and ACJ are performed on these image sequences to detect the junctions. The repeatability for ACJ is discussed in~\cite{XiaDG14}, however, their definition for corresponding junction just concentrates on the location and branch of junctions while ignoring the scale coherence. Therefore, we are going to define the corresponding ACJ and ASJ with scale information here. For the original image $I_0$ and the scaled image $I_s = s(I_0)$, the corresponding ACJ junctions should have close locations, branch orientations as well as scales. Meanwhile, two junctions with different number of branches cannot be identified as correspondence. More precisely, we define two ACJ junctions  $\jmath_1 = \{\bm{p}_1,r_1,\{\theta_i\}_{i=1}^M\}$ and $\jmath_2 = \{\bm{p}_2,r_2,\{\theta_i^s\}_{i=1}^M\}$ detected in $I_0$ and $I_s$ if they follow 
\begin{equation}
\left\|s\cdot \bm{p}_1-\bm{p}_2\right\|_2 < 3,
\label{eq:corresponding-loc}
\end{equation}
\begin{equation}
\left|s\cdot r_1-r_2\right|<3,
\label{eq:corresponding-iso-scale}
\end{equation}
\begin{equation}
\max_{\theta\in\{\theta_i\}_{i=1}^M}\min_{\theta'\in\{\theta_i\}_{i=1}^M} d_{2\pi}(\theta,\theta') < \frac{\pi}{20},
\label{eq:corresponding-ang}
\end{equation}
where the angular distance $d_{2\pi}(\theta,\theta') = \min(\left|\theta-\theta'\right|,2\pi-\left|\theta-\theta'\right|)$.
Similar to the above, the correspondence for two junctions $\jmath_1 = \{\bm{p}_1,\{r_i,\theta_i\}_{i=1}^M\}$ and $\jmath_2 = \{\bm{p}_2,\{r_i^s,\theta_i\}_{i=1}^M\}$ detected by ASJ should satisfy the inequalities \eqref{eq:corresponding-loc}, \eqref{eq:corresponding-ang} as well as 
\begin{equation}
\max_{r\in\{r_i\}_{i=1}^M}\min_{r'\in\{r_i^s\}_{i=1}^M}  \left|s\cdot r-r'\right| < 3.
\end{equation}
In this experiment, the set of scale factors is $\left\{1.0, 0.9, 0.8, \ldots, 0.3\right\}$ and the results are shown in Fig.~\ref{fig:rep-exp}. Observing the repeatability curve, our proposed ASJ performs better than ACJ. The repeatability rate reported in \cite{XiaDG14} is higher, however, it just demonstrate the accuracy of locations and orientation of branches. In our experiment, the scale difference are also considered here. 

As reported in \cite{XiaDG14}, the scale of ACJ represents the length of shortest branch and it is roughly linear through the scale factors\cite{XiaDG14}. Theoretically, if a detected ACJ has scale $r$ in original image, its correspondence in the scaled image $s(I)$ should be close enough to $s\cdot r$. However, the upper bound of scale is required for ACJ algorithm as input and it is recommend to be set as in the range of $[12,30]$ constantly\cite{XiaDG14} for the sake of computational speed. As a matter of fact, the junctions in indoor images usually have large scale branches and they cannot be bound with a relative small constant. 

\begin{figure*}[htb!]
	\centering
	\includegraphics[height=0.13\textheight]{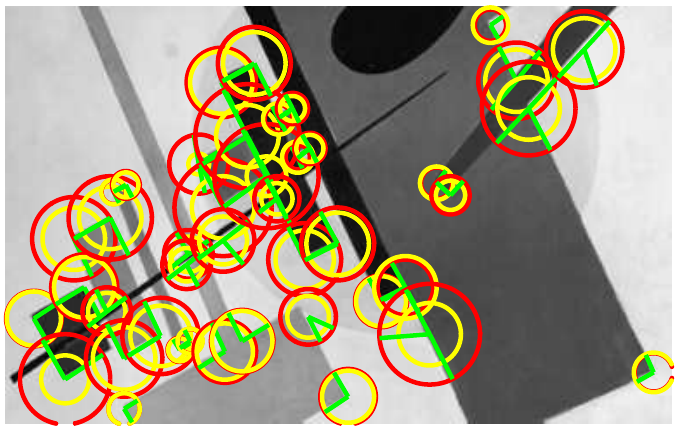}	
	\includegraphics[height=0.13\textheight]{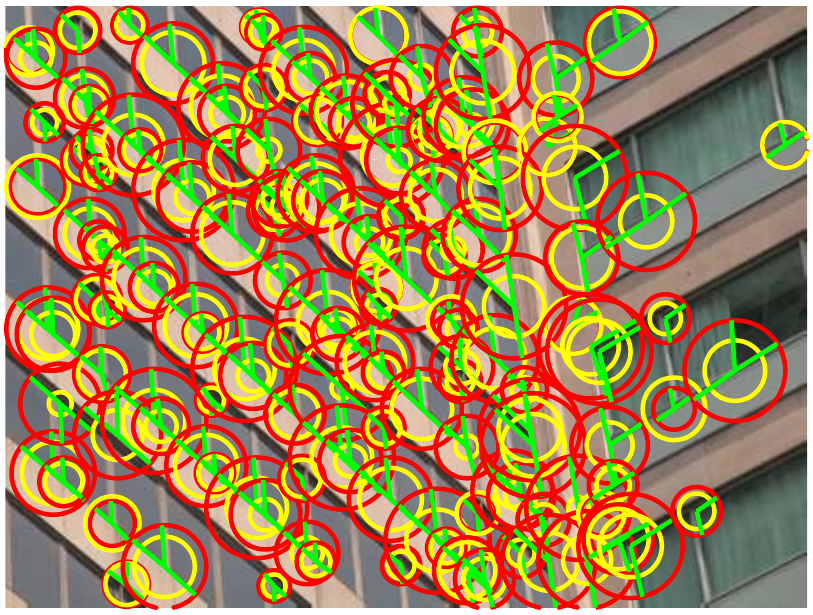}	
	\includegraphics[height=0.13\textheight]{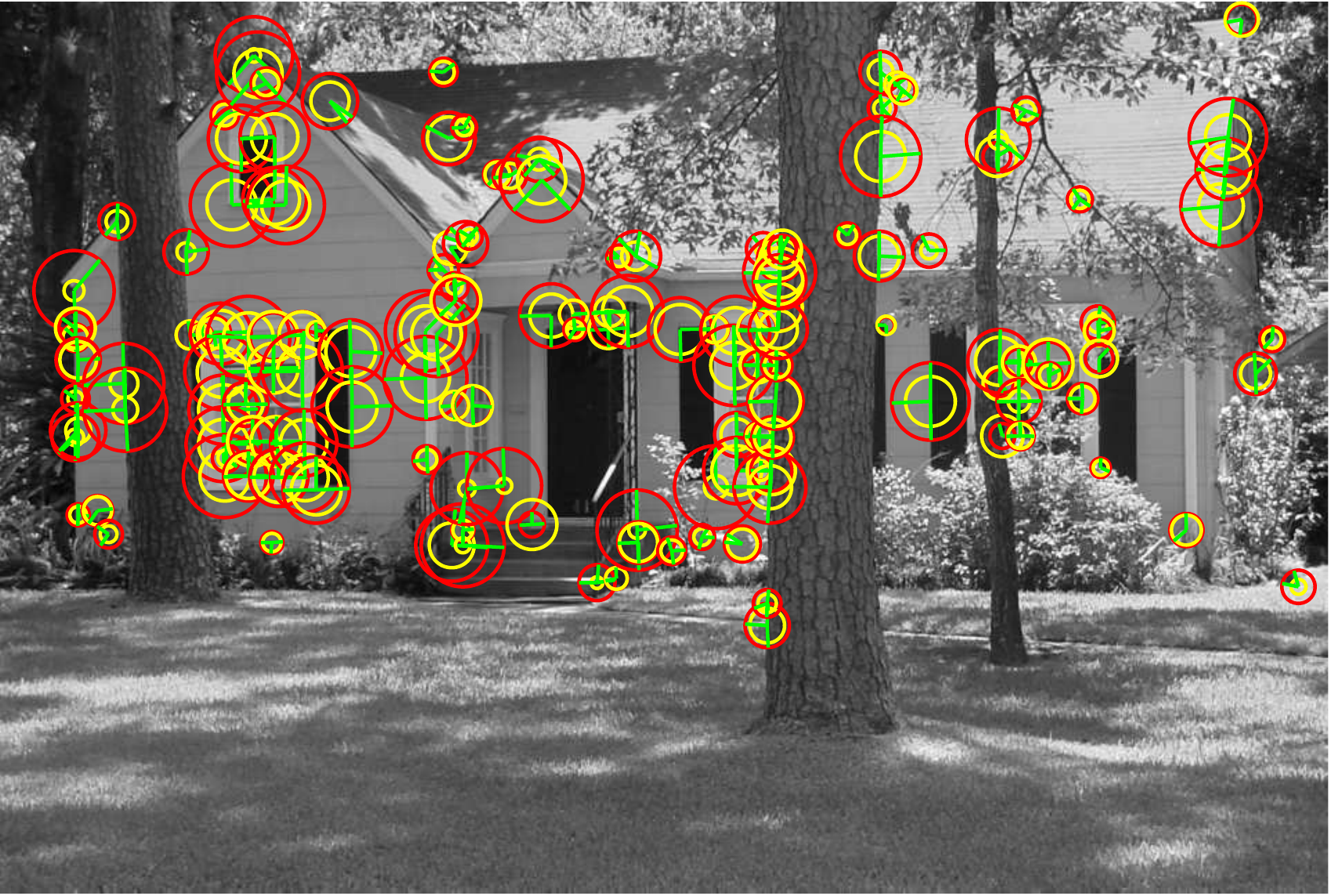}			
	\includegraphics[height=0.13\textheight]{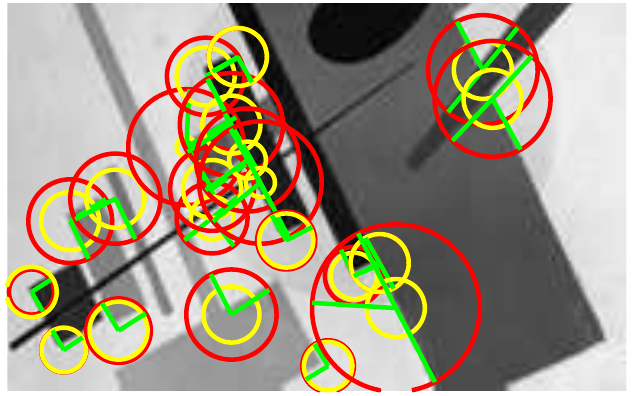}	
	\includegraphics[height=0.13\textheight]{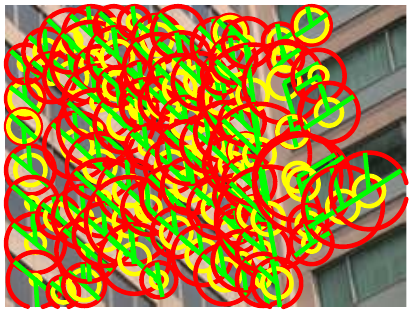}	
	\includegraphics[height=0.13\textheight]{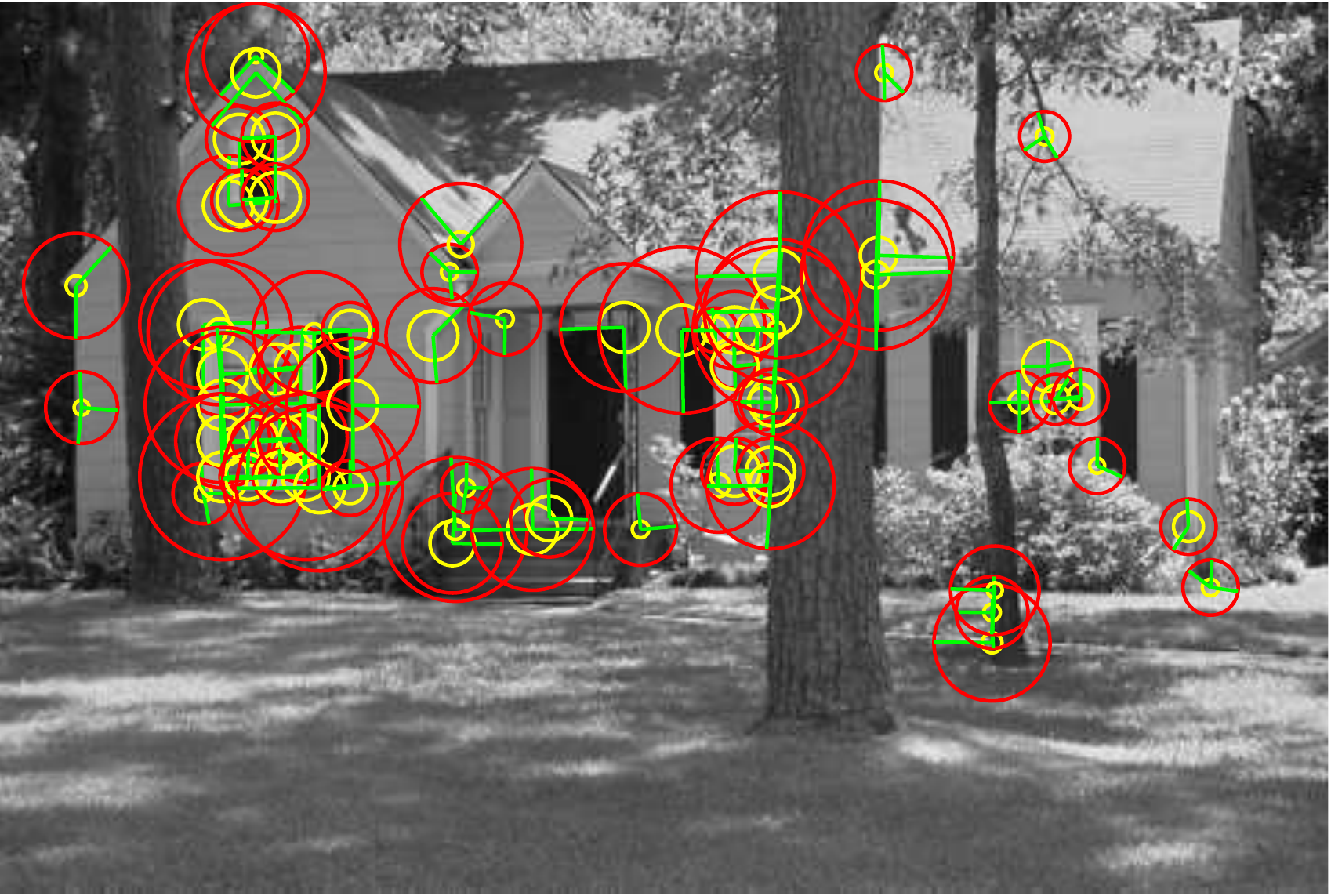}			
	\caption{The scale consistency between the original image and scaled image. The \emph{yellow circles} represent the scale estimated in the scaled image with scale factor $s$ while \emph{red circles} represent the scales detected in original images. The scale factors  are $s = 0.6$ for the top row and $s=0.3$ for the bottom.}
	\label{fig:scale-inaccurate}
\end{figure*}

To demonstrate this fact, we compare the detected junctions in Fig.~\ref{fig:scale-inaccurate}. In this experiment, the junctions are detected by ACJ firstly in original image $I$ and scaled image $s(I)$ with the factor $s$ firstly. Then we find the corresponding ACJ in the image pair by using the inequalities \eqref{eq:corresponding-ang} and \eqref{eq:corresponding-loc} while ignoring the inequality \eqref{eq:corresponding-iso-scale}. For the sake of comparing the scale of junctions with respect to factor $s$, all the correspondences are shown with colored circle. 
In Fig.~\ref{fig:scale-inaccurate}, a correspondence of  $\jmath_2 = {\bm{p}_2,r_2,\{\theta_i^s\}_{i=1}^M}$ in image $s(I)$ which has scale $r_1$ is shown with a yellow circle with the radius $r_1\cdot s$. The red circle and green line segments present the junction $\jmath_2$. We can find out that there exist several correspondences which do not have consistent scales. If a junction is formed by several line segments of which lengths are more than $1/s$ time of maximal radius threshold of ACJ, the scale of junction will be equal to the threshold in the original image. When the image is zoomed with factor $s$, the scale will not be decreased since it is still larger than the threshold. This is the reason why the repeatability is lower when we use the inequality \eqref{eq:corresponding-iso-scale} to calculate it. 

\begin{figure*}[htb!]
	\centering
		\includegraphics[width=0.25\textwidth]{figures/cmpimg1}
		\includegraphics[width=0.25\textwidth]{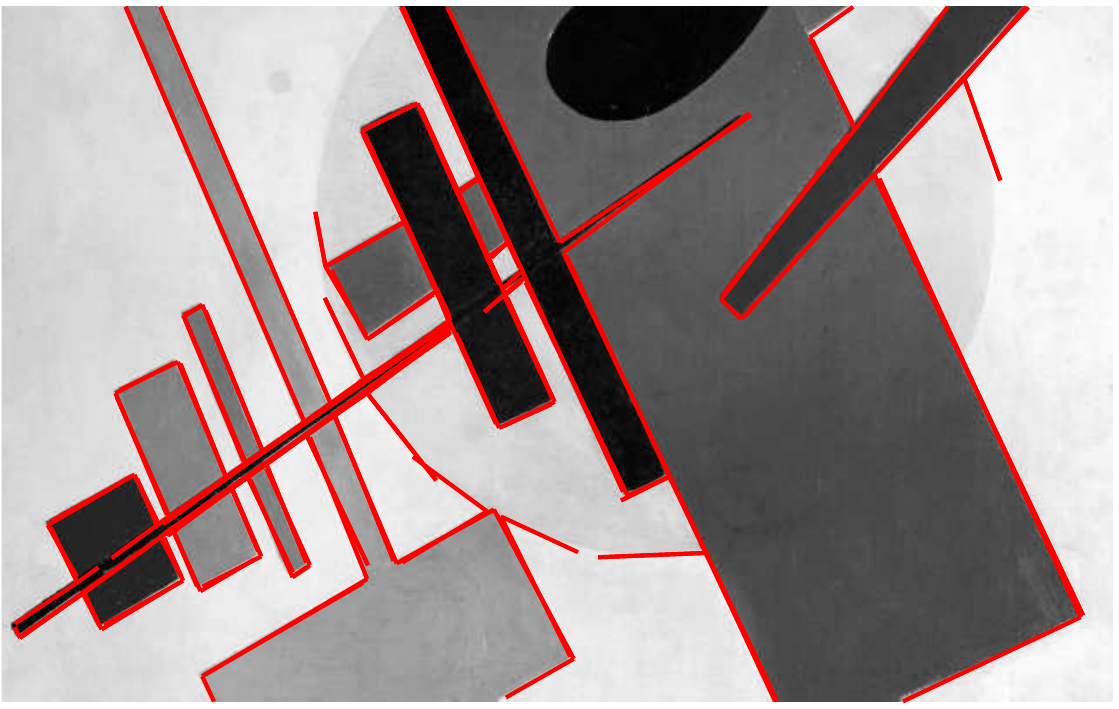}
		\includegraphics[width=0.25\textwidth]{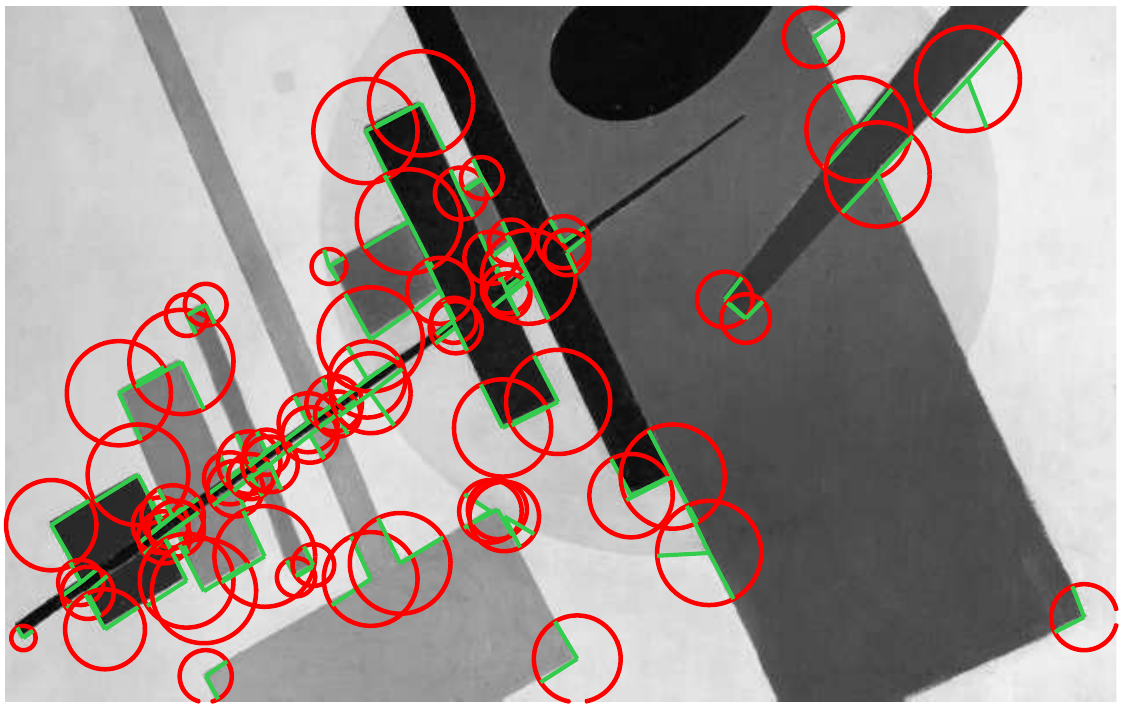}
	\\
	\vspace{1mm}
		\includegraphics[width=0.25\textwidth]{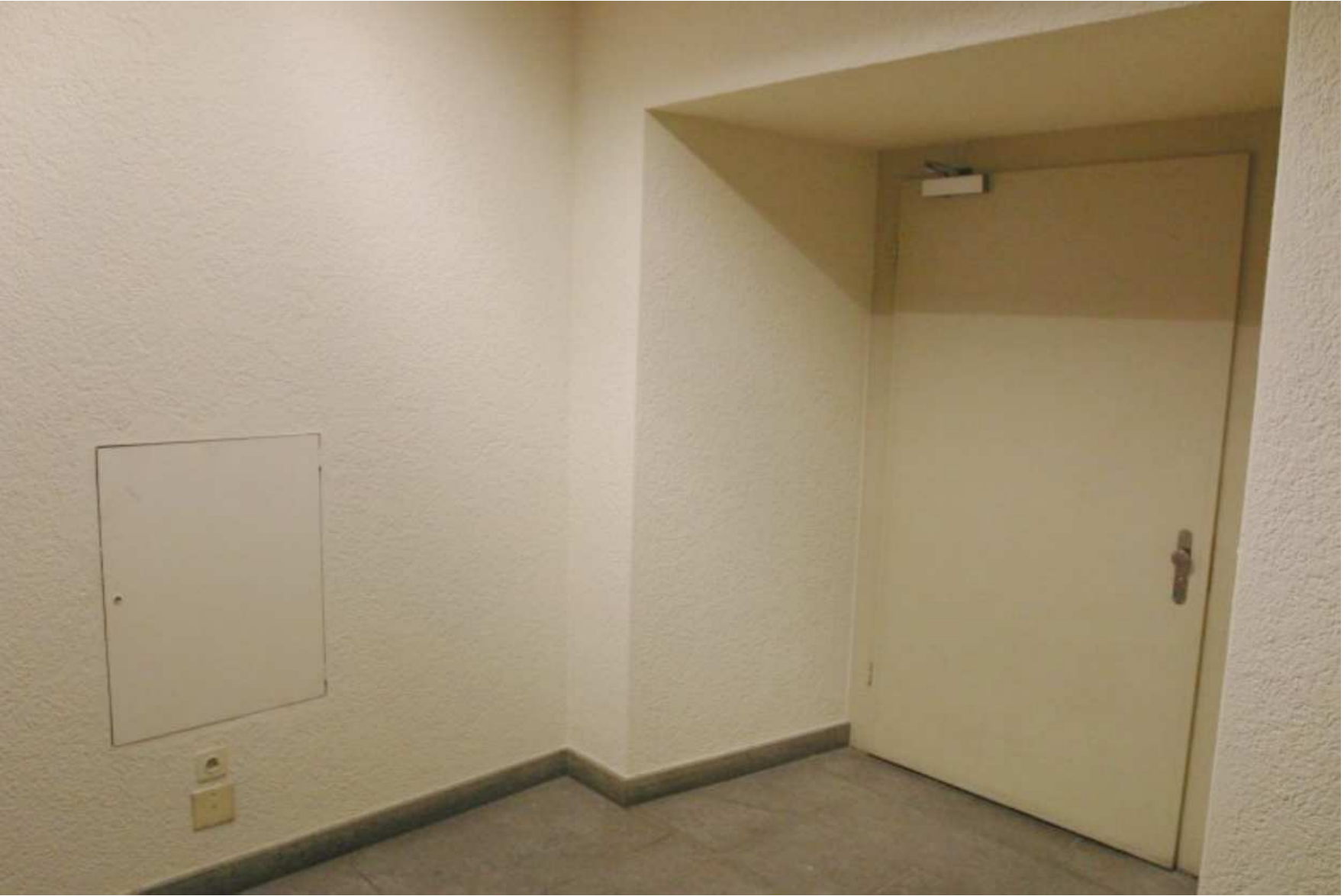}
		\includegraphics[width=0.25\textwidth]{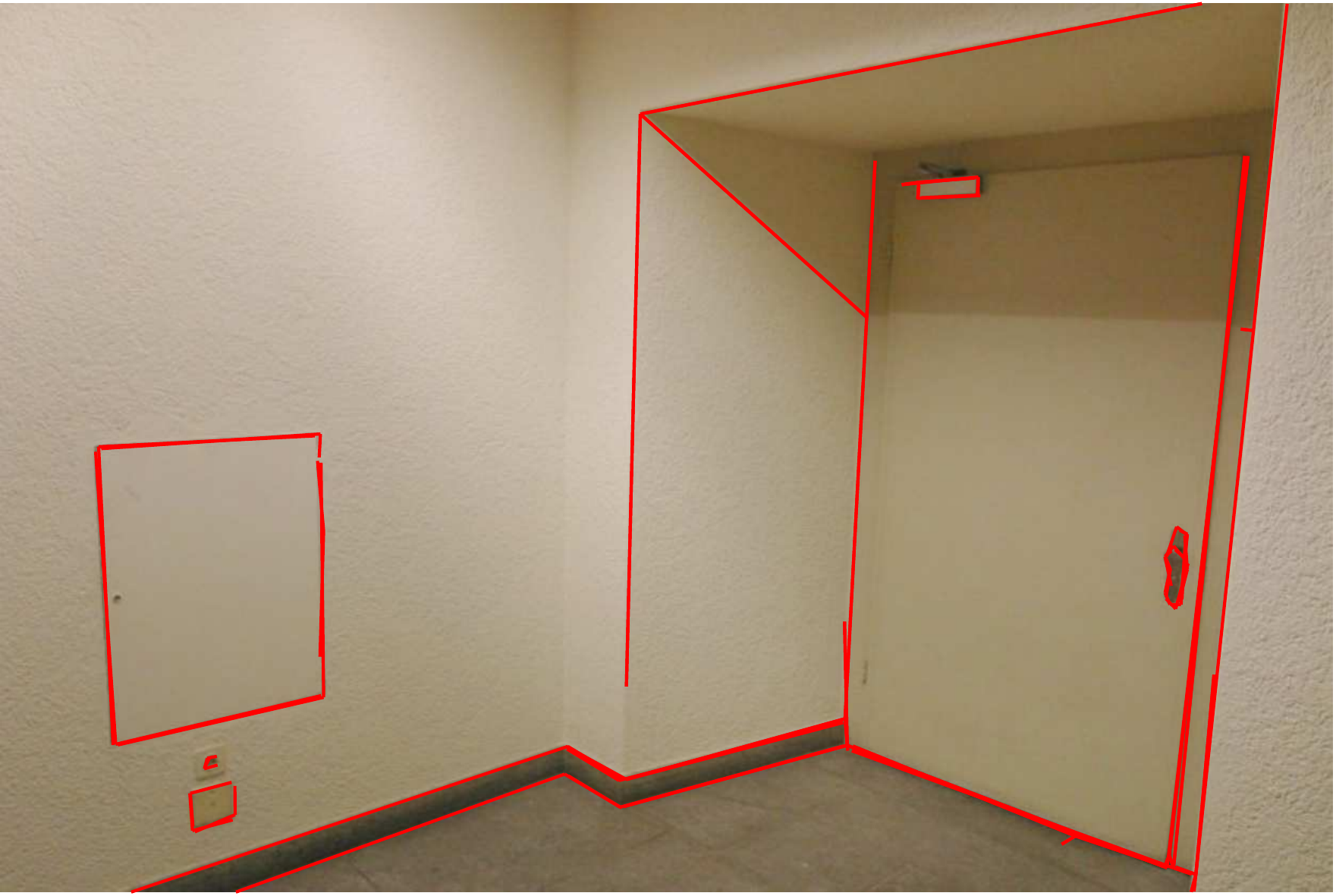}
		\includegraphics[width=0.25\textwidth]{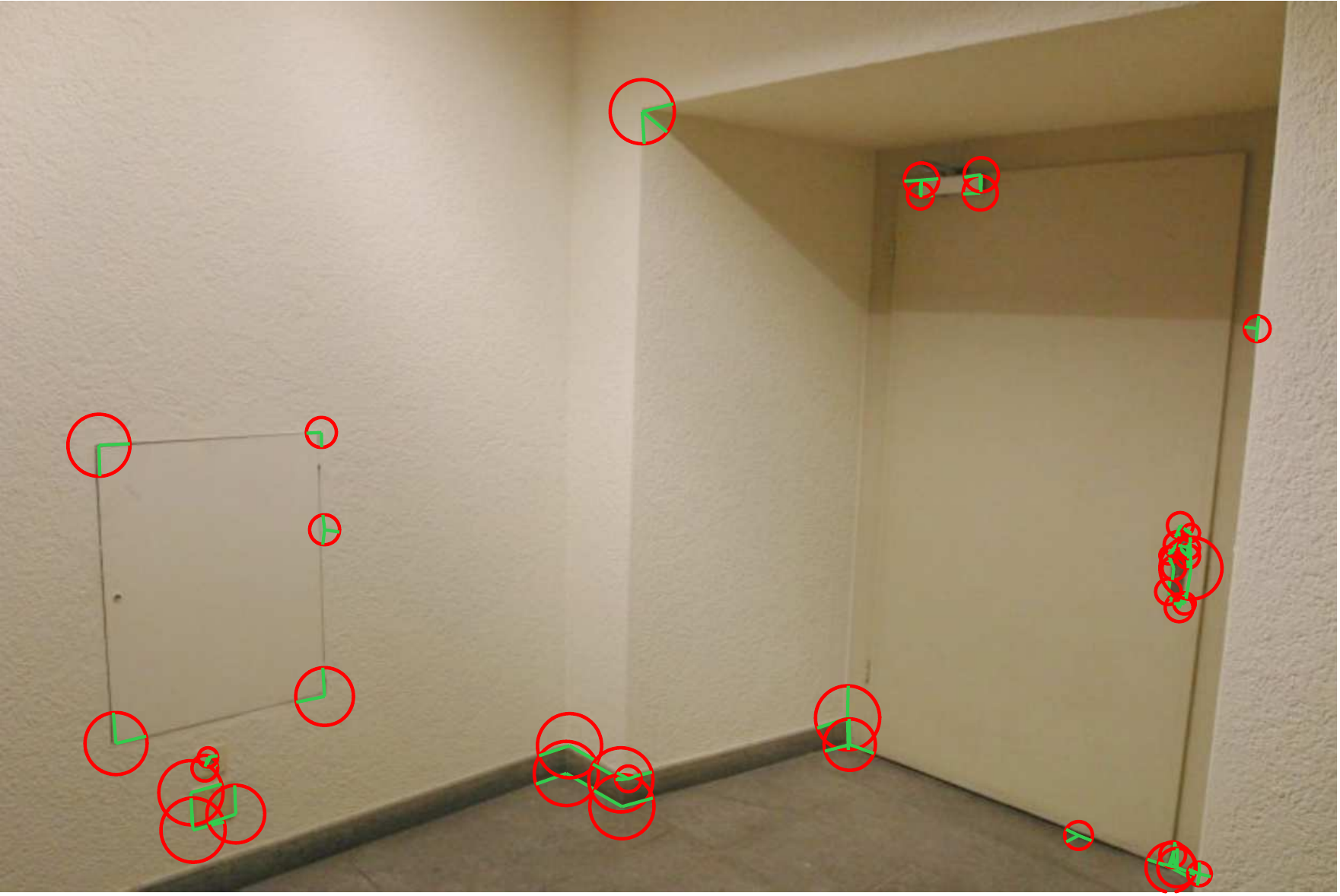}
	\\
	\vspace{1mm}
		\includegraphics[width=0.25\textwidth]{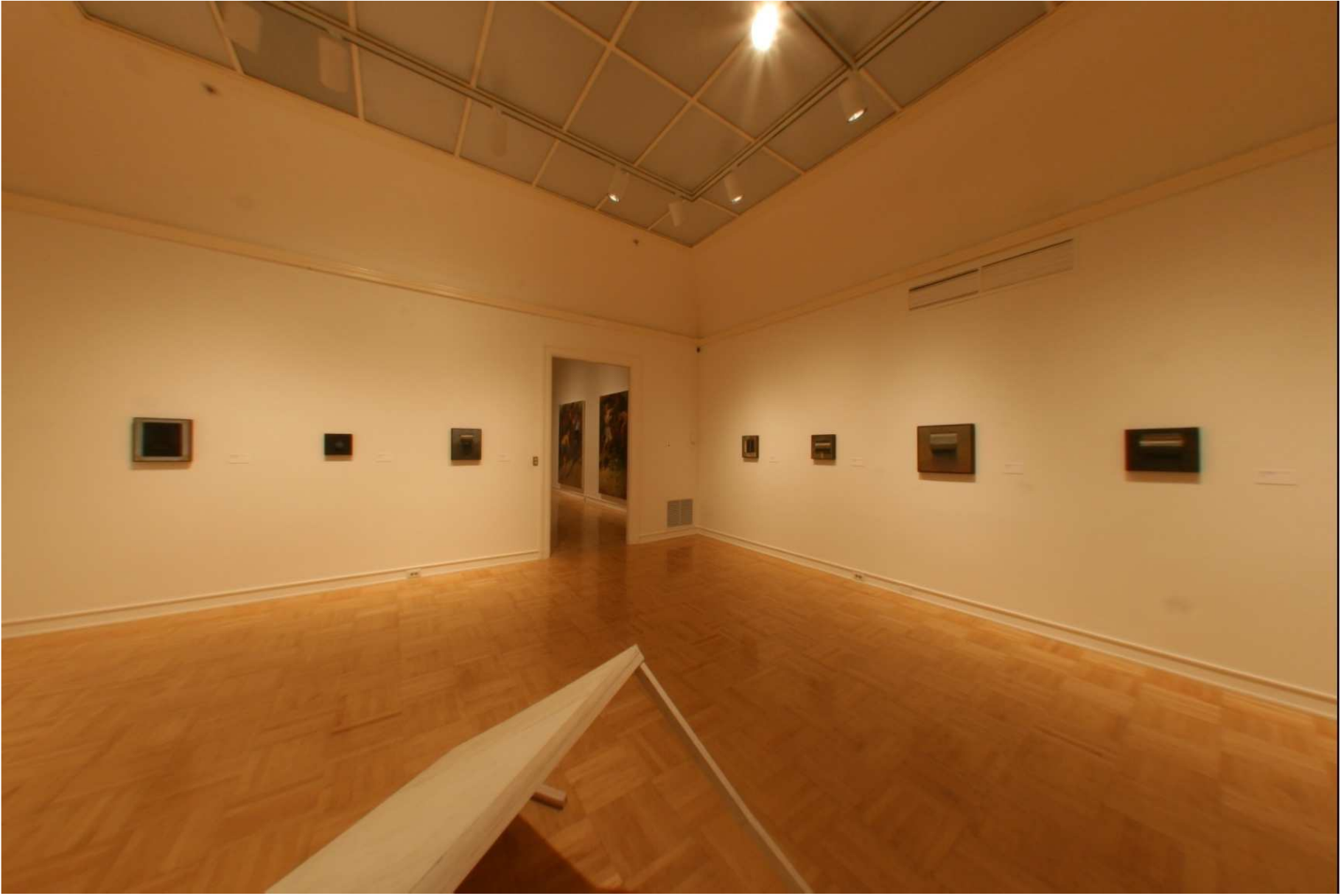}
		\includegraphics[width=0.25\textwidth]{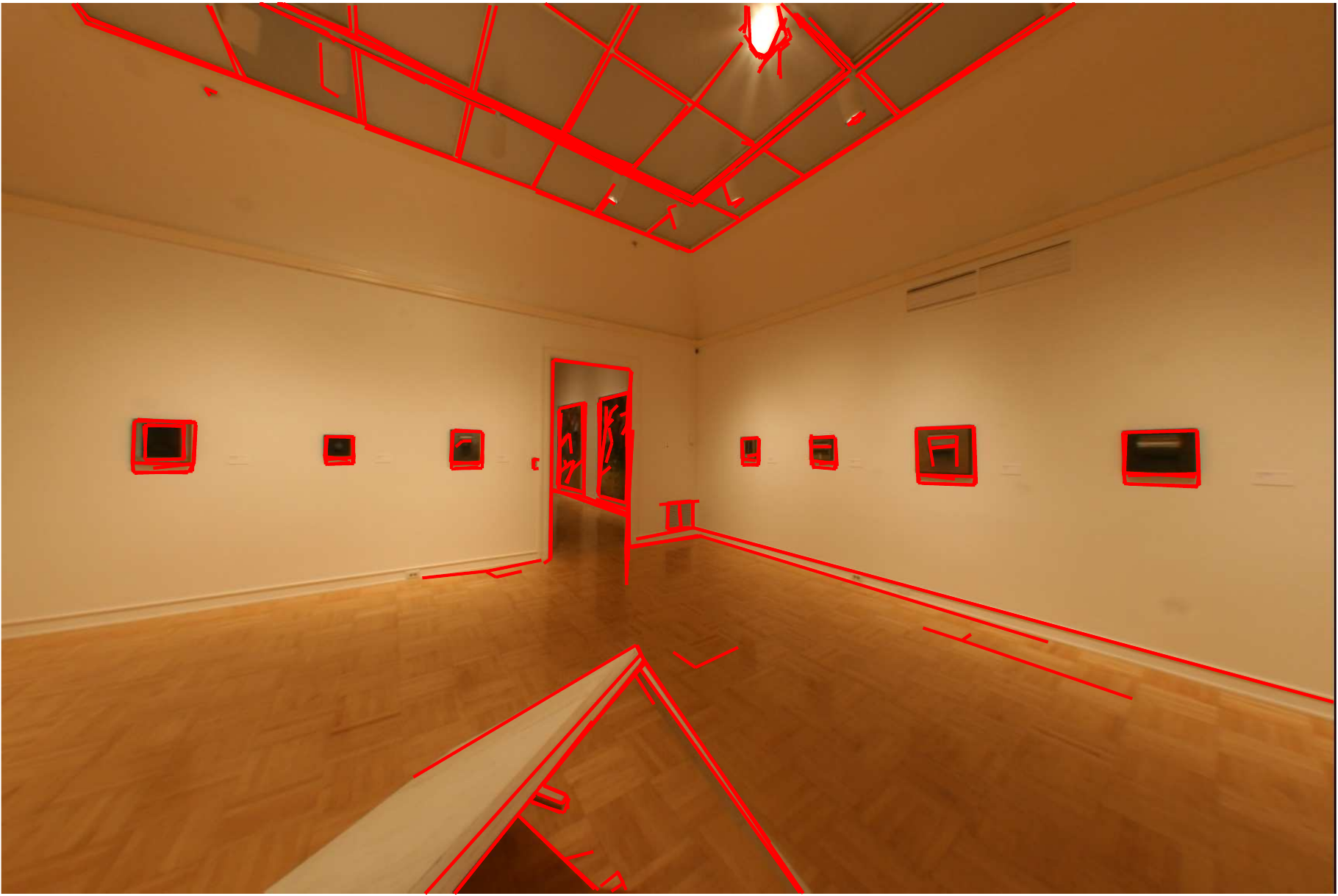}
		\includegraphics[width=0.25\textwidth]{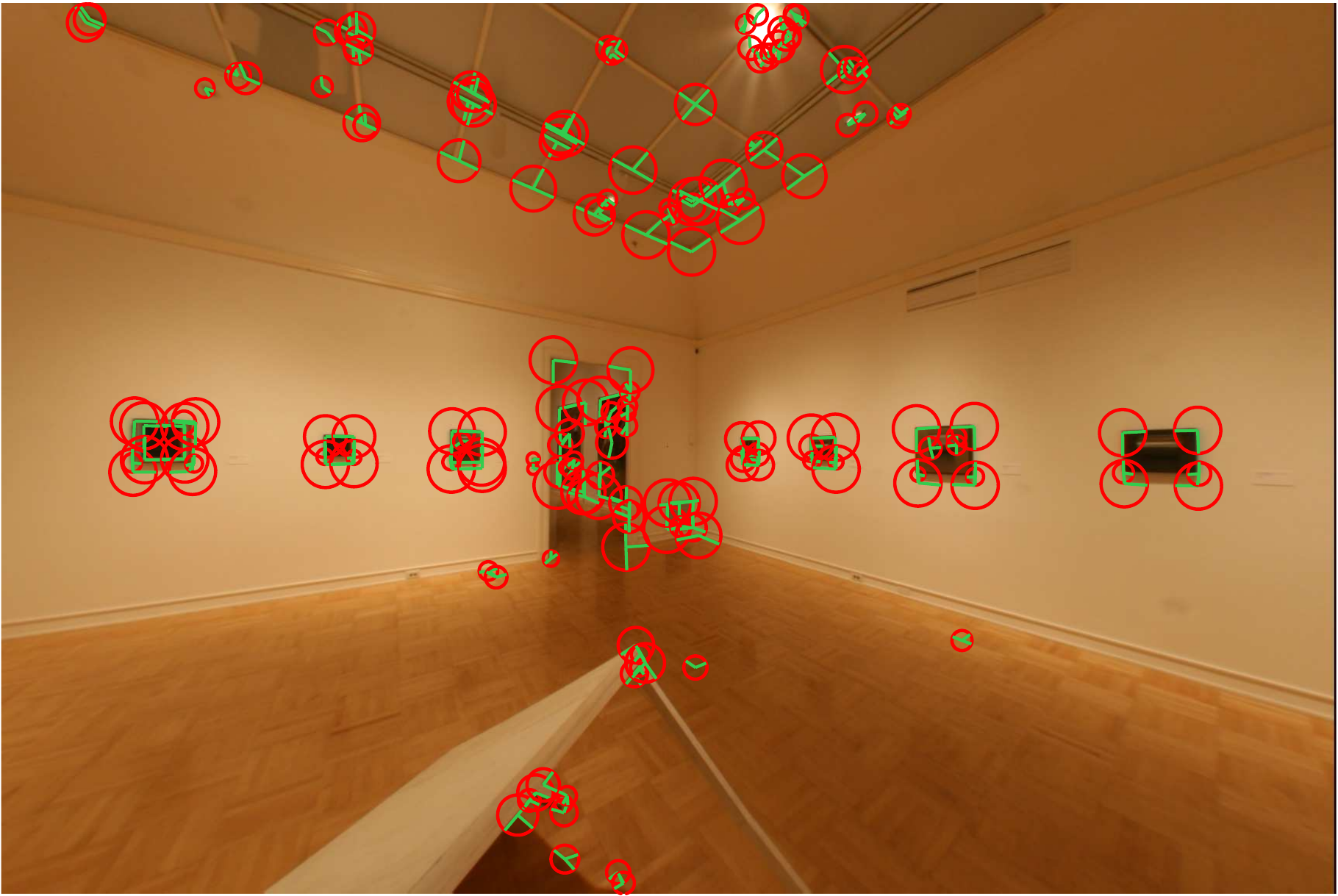}
	\caption{Some results of ASJ for the input images in the first column are shown in the middle column. The junctions detected by ACJ are shown in the right column for comparison.}
	\label{fig:example-asj}
\end{figure*}

In the final of this subsection, some example results of ASJ detector for indoor images are shown in Fig.~\ref{fig:example-asj}.
The \emph{anisotropic-scale} junction are shown in the middle column and the results of ACJ are listed in the right column.
Observing the results, we can find that ASJ has the ability to detect more geometric structure than ACJ. The anisotropic-scale branches of a junction can depict the layout of indoor scenes. By contrast, the results of ACJ just represent the very local information. For example, there are several rectangles in the Fig.~\ref{fig:example-asj}, our ASJ can produce the boundary of the rectangle while ACJ just detect the corner points and orientations around the corners of rectangle.
\subsection{ASJ Matching}
\begin{figure*}[htb!]
	\centering
	\subfigure[]{
		\begin{minipage}{0.13\textwidth}
			\includegraphics[height=0.07\textheight]{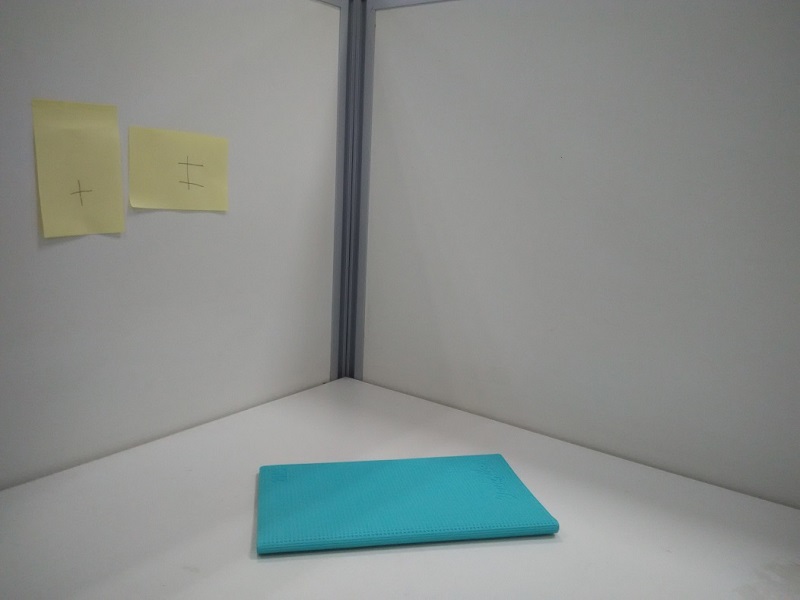}
			\\
			\includegraphics[height=0.07\textheight]{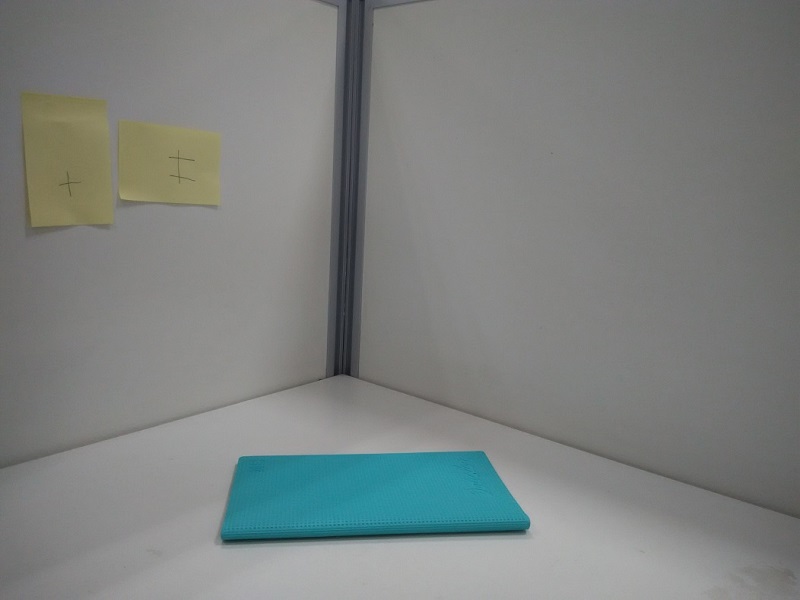}
		\end{minipage}
	}	
	\subfigure[]{
		\begin{minipage}{0.13\textwidth}
			\includegraphics[height=0.07\textheight]{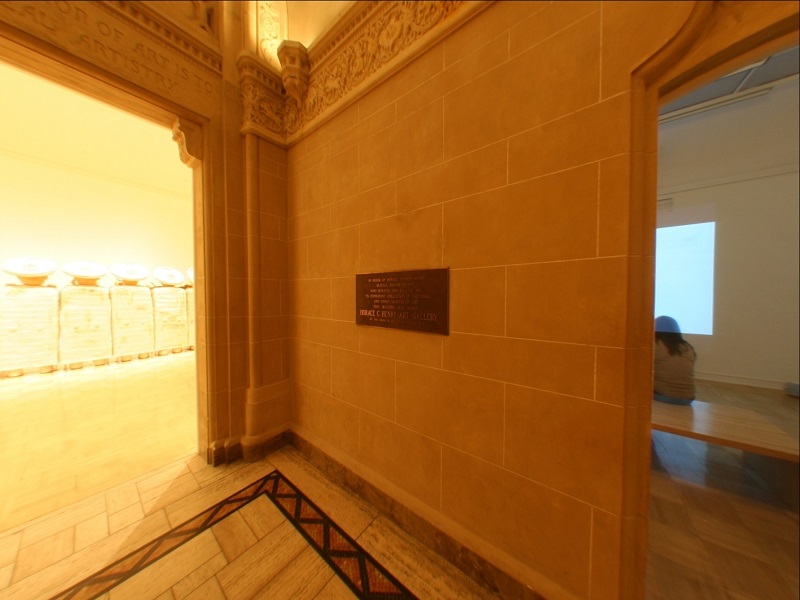}
			\\
			\includegraphics[height=0.07\textheight]{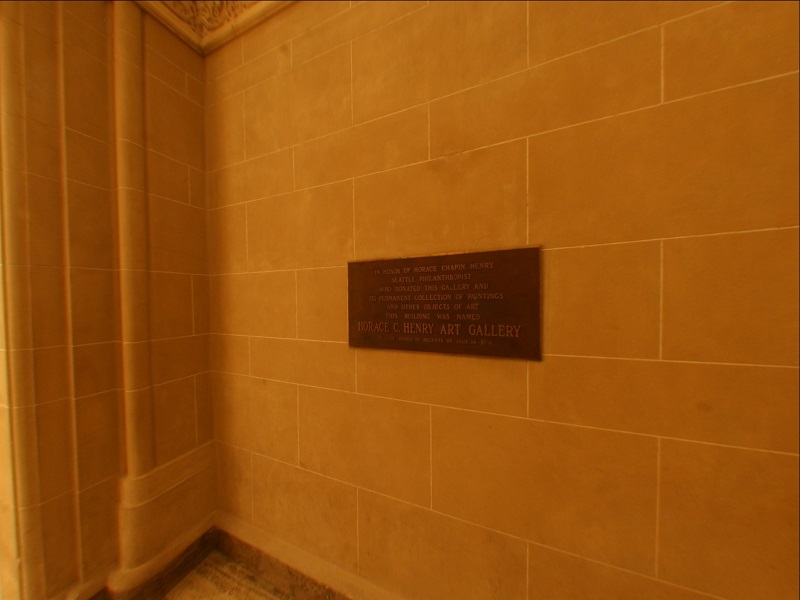}
		\end{minipage}
	}	
	\subfigure[]{
		\begin{minipage}{0.13\textwidth}
			\includegraphics[height=0.07\textheight]{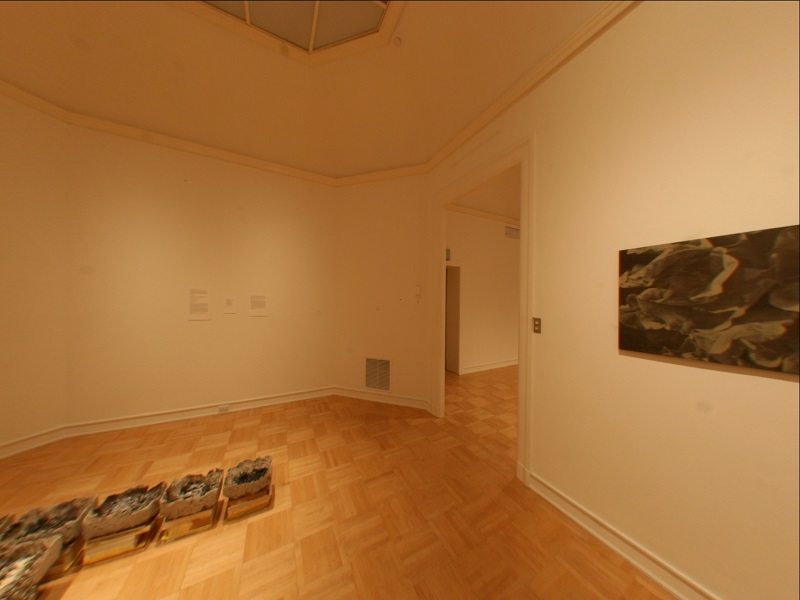}
			\\
			\includegraphics[height=0.07\textheight]{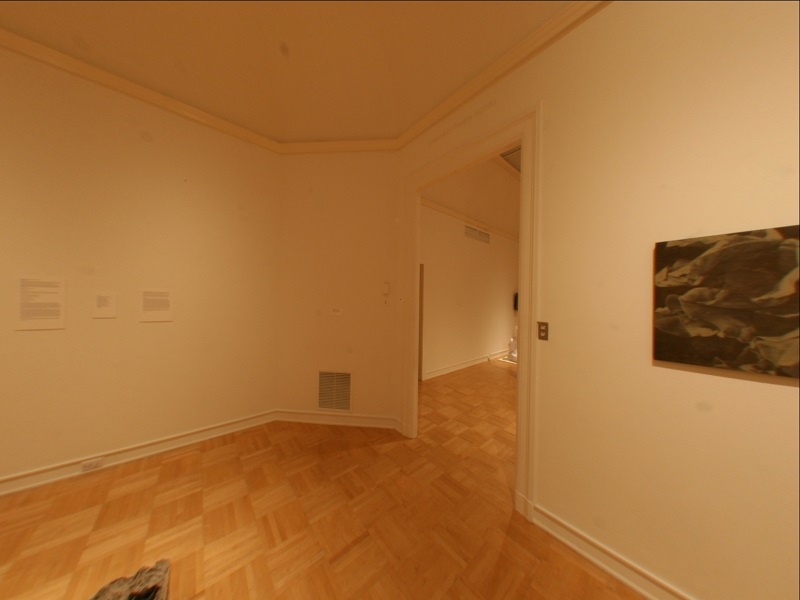}
		\end{minipage}
	}	
	\subfigure[]{
		\begin{minipage}{0.13\textwidth}
			\includegraphics[height=0.07\textheight]{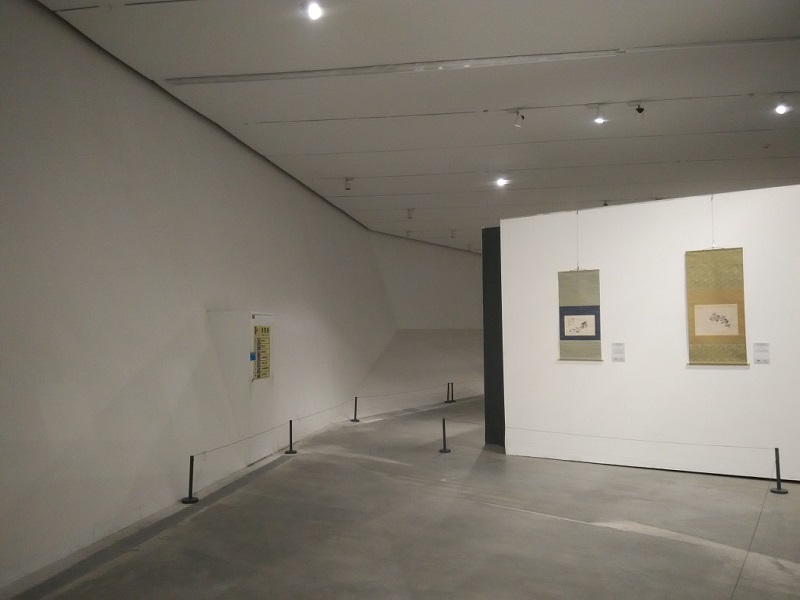}\\
			\includegraphics[height=0.07\textheight]{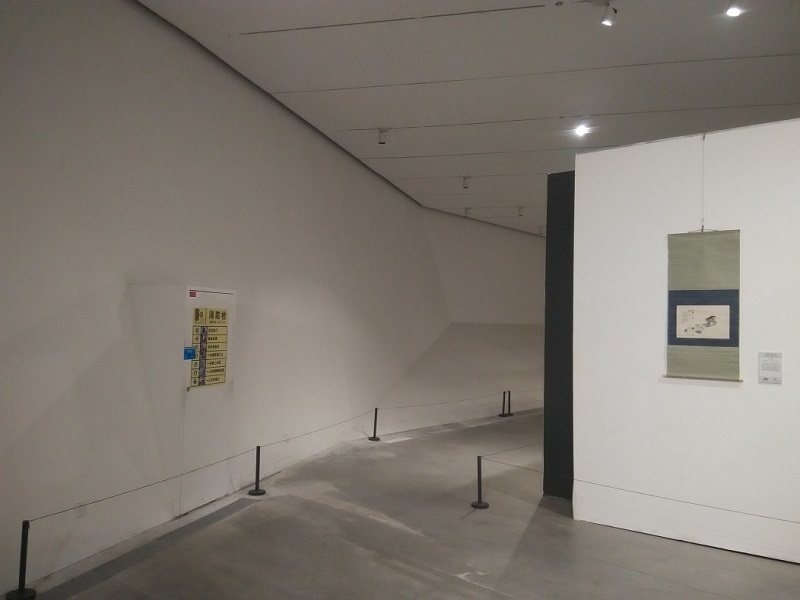}
		\end{minipage}
	}	
	\subfigure[]{
		\begin{minipage}{0.13\textwidth}
			\includegraphics[height=0.07\textheight]{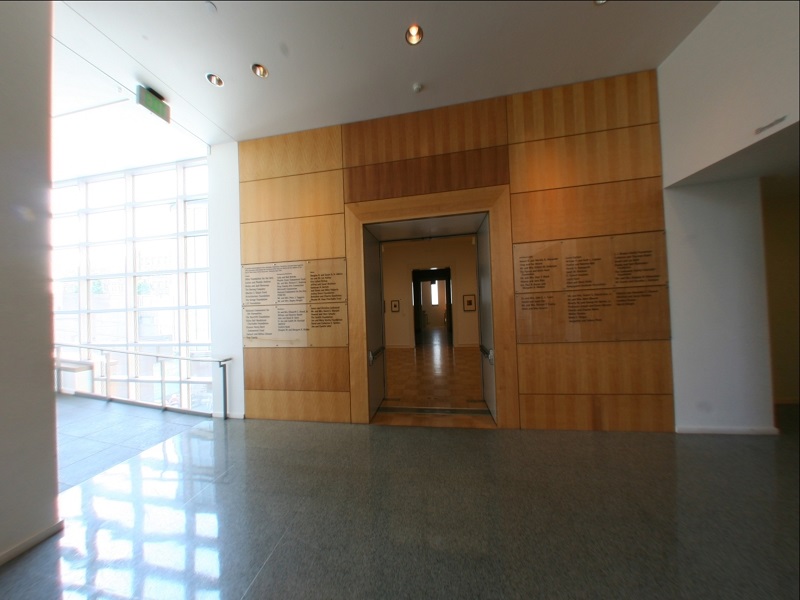}\\
			\includegraphics[height=0.07\textheight]{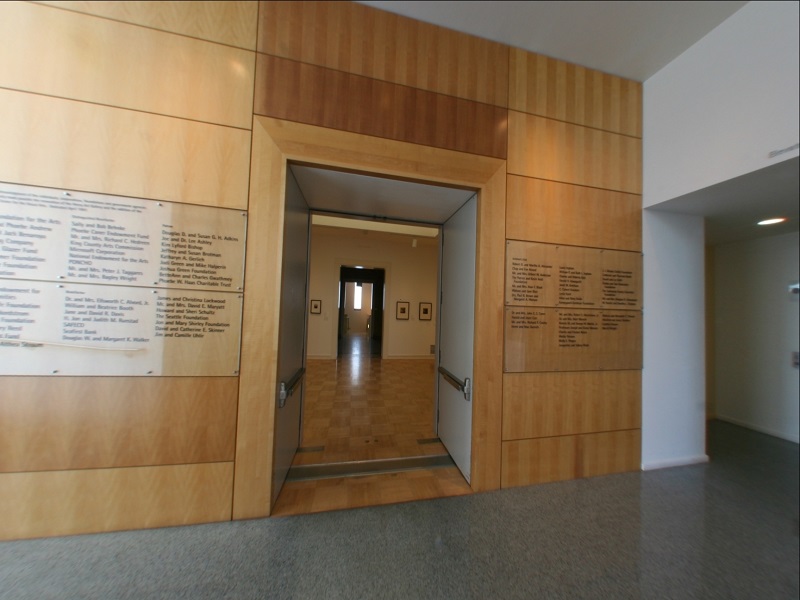}
		\end{minipage}
	}	
	\subfigure[]{
		\begin{minipage}{0.13\textwidth}
			\includegraphics[height=0.07\textheight]{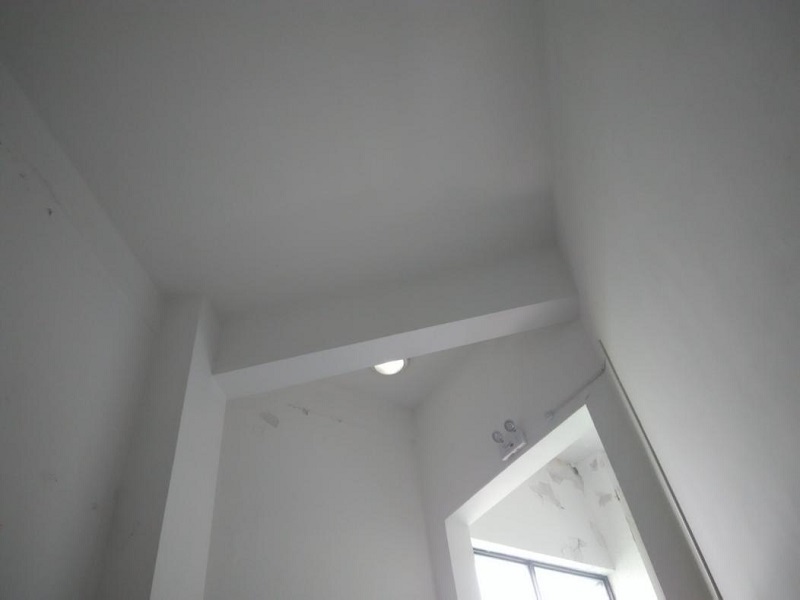}\\
			\includegraphics[height=0.07\textheight]{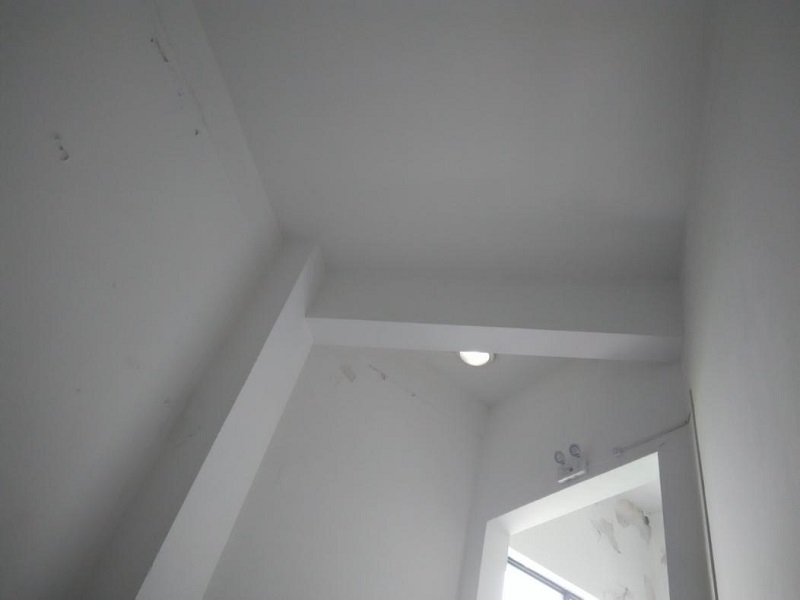}
		\end{minipage}
	}	
	\subfigure[]{
		\begin{minipage}{0.13\textwidth}
			\includegraphics[height=0.07\textheight]{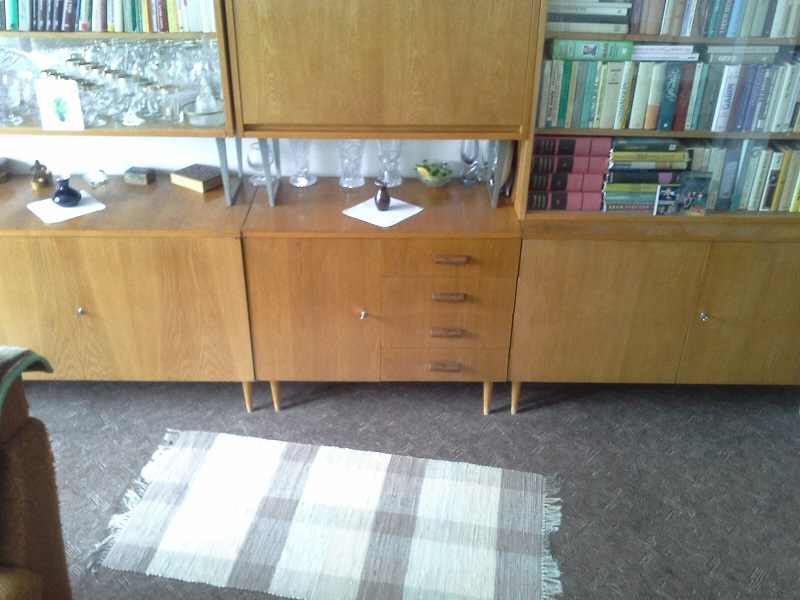}\\
			\includegraphics[height=0.07\textheight]{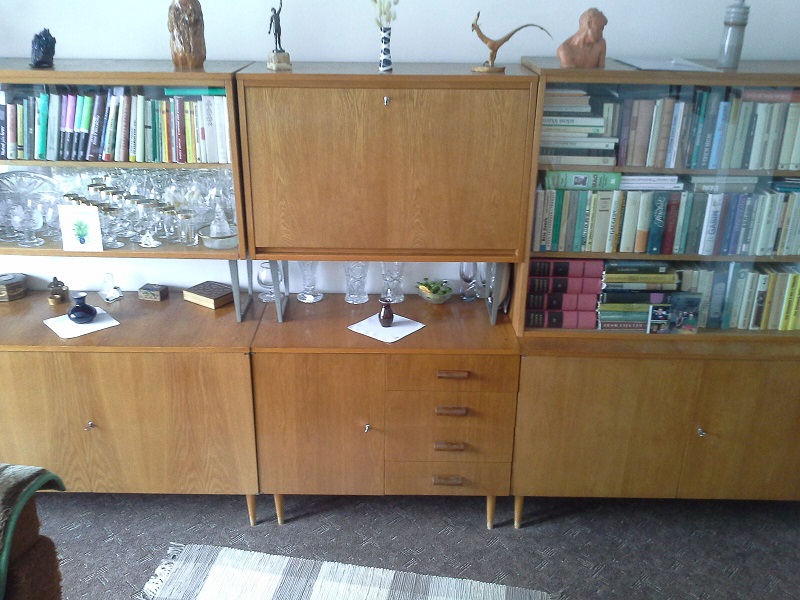}
		\end{minipage}
	}	
	\subfigure[]{
		\begin{minipage}{0.13\textwidth}
			\includegraphics[height=0.07\textheight]{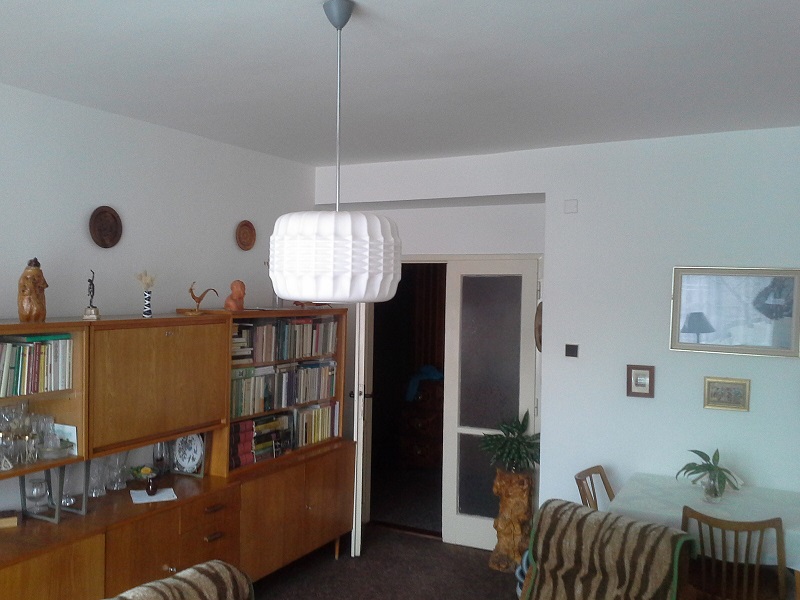}\\
			\includegraphics[height=0.07\textheight]{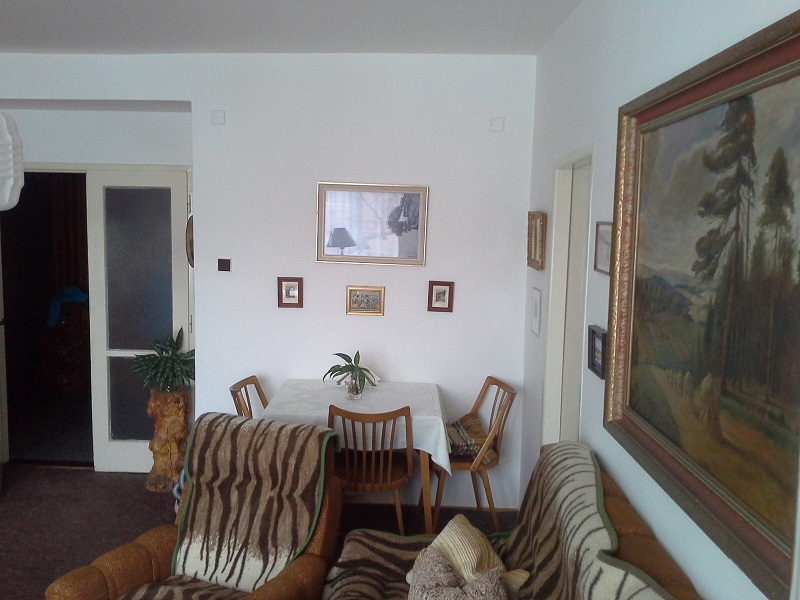}
		\end{minipage}
	}	
	\subfigure[]{
		\begin{minipage}{0.13\textwidth}
			\includegraphics[height=0.07\textheight]{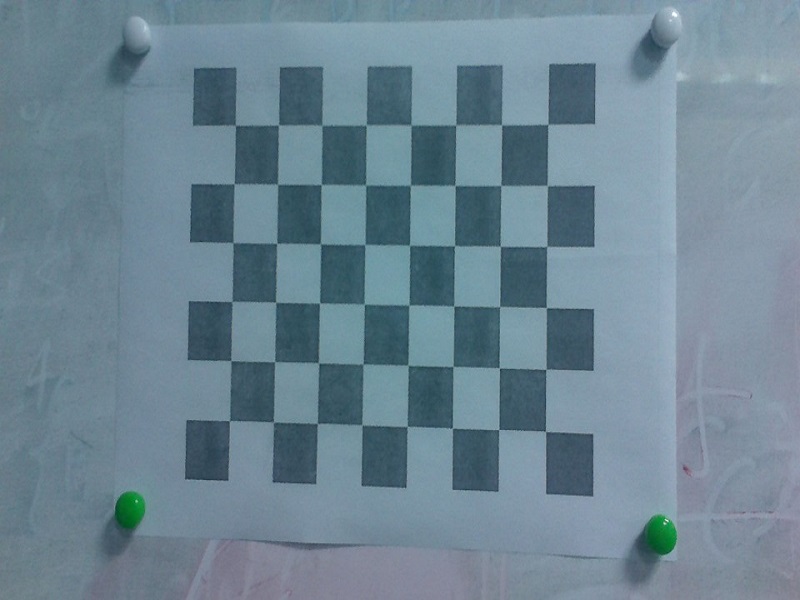}\\
			\includegraphics[height=0.07\textheight]{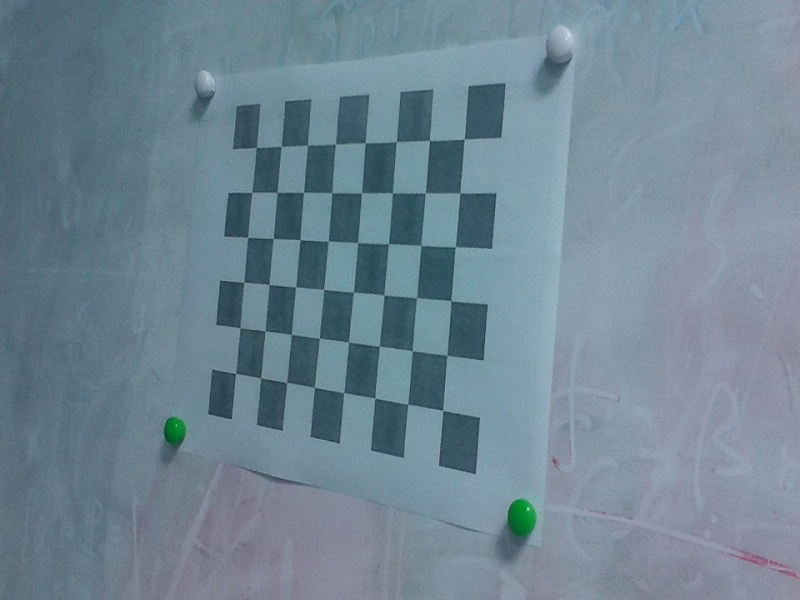}
		\end{minipage}
	}	
	\subfigure[]{
		\begin{minipage}{0.13\textwidth}
			\includegraphics[height=0.07\textheight]{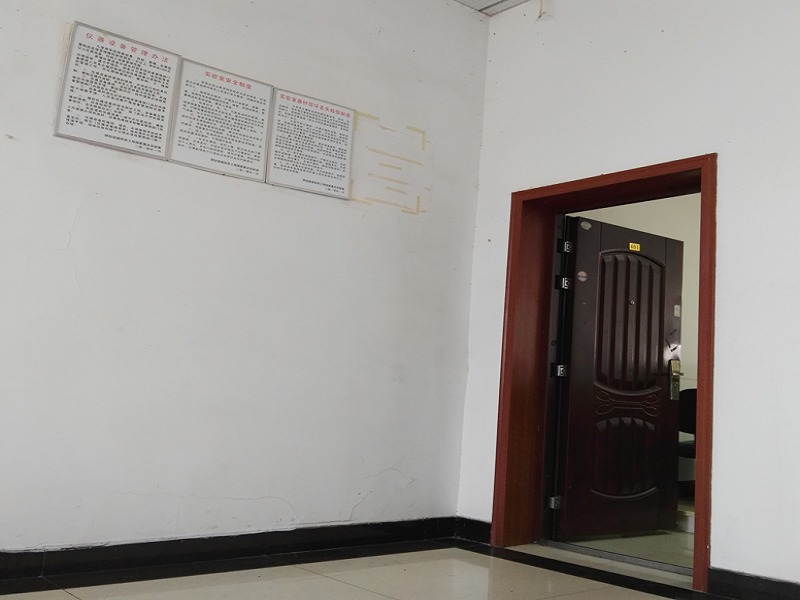}\\
			\includegraphics[height=0.07\textheight]{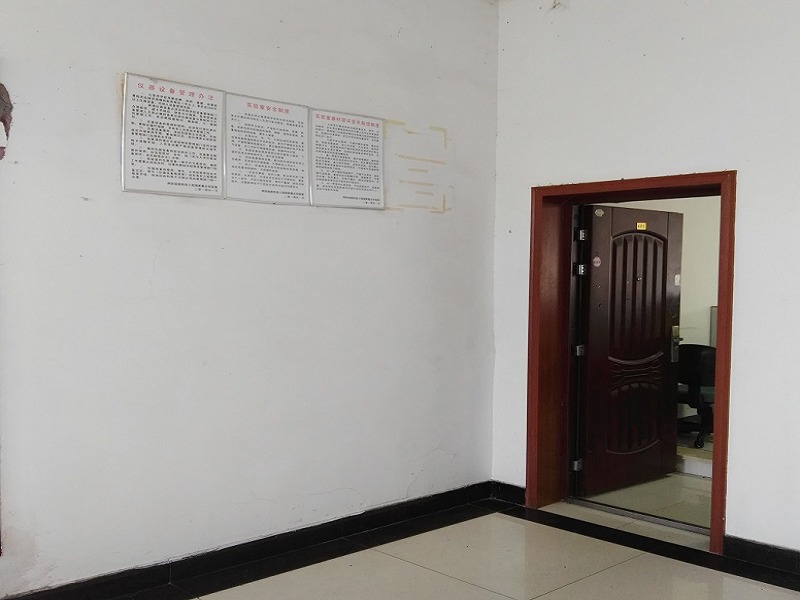}
		\end{minipage}
	}	
	\subfigure[]{
		\begin{minipage}{0.13\textwidth}
			\includegraphics[height=0.07\textheight]{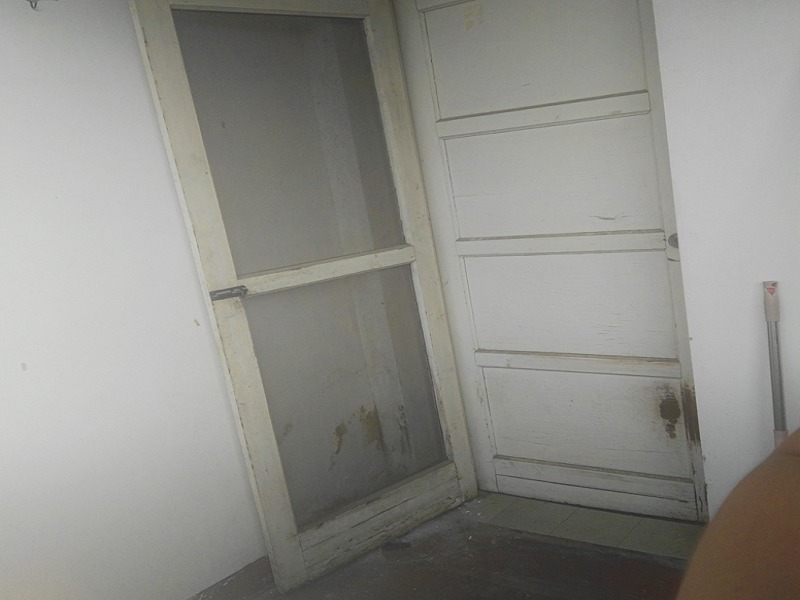}\\
			\includegraphics[height=0.07\textheight]{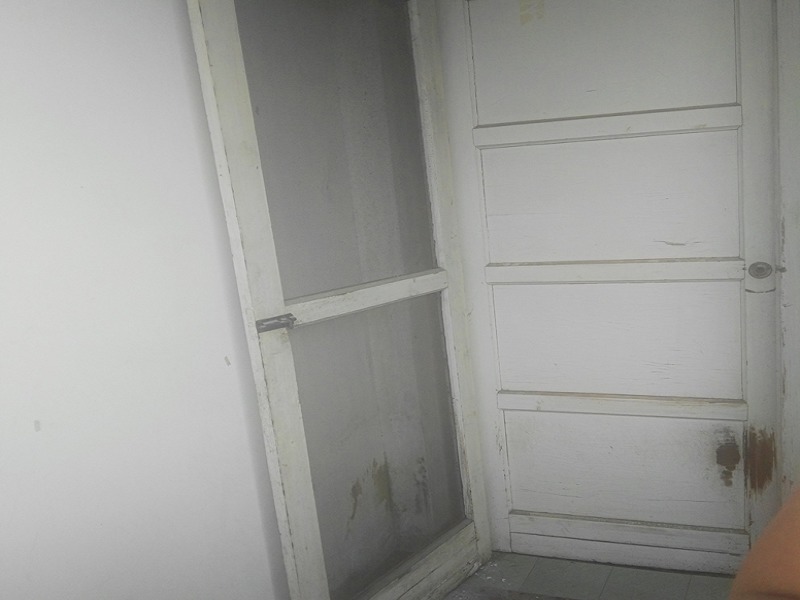}
		\end{minipage}
	}	
	\subfigure[]{
		\begin{minipage}{0.13\textwidth}
			\includegraphics[height=0.07\textheight]{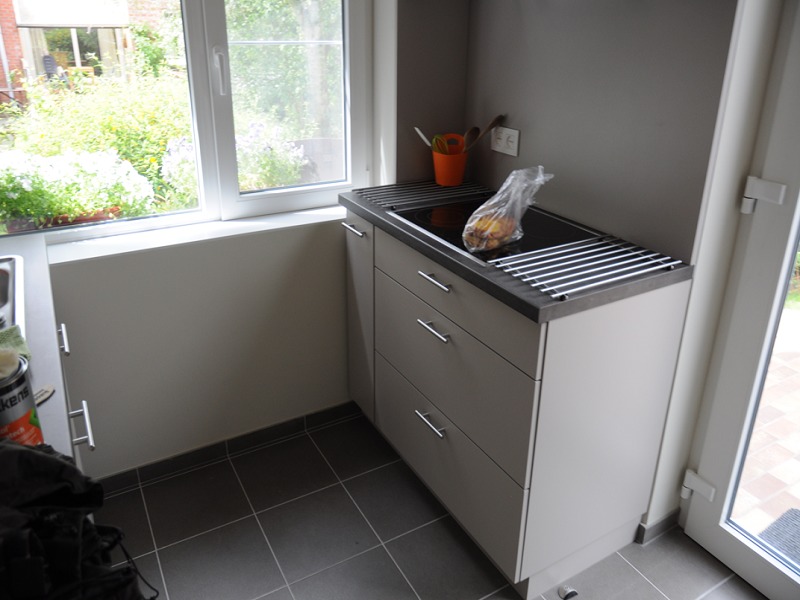}\\
			\includegraphics[height=0.07\textheight]{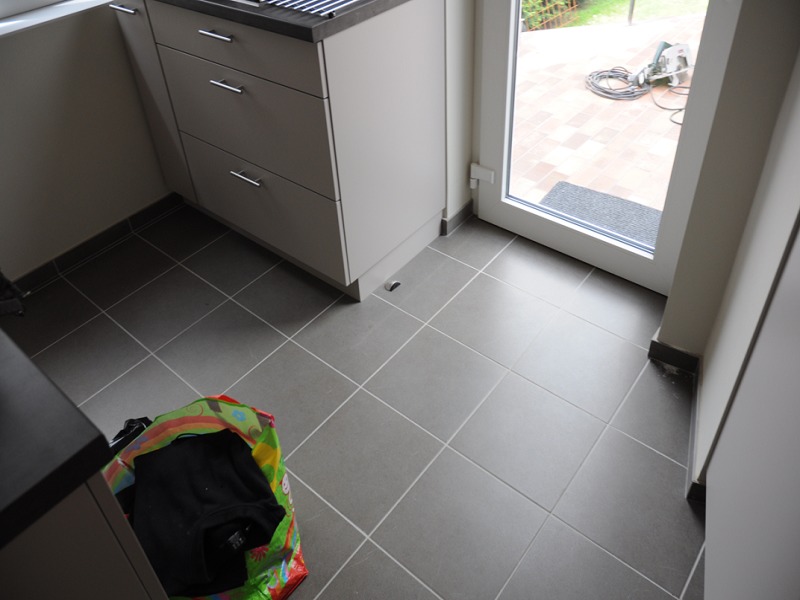}
		\end{minipage}
	}								
	\caption{Some example of collected indoor images used for comparison of different matching approaches. }
	\label{fig:imageshow}
\end{figure*}
In order to evaluate our approach, we collect more than 100 images to perform our proposed approach ASJ. Some of the collected images are from indoor 3D reconstruction dataset~\cite{FurukawaCSS09,SrajerSPP14} while others are taken by ourselves. As shown in Fig.~\ref{fig:imageshow}, the collected images are less texture than natural images. Some of them contain large viewpoint changes and indistinct texture repeated regions such as Fig.~\ref{fig:imageshow}(b), Fig.~\ref{fig:imageshow}(i) and Fig.~\ref{fig:imageshow}(l).

We define that two junctions are matched, only if the junction centers and orientations of branches are corresponding.
In this sense, our
matching result is somewhat beyond of local features and can be compared with
existing approaches in different settings:
\begin{itemize}
	\item[-] It is comparable to keypoint matching methods, if we regard junctions as a specific corner points with two orientations;
	\item[-] It is also comparable to line segment matching ones, if we take junctions as several intersecting line segments.
\end{itemize}

 For key-points matching, we compare the results of matched junctions with SIFT~\cite{Lowe04}, Affine-SIFT~\cite{MorelY09,YuM11}, Hessian-Affine~\cite{PerdochCM09}, EBR and IBR in~\cite{TuytelaarsG04}.

Meanwhile, we compare maching accuracy with existing approaches LPI~\cite{FanWH12B} and LJL~\cite{LiYLLZ16} for matched line segments that measures the proportion of matches if their endpoints are corresponding. This rule is more strict for assessing line segment matching results. Interestingly, the approaches LPI~\cite{LiYLLZ16} and LJL~\cite{LiYLLZ16} use the epipolar geometry without outliers to assist their line segment matcher, our proposed method without epipolar geometry achieves better accuracy.

The implementation for Affine-SIFT~\cite{MorelY09,YuM11}, Hessian Affine~\cite{PerdochCM09}, LPI~\cite{FanWH12B} and LJL~\cite{LiYLLZ16} are getting from authors' homepage. EBR and IBR~\cite{TuytelaarsG04} are got from VGG's website\footnote{\url{http://www.robots.ox.ac.uk/~vgg/research/affine/descriptors.html#binaries}}. The version of SIFT detector is provided by VLFeat\footnote{\url{http://www.vlfeat.org}}. The descriptor used in our experiment is SIFT and the mismatches are filtered according to the ratio test with threshold $1.5$ for ASJ , SIFT~\cite{Lowe04}, Hessian-Affine~\cite{PerdochCM09}, EBR~\cite{TuytelaarsG04} and IBR~\cite{TuytelaarsG04} by comparing the $\ell_2$-distance, which is the default threshold for computing matches from descriptor in VLFeat. Remarkably, the implementation of Affine-SIFT~\cite{MorelY09,YuM11} provided by its authors use threshold $1.33$ since they calculate the distance with $\ell_1$ norm and we keep it unchanged. Since the released code for Affine-SIFT~\cite{MorelY09,YuM11} produce the matched result with outliers filtering, we remove this procedure in all fairness, which makes the results in our experiment different from the released executable program. All of parameters for compared approaches are the default value which is provided by their authors.

\begin{table*}[htb!]
	\newcommand{\tabincell}[2]{\begin{tabular}{@{}#1@{}}#2\end{tabular}}
	
	\caption{Comparison of different matching methods. The number of correct matches, number of total matches and the matching accuracy for the comparision with key-points matching results are reported in the first row. The results for key point matching approaches SIFT~\cite{Lowe04}, Affine-SIFT~\cite{MorelY09,YuM11}, Hessian-Affine~\cite{PerdochCM09}, EBR~\cite{TuytelaarsG04} and IBR~\cite{TuytelaarsG04} are list in the 3-th row to 7-th row. The average matching accuracy for all collected images is reported in the last column.} 		
	\resizebox{\textwidth}{!}{
	\begin{tabular}{{c|c|cccccccccccc|c}}
		\hline
		\multicolumn{2}{r|}{\multirow{2}*{\diagbox{Methods}{Image pairs}}} &
		\multirow{2}*{(a)} &
		\multirow{2}*{(b)} &
		\multirow{2}*{(c)} &
		\multirow{2}*{(d)} &
		\multirow{2}*{(e)} &
		\multirow{2}*{(f)} &
		\multirow{2}*{(g)} &
		\multirow{2}*{(h)} &
		\multirow{2}*{(i)} &
		\multirow{2}*{(j)} &
		\multirow{2}*{(k)} &
		\multirow{2}*{(l)} & \multirow{2}*{\tabincell{c}{Average\\accuracy}}\\
		\multicolumn{2}{r|}{} & & & & & & & & & & & & \\\hline
		\multirow{3}*{\tabincell{c}{Ours\\(Junctions)}}     &
		\#correct &
		12 & 26 & 12 & 16 &50 &14 &197&65  &119 & 37 &17 &33
		& \multirow{3}*{\bf{85.17}\%}\\ \cline{2-14}
		\multirow{3}*{} &
		\#total  &
		12 & 29 & 13 & 20 & 60& 15& 214&69 & 121& 45&19 &40 & \\\cline{2-14}
		\multirow{3}*{} &
		accuracy (\%)   &
		\textbf{100.00} & \textbf{89.66} & \textbf{92.31} & \textbf{80.00} & \textbf{83.33} & 93.33 & 92.06 & 94.20 & \textbf{98.35} & \textbf{82.22} & 89.47 & \textbf{82.50} & \\\hline	
		\multirow{3}*{\tabincell{c}{SIFT~\cite{Lowe04}}}     &
		\#correct &
		128   & 476   & 559   & 115   & 435   & 74    & 708   & 199   & 147   & 200   & 103   & 65  & \multirow{3}*{62.69\%}\\ \cline{2-14}
		\multirow{3}*{} &
		\#total  &
		206   & 700   & 839   & 222   & 652   & 287   & 770   & 261   & 330   & 299   & 191   & 161 & \\\cline{2-14}
		\multirow{3}*{} &
		accuracy (\%)   &
		62.14 & 68.00    & 66.63 & 51.80 & 66.72 & 25.78 & 91.95 & 76.25 & 44.55 & 66.89 & 53.93 & 40.37  & \\\hline	
		\multirow{3}*{\tabincell{c}{Affine-SIFT~\cite{YuM11}}}     &
		\#correct &
		135   & 183   & 433   & 364   & 430   & 119   & 4141  & 1271  & 136   & 172   & 196   & 163
		& \multirow{3}*{82.85\%}\\ \cline{2-14}
		\multirow{3}*{} &
		\#total  &
		141   & 240   & 519   & 480   & 549   & 133   & 4205  & 1326  & 240   & 263   & 224   & 264
		& \\\cline{2-14}
		\multirow{3}*{} &
		accuracy (\%)   &
		95.74  & 76.25  & 83.43  & 75.83  & 78.32  & 89.47  & \textbf{98.48}  & \textbf{95.85}  & 56.67  & 65.40  & 87.50  & 61.74
		& \\\hline	
		\multirow{3}*{\tabincell{c}{Hessian-Affine~\cite{PerdochCM09}}}     &
		\#correct &
		24    & 13    & 96    & 82    & 66    & 17    & 640   & 226   & 32    & 114   & 38    & 29
		& \multirow{3}*{79.68\%}\\ \cline{2-14}
		\multirow{3}*{} &
		\#total  &
		26    & 34    & 132   & 105   & 109   & 18    & 671   & 248   & 63    & 144   & 41    & 49
		& \\\cline{2-14}
		\multirow{3}*{} &
		accuracy (\%)   &
		92.31  & 38.24  & 72.73  & 78.10  & 60.55  & \textbf{94.44} & 95.38  & 91.13  & 50.79  & 79.17  & \textbf{92.68}  & 59.18
		& \\\hline
		\multirow{3}*{\tabincell{c}{EBR~\cite{TuytelaarsG04}}}     &
		\#correct &
		0     & 0     & 10    & 0     & 0     & 0     & 64    & 20    & 28    & 14    & 0     & 0
		& \multirow{3}*{32.56\%}\\ \cline{2-14}
		\multirow{3}*{} &
		\#total  &
		1     & 1     & 16    & 15    & 10    & 2     & 75    & 21    & 46    & 34    & 2     & 4
		& \\\cline{2-14}
		\multirow{3}*{} &
		accuracy (\%)   &
		0.00  & 0.00  & 62.50  & 0.00  & 0.00  & 0.00  & 85.33  & 95.24  & 60.87  & 41.18  & 0.00  & 0.00
		& \\\hline	
		\multirow{3}*{\tabincell{c}{IBR~\cite{TuytelaarsG04}}}     &
		\#correct &
		0     & 0     & 28    & 11    & 14    & 0     & 46    & 0     & 0     & 10    & 0     & 0
		& \multirow{3}*{31.84\%}\\ \cline{2-14}
		\multirow{3}*{} &
		\#total  &
		4     & 9     & 39    & 16    & 25    & 9     & 63    & 12    & 10    & 21    & 8     & 5
		& \\\cline{2-14}
		\multirow{3}*{} &
		accuracy (\%)   &
		0.00  & 0.00  & 71.79  & 68.75  & 56.00  & 0.00  & 73.02  & 0.00  & 0.00  & 47.62  & 0.00  & 0.00 		
		& \\\hline	
	\end{tabular}
	}
	\label{tab:matchkeypoint_results}
\end{table*}
\subsubsection{Matching results for key-points matching}
As shown in Tab.~\ref{tab:matchkeypoint_results}, our proposed feature ASJ is compare with most widely used feature detectors. In the sense for key-points matching, we regard an ASJ as a key-point with two specific orientations. The matching accuracy for ASJ is better than other key-points matches in most cases. Representatively, in Fig.~\ref{fig:imageshow}(i), the indistinct repeated region in chessboard are matched very well with the accuracy $98.35\%$ since ASJs makes corner points contain more global information than other approaches, which represents the relative position with meaningful orientations in images.

Comparing with the most related approach EBR and IBR~\cite{TuytelaarsG04}, our proposed approach ASJ handles straight edges in a better way which can produce more key-points and more correct correspondences. In many cases as shown in Tab.~\ref{tab:matchkeypoint_results}, the results of EBR and IBR illustrate their limitation in indoor images which are dominated by straight edges.

\begin{figure*}[t!]
\centering
\subfigure[]
{	
\begin{minipage}[b]{0.31\linewidth} 
	\includegraphics[width=0.48\linewidth]{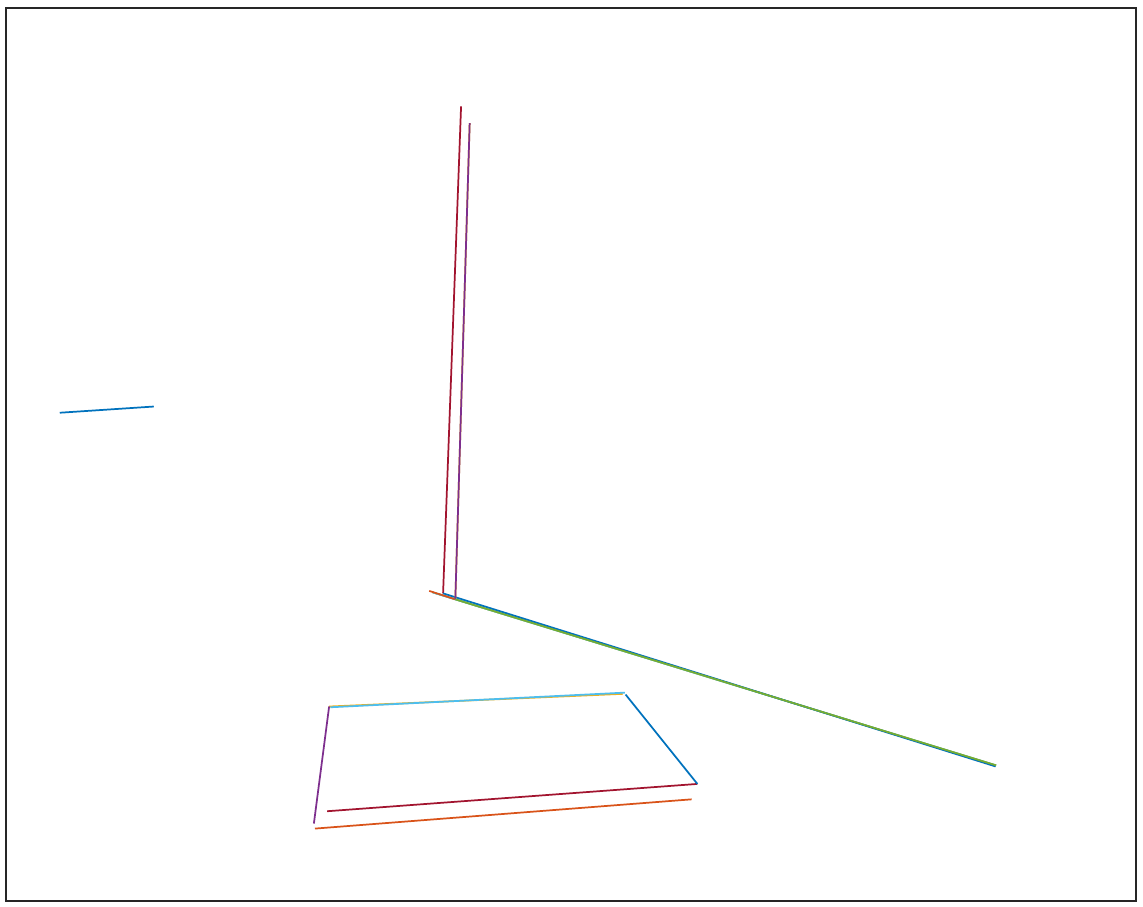}
	\includegraphics[width=0.48\linewidth]{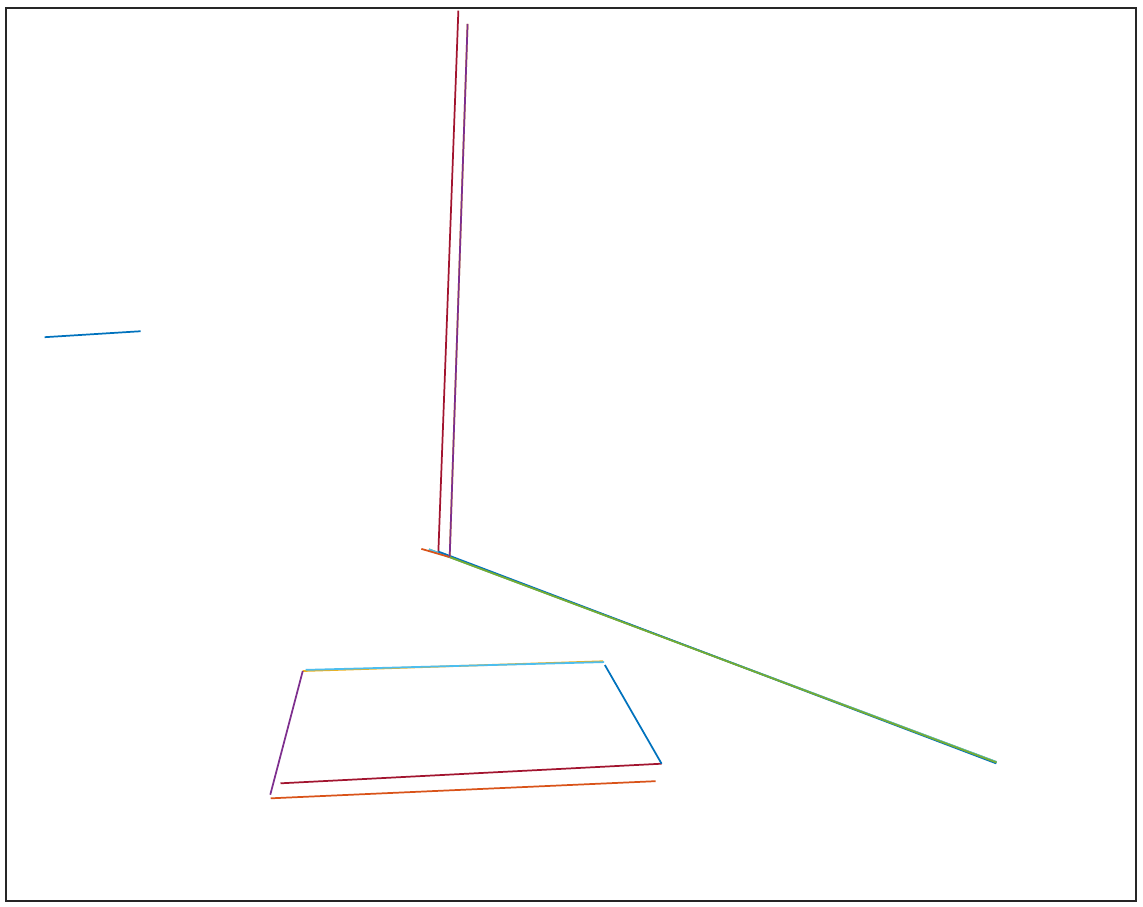}\\
	\includegraphics[width=0.48\linewidth]{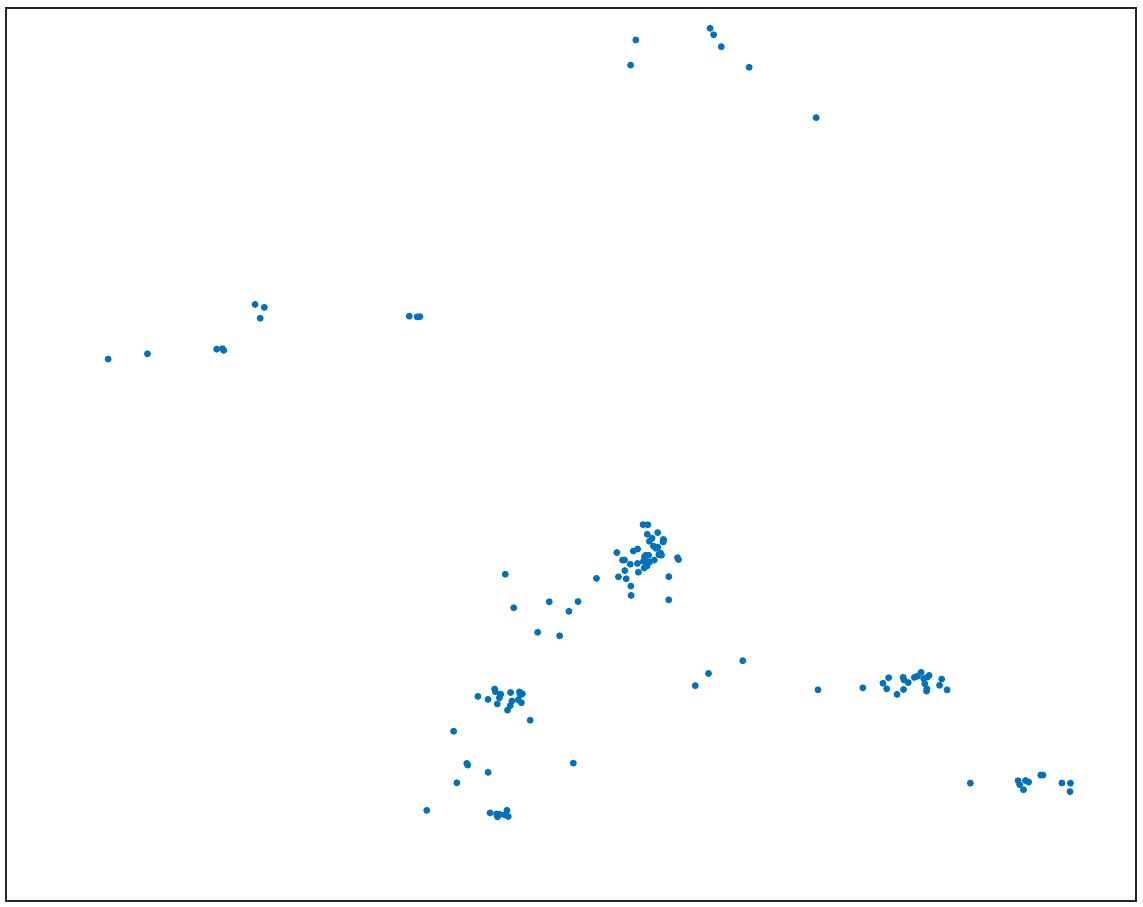}
	\includegraphics[width=0.48\linewidth]{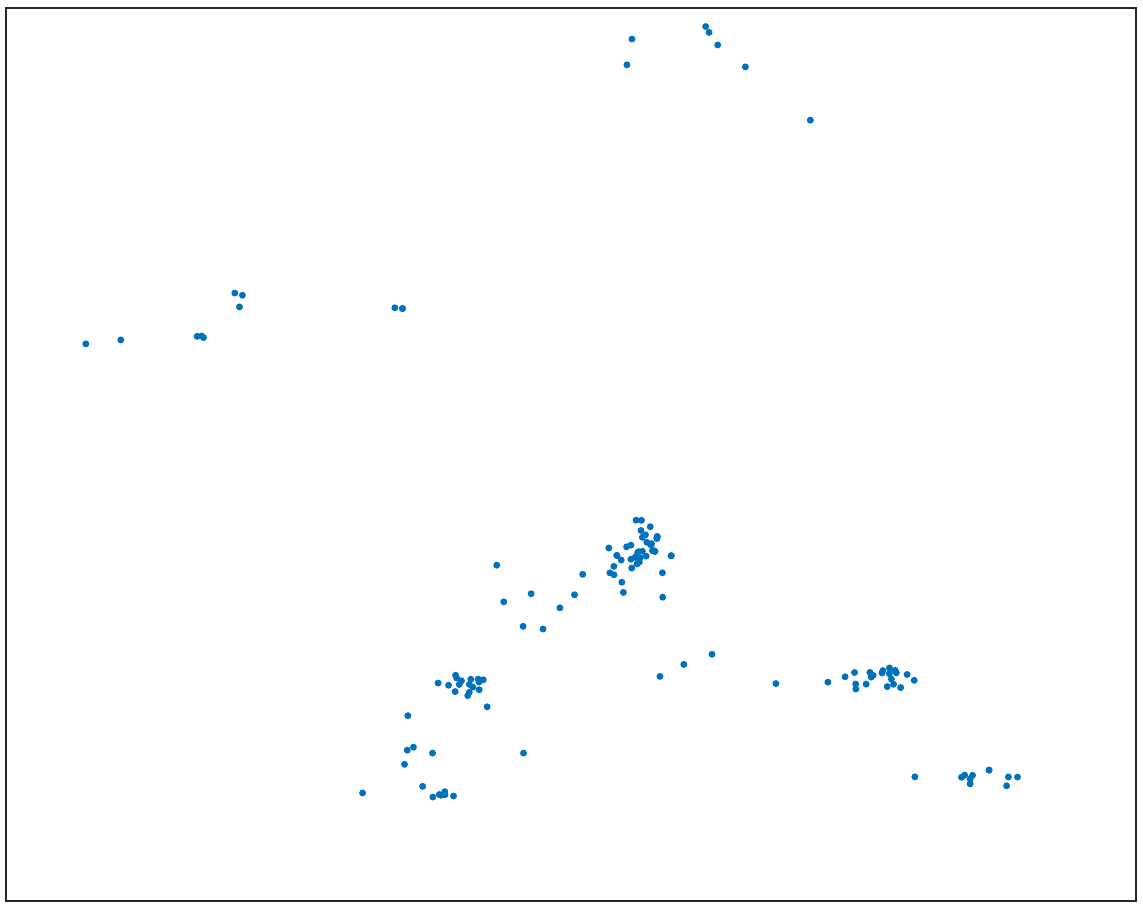}	
\end{minipage}
}
\subfigure[]
{
\begin{minipage}[b]{0.31\linewidth} 
	\includegraphics[width=0.48\linewidth]{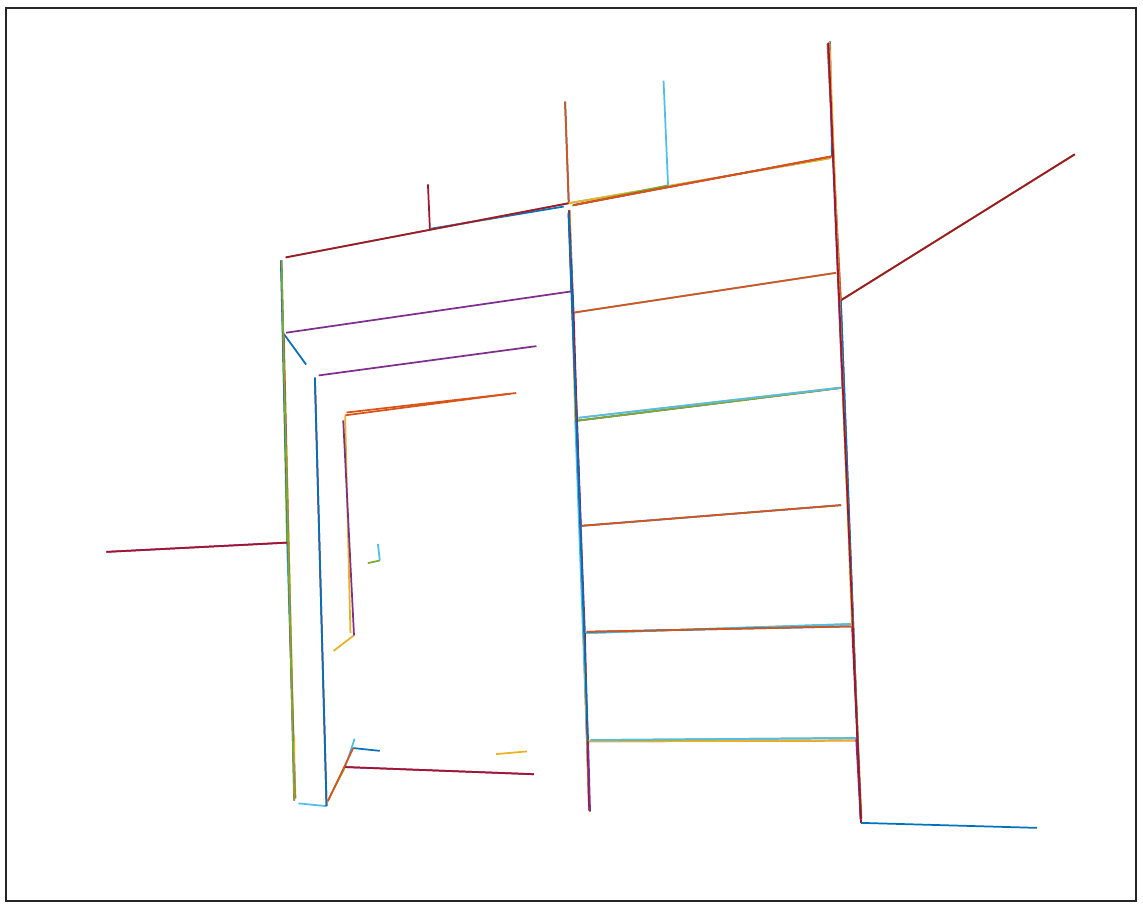}
	\includegraphics[width=0.48\linewidth]{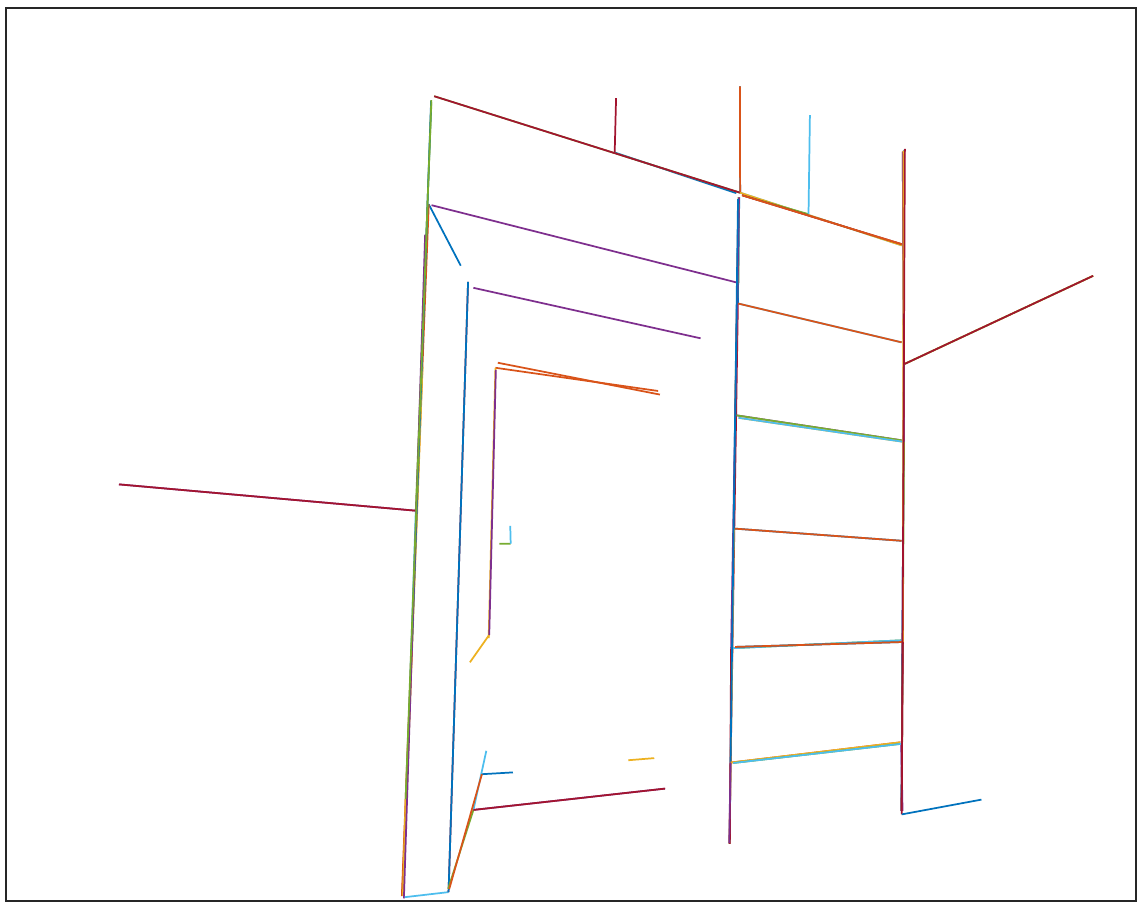}\\
	\includegraphics[width=0.48\linewidth]{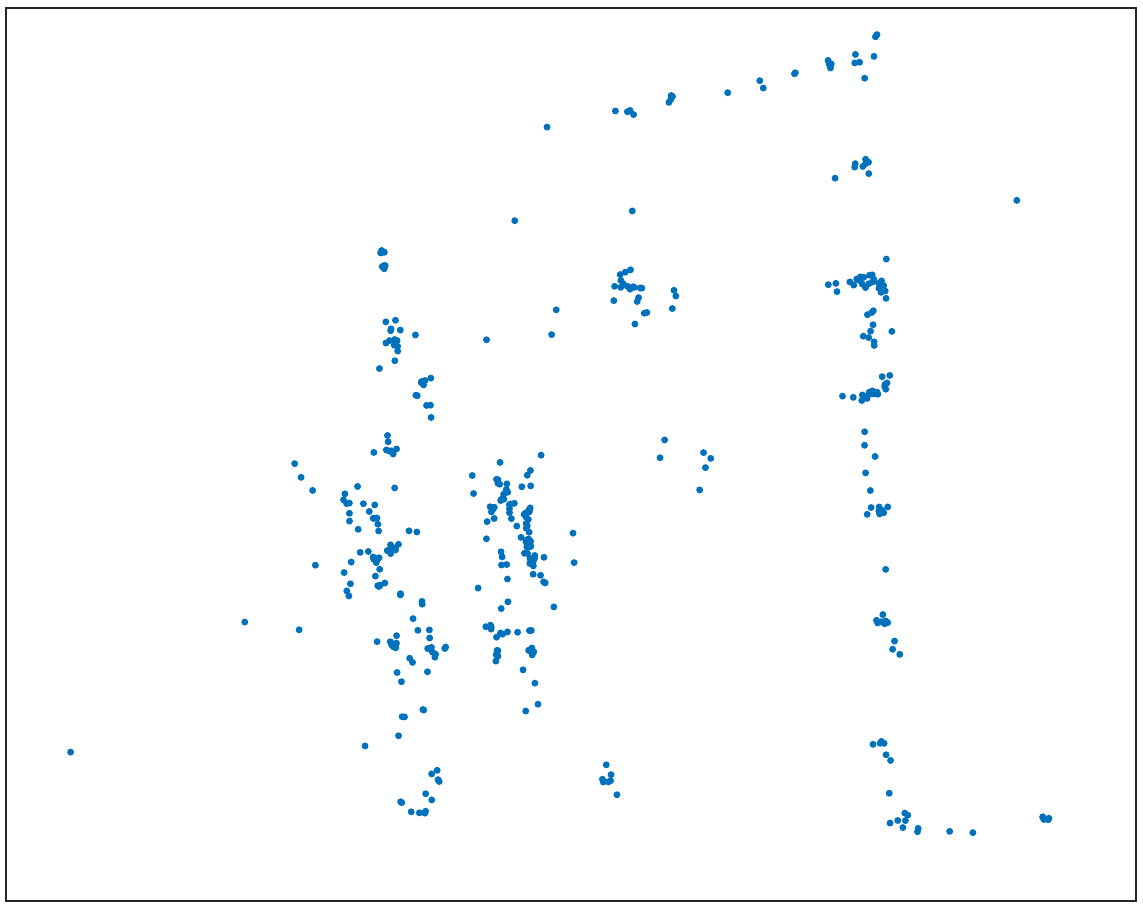}
	\includegraphics[width=0.48\linewidth]{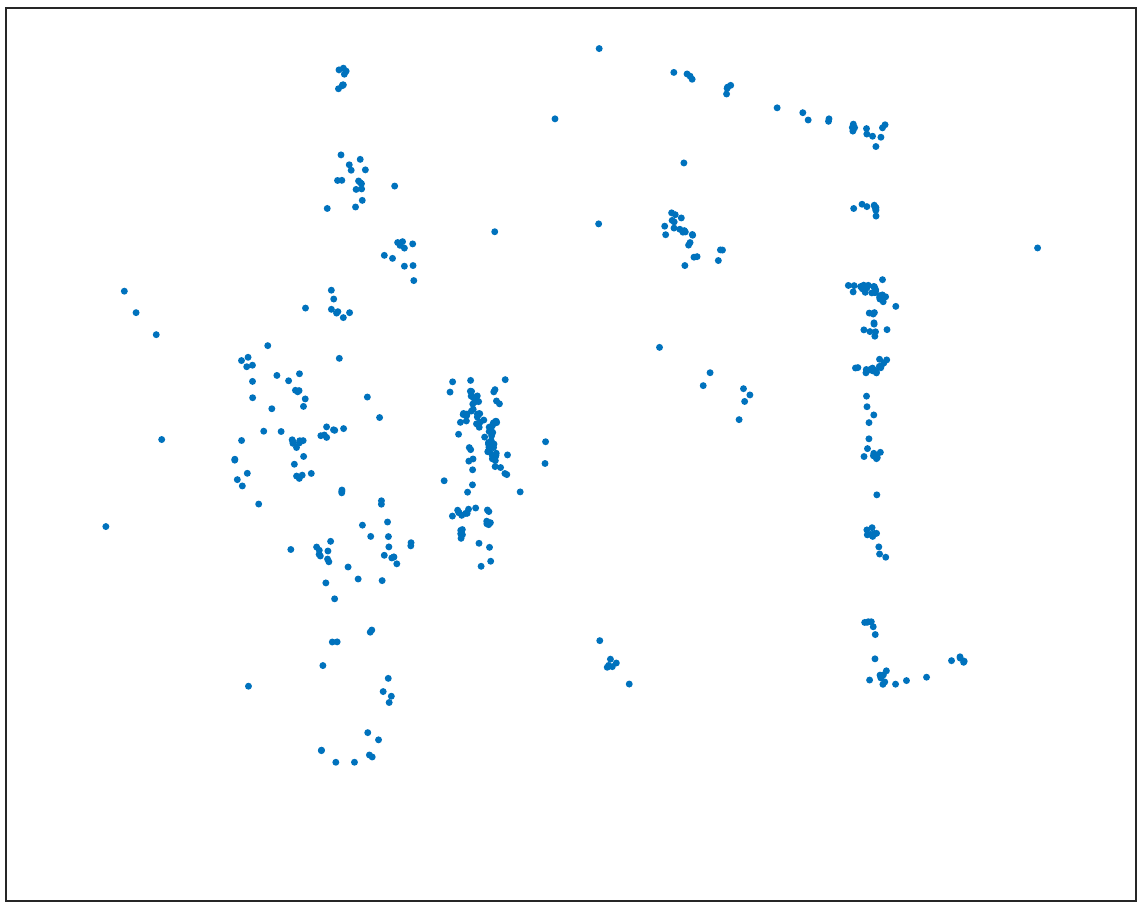}	
\end{minipage}
}
\subfigure[]
{
\begin{minipage}[b]{0.31\linewidth} 
	\includegraphics[width=0.48\linewidth]{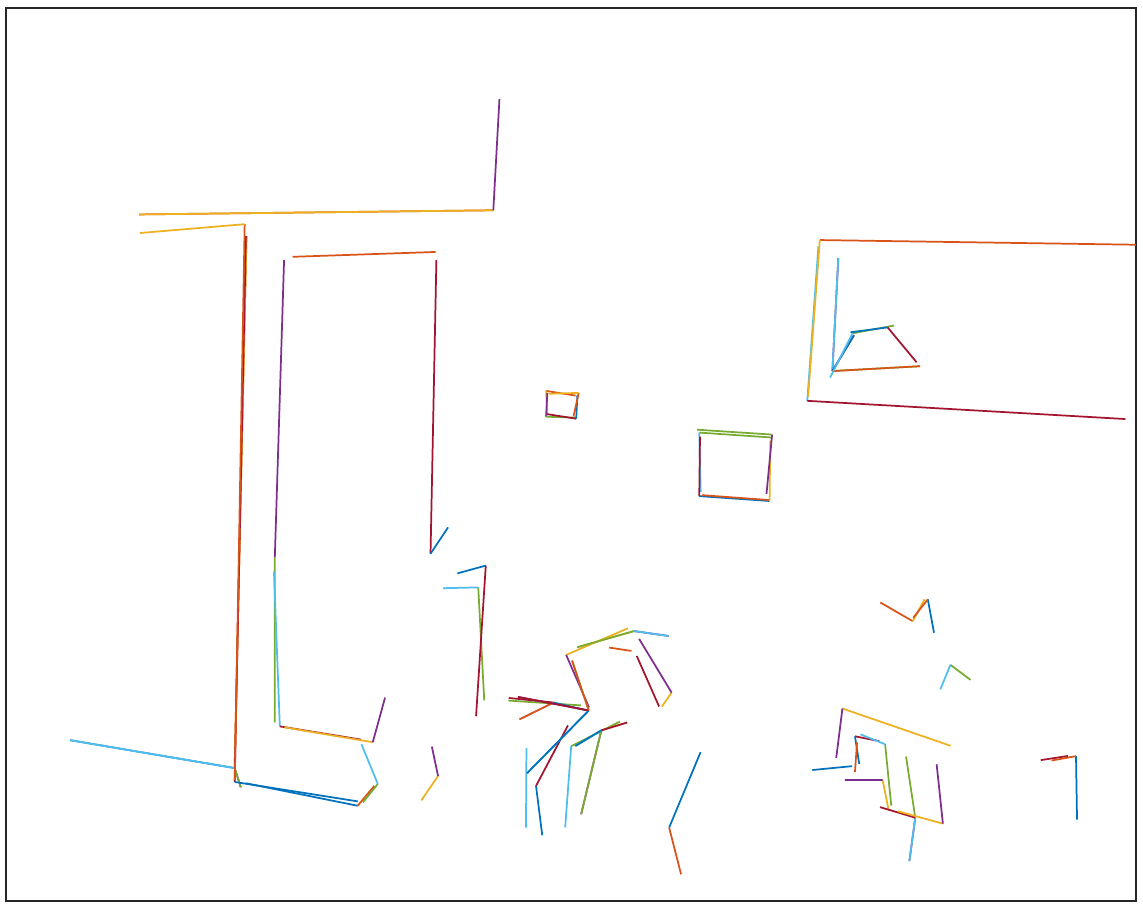}
	\includegraphics[width=0.48\linewidth]{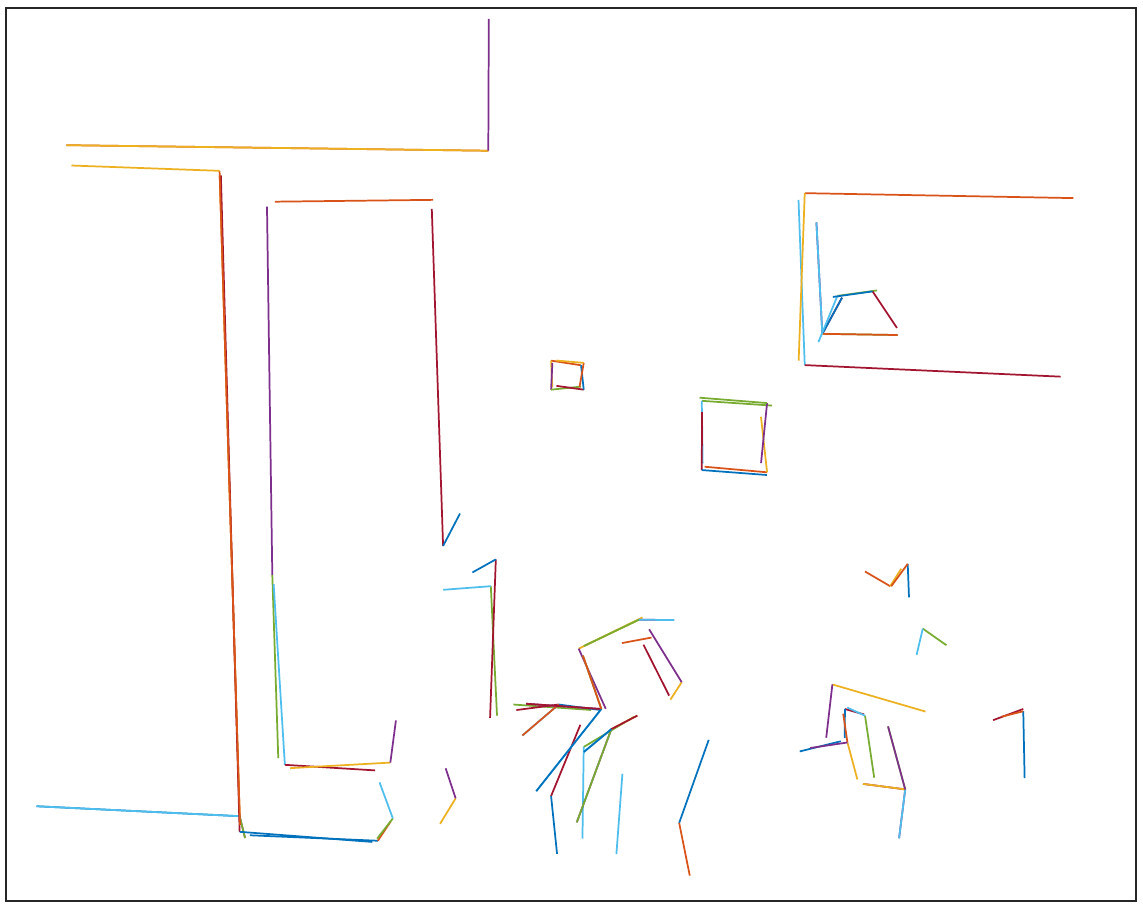}\\
	\includegraphics[width=0.48\linewidth]{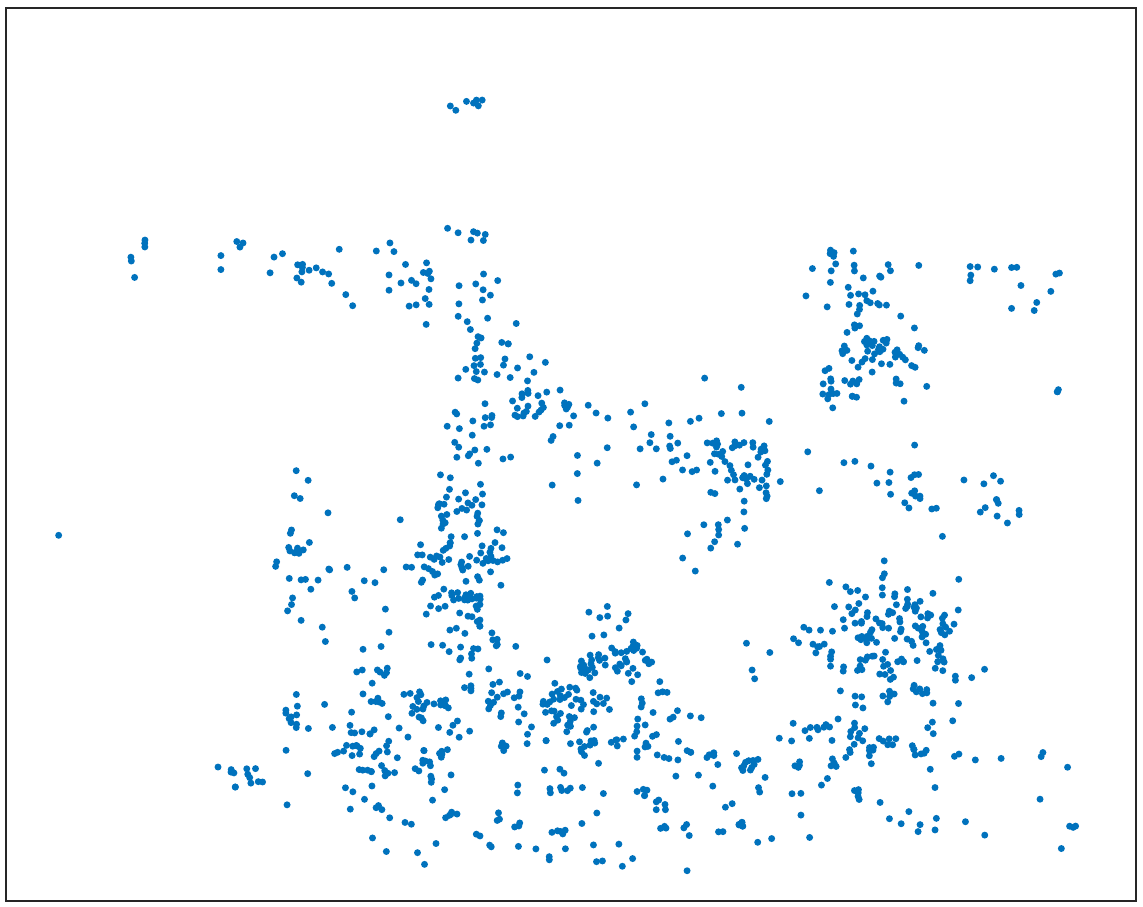}
	\includegraphics[width=0.48\linewidth]{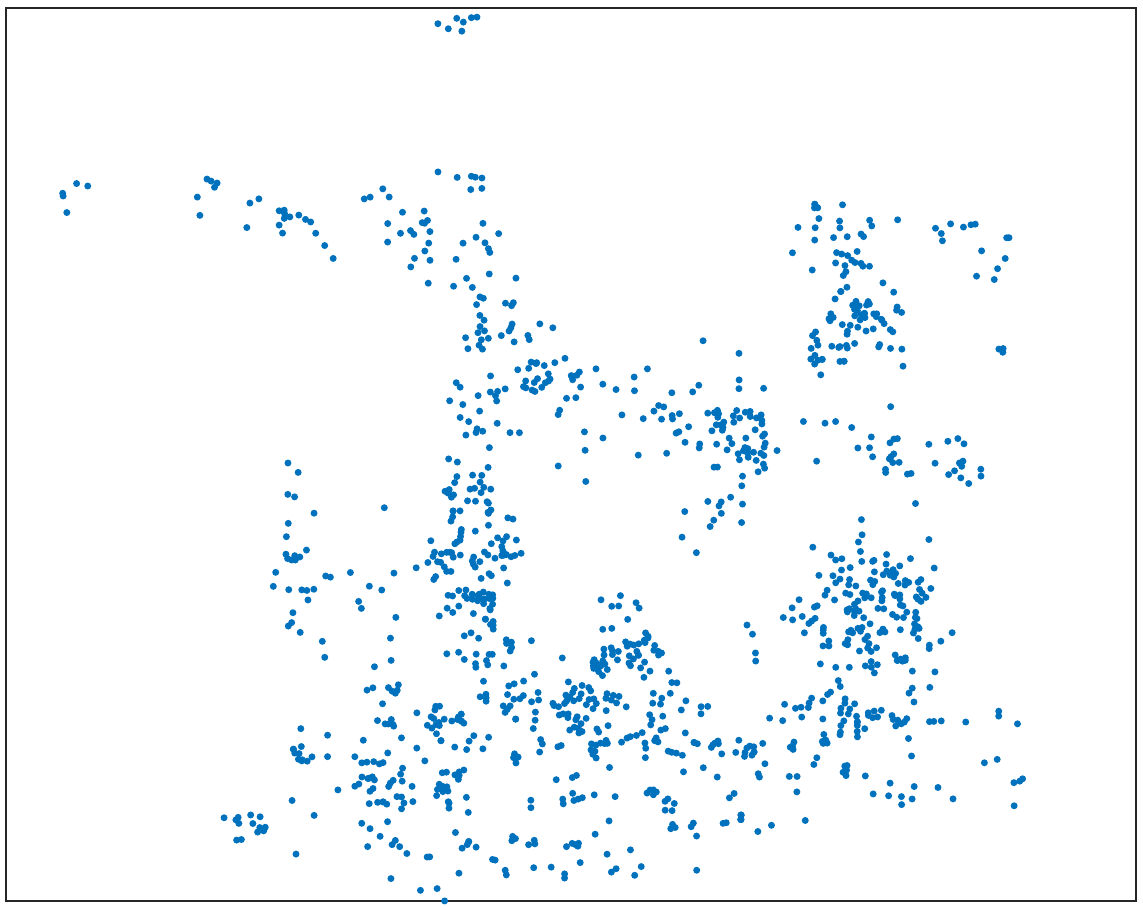}	
\end{minipage}
}
\caption{Top row: plotted correct matched ASJ in image pairs Fig.~\ref{fig:imageshow}(a), Fig.~\ref{fig:imageshow}(e) and Fig.~\ref{fig:imageshow}(h).
Bottom row: plotted correct matched keypoints by using Affine-SIFT~\cite{MorelY09,YuM11}. Although the number of correct matches for Affine-SIFT is more than ASJ, the ASJ can represent structure information for the input images while plotted key-points are confused if we do not have input image for reference.}
\label{fig:plot-ASJ-KPT}
\end{figure*}

\begin{figure*}[htb!]
\centering
\subfigure[(\#correct matches, \#total matches) = (16, 20)]
{
	\includegraphics[width=0.8\linewidth]{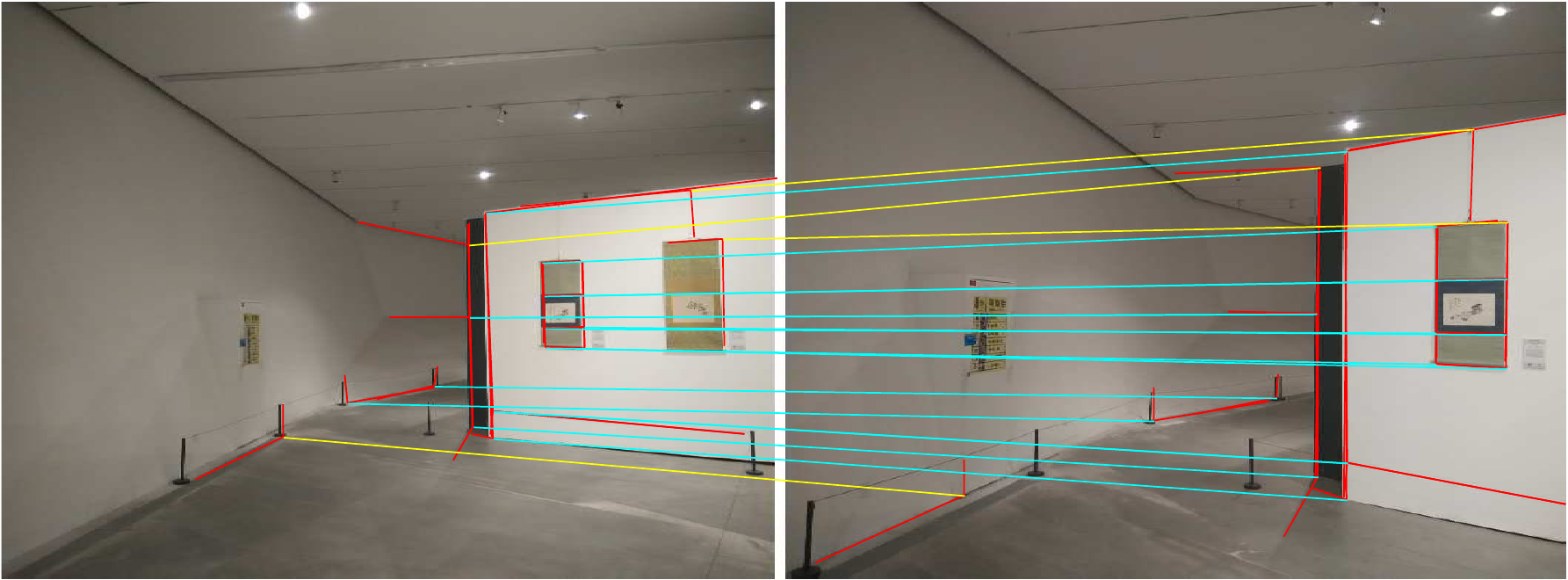}
}
\subfigure[(\#correct matches, \#total matches) = (119, 121)]
{
	\includegraphics[width=0.8\linewidth]{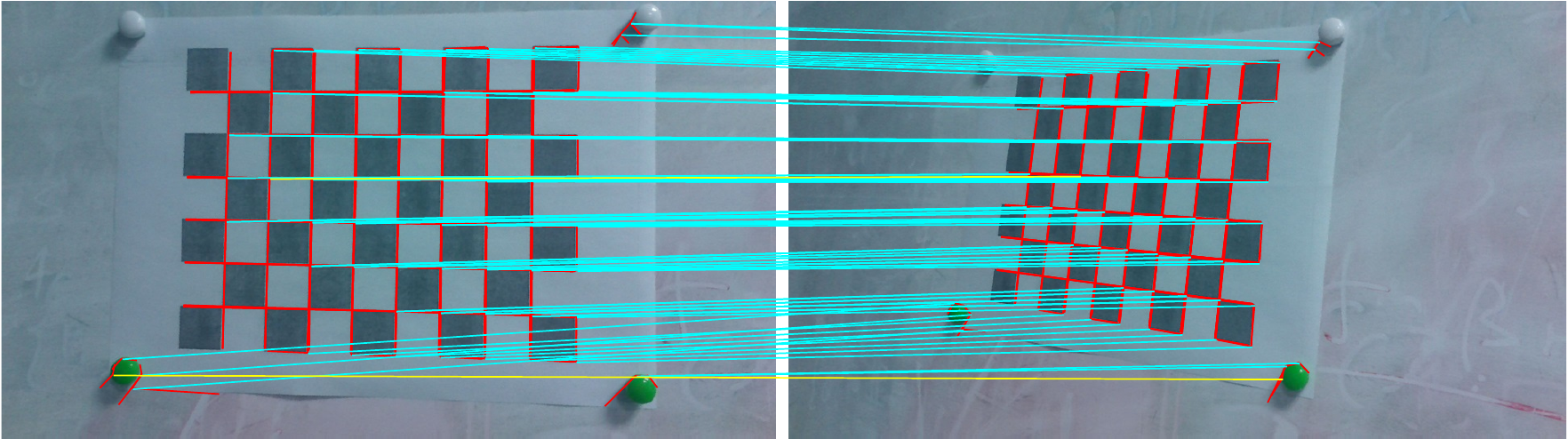}
}
\caption{Matched ASJs for image pairs Fig.~\ref{fig:imageshow} (d) and Fig.~\ref{fig:imageshow} (i) are shown in the sub-figures (a) and (b) respectively. The false matches are connected as yellow lines while correct matches are connected by cyan lines.}
\label{fig:match_ASJ}
\end{figure*}

In the aspect of absolute number of correct matches, ASJ is less than other approaches significantly. The approaches matching most number of correct matches are Affine-SIFT and SIFT. Since the junctions detected in indoor images represents the meaningful junctions in the scene, the fact that absolute number is less than SIFT key-points is not surprising. Nevertheless, ASJ represents the structure information compactly for scenes than key-points. To illustrate this, we plot the correct matched key-points and ASJs in the clean background, the structure of the scene can be represented by ASJs with their branches while plotted key-points are hard to understand without their input images. As shown in Fig.~\ref{fig:plot-ASJ-KPT}, the matched ASJs represents the geometric information with small number of ASJs (12 for Fig.~\ref{fig:plot-ASJ-KPT} (a), 50 for Fig.~\ref{fig:plot-ASJ-KPT} (b) and 65 for Fig.~\ref{fig:plot-ASJ-KPT}(c) while matched ASIFT key-points show confused results even though the amount of matches are much more than ASJs. Some example of match results are shown in Fig.~\ref{fig:match_ASJ}.

\subsubsection{Matching results for line-segments matching}
\begin{table*}[htb!]
	\newcommand{\tabincell}[2]{\begin{tabular}{@{}#1@{}}#2\end{tabular}}
	\centering
	\caption{Comparison of different matching methods for line segment matching. The number of correct matches are counted by the rule that endpoints of corresponding line-segments are correct. We compare ASJ with state-of-the-art approaches LPI~\cite{FanWH12B} and LJL~\cite{LiYLLZ16} and report the number of correct matches, number of total matches and the matching accuracy in this table. The average matching accuracy is also compare in the last column.
	}
	\resizebox{\textwidth}{!}{
	\begin{tabular}{{c|c|cccccccccccc|c}}
		\hline
		\multicolumn{2}{r|}{\multirow{2}*{\diagbox{Methods}{Image pairs}}} &
		\multirow{2}*{(a)} &
		\multirow{2}*{(b)} &
		\multirow{2}*{(c)} &
		\multirow{2}*{(d)} &
		\multirow{2}*{(e)} &
		\multirow{2}*{(f)} &
		\multirow{2}*{(g)} &
		\multirow{2}*{(h)} &
		\multirow{2}*{(i)} &
		\multirow{2}*{(j)} &
		\multirow{2}*{(k)} &
		\multirow{2}*{(l)} & \multirow{2}*{\tabincell{c}{Average\\accuracy}}\\
		\multicolumn{2}{r|}{} & & & & & & & & & & & & \\\hline
		\multirow{3}*{\tabincell{c}{Ours\\(Line segments)}}     &
		\#correct &
			15	&30	&14	&26	&85	&21	&349	&121	&232	&49	&27	&48
		    & \multirow{3}*{\bf{71.55}\%}\\ \cline{2-14}
		\multirow{3}*{} &
		\#total  &
		    24	&58	&26	&40	&120	&30	&428	&138	&242	&90	&38	&80
		     & \\\cline{2-14}
		\multirow{3}*{} &
		accuracy (\%)   &
		\textbf{62.50} & \textbf{51.72} &	53.85 &	65.00 &	\textbf{70.83} &	\textbf{70.00} &	\textbf{81.54} &	\textbf{87.68} &	\textbf{95.87} &	\textbf{54.44} &	71.05 &	\textbf{60.00}
		     & \\\hline			
		\multirow{3}*{\tabincell{c}{LPI~\cite{FanWH12B}}}     &
		\#correct &
		5	&0	&15	&19	&53	&3	&123	&60	&33	&17	&11	&16
				   	
		 & \multirow{3}*{48.83\%}\\ \cline{2-14}
		\multirow{3}*{} &
		\#total  &
		9	&0	&18	&29	&90	&9	&193	&102	&59	&38	&15	&40		
		     & \\\cline{2-14}
		\multirow{3}*{} &
		accuracy (\%)   &
		55.56 &	0.00  &	\textbf{83.33} &	\textbf{65.52} &	58.89 &	33.33 &	63.73 &	58.82 &	55.93 &	44.74 &	\textbf{73.33} &	40.00 			
		 & \\\hline		
		\multirow{3}*{\tabincell{c}{LJL~\cite{LiYLLZ16}}}     &
		\#correct &
		8	& 24	& 26	& 37	& 148	& 4	&221	&113	&129	&50	&26	&22
		 & \multirow{3}*{52.95\%}\\ \cline{2-14}
		\multirow{3}*{} &
		\#total  &
		30	&79	&32	&64	&251	&17	&376	&186	&138	&131	&50	&102		
		 & \\\cline{2-14}
		\multirow{3}*{} &
		accuracy (\%)   &
		26.67 &	30.38 &	81.25 &	57.81 &	58.96 &	23.53 &	58.78 &	60.75 &	93.48 &	38.17 &	52.00 &	21.57 		
 & \\\hline					
	\end{tabular}
	}
	\label{tab: matchlsg_results}
\end{table*}

\begin{figure*}[htb!]
\centering
\includegraphics[width=0.9\linewidth]{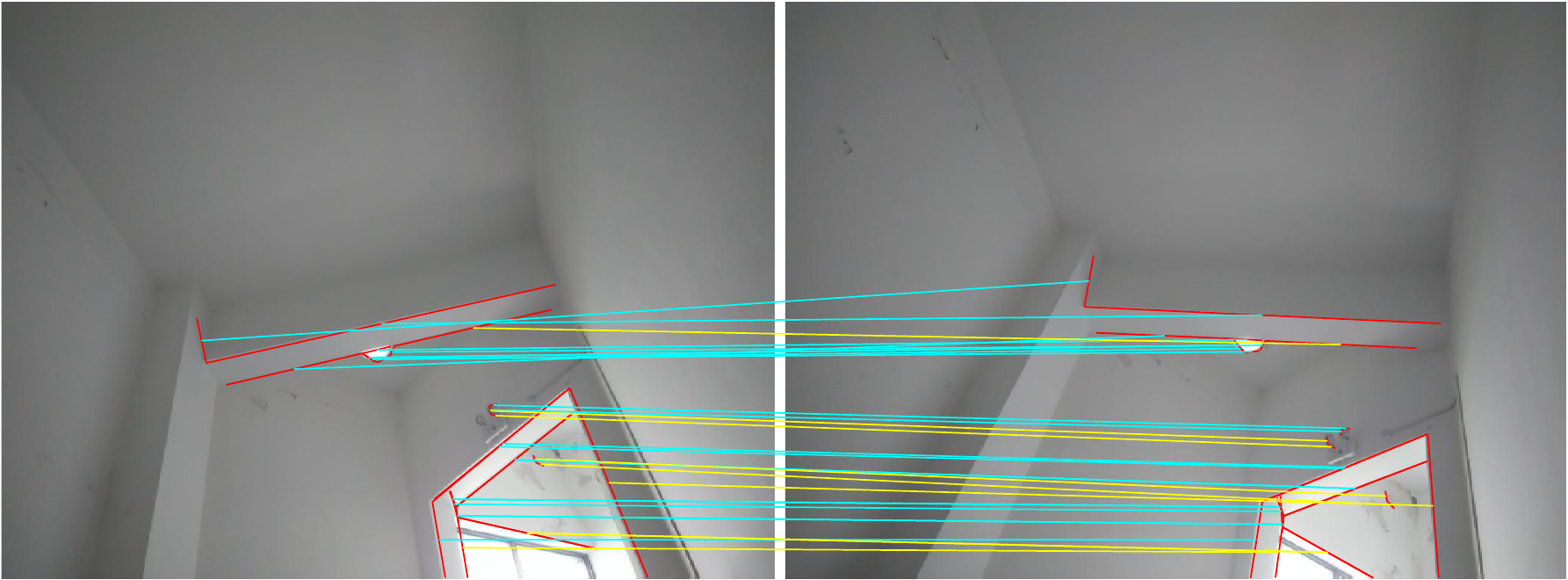}
\caption{Matched line-segments for image pair Fig.~\ref{fig:imageshow} (f). (\#correct matches, \#total matches) = (21, 30). Midpoints of matched line-segments are connect by cyan lines (if they are correct) or yellow lines (mismatches).}
\label{fig:plot-ASJ-LSG-f}
\end{figure*}

\begin{figure*}[htb!]
\centering
\includegraphics[width=0.24\linewidth]{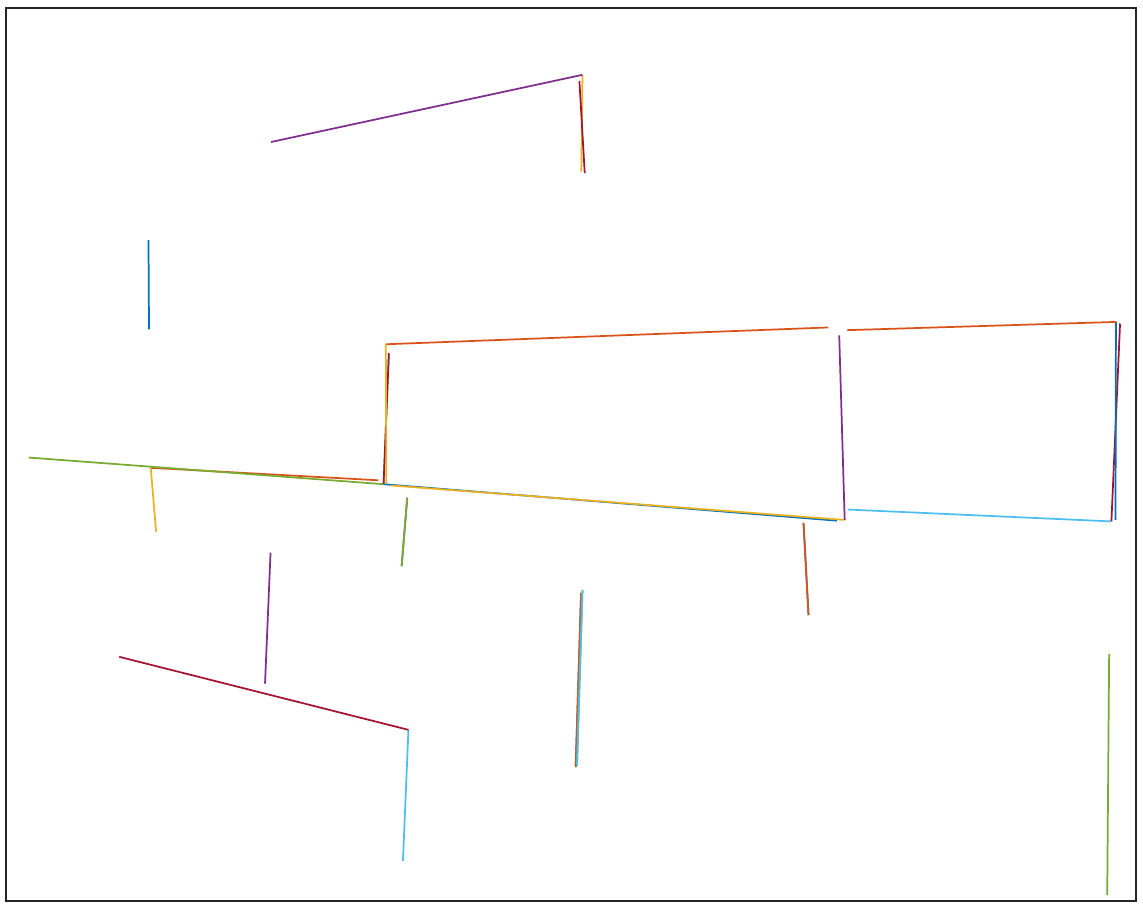}
\includegraphics[width=0.24\linewidth]{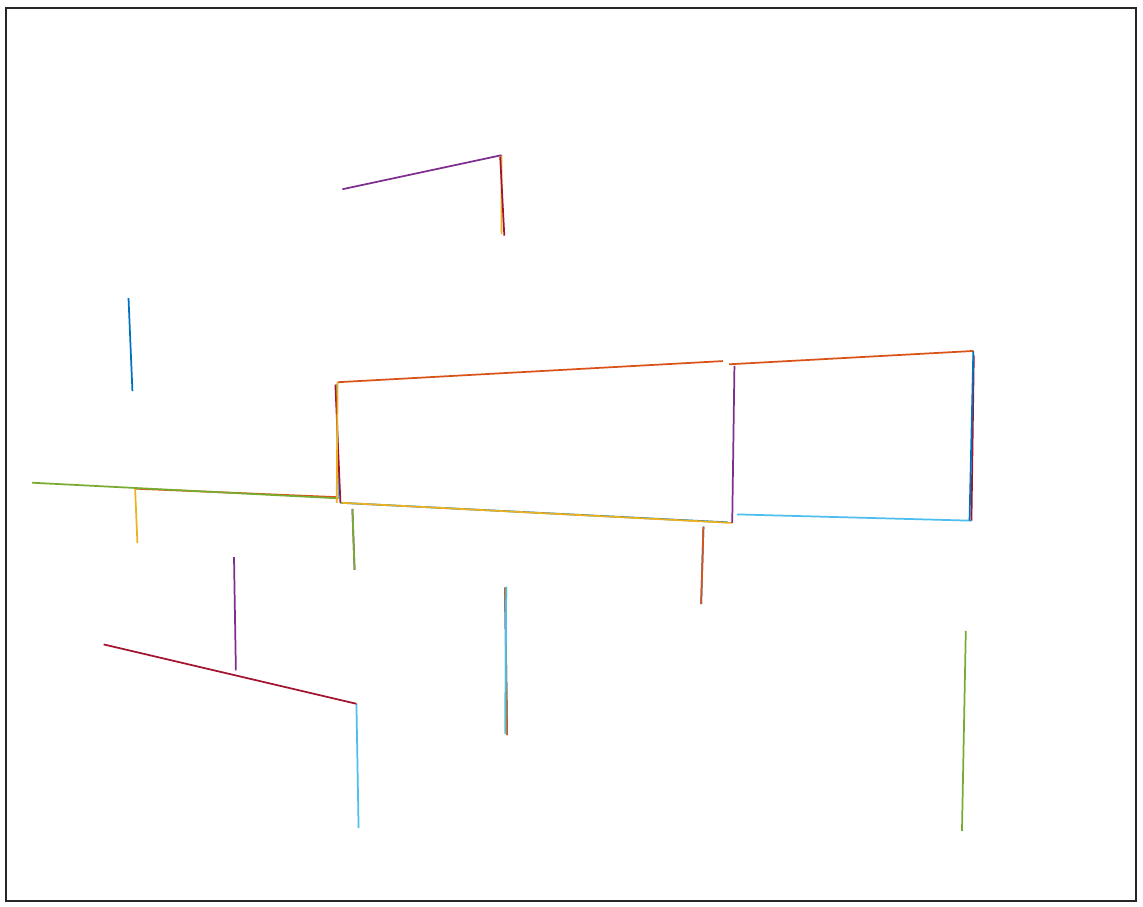}
\includegraphics[width=0.24\linewidth]{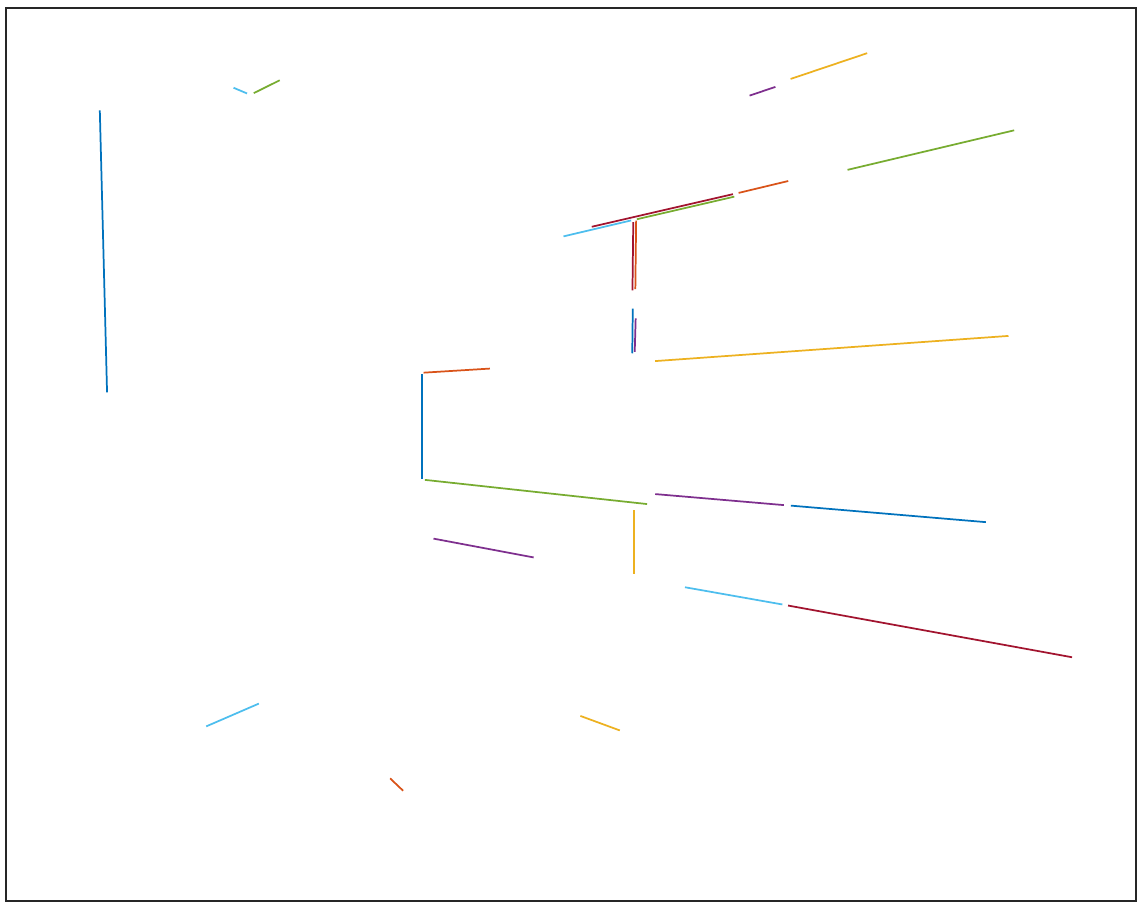}
\includegraphics[width=0.24\linewidth]{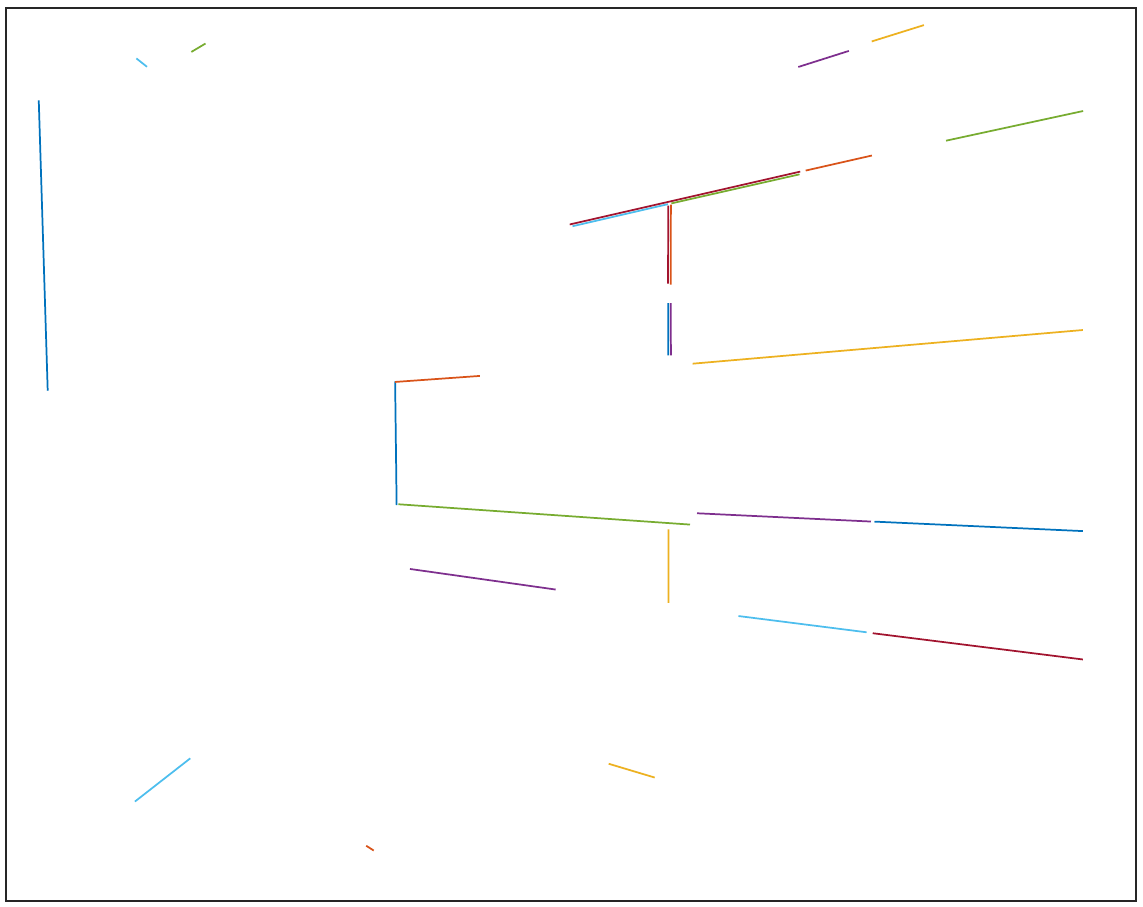}
\caption{Matched line segments for image pair Fig.~\ref{fig:imageshow} (b). Left and mid-left: correct matched line-segments for ASJ; Right and mid-right: correct matched line-segments for LJL~\cite{LiYLLZ16}. The result of ASJ covers the scene more complete benefiting with the anisotropic scales for branches of junctions.}
	\label{fig:matched-lsg-b}
\end{figure*}

We evaluate the matched line-segments with state-of-the-art approaches LPI~\cite{FanWH12B} and LJL~\cite{LiYLLZ16} for the comparison in a more strict rule that compare endpoints of corresponding line-segments instead of their line equation. For the example image pairs shown in Fig.~\ref{fig:imageshow}, our proposed method is better than existing methods in considerable advantage for most cases. Some matched results for line segments are shown in  Fig.~\ref{fig:plot-ASJ-LSG-f} and Fig.~\ref{fig:matched-lsg-b}. The number of correct matched line-segments are also comparable with other approaches. 
Besides of the matching accuracy, the result shown in 
 Fig.~\ref{fig:matched-lsg-b} for our method cover the scene more complete.

Different from the approaches LPI~\cite{FanWH12B} and LJL~\cite{LiYLLZ16}, our approach performs better while not using any pre-estimated geometric information. As shown the Tab.~\ref{tab: matchlsg_results}, we will find that key-point driven approach for line segment matching is possible to be failed because of the erroneous estimated geometric relationship. Observing the failed case reported in Tab.~\ref{tab: matchlsg_results}, the image pair in Fig.~\ref{fig:matched-lsg-b} is dominant by repeated texture and severe viewpoint change which are challenging for key-point matching. In such scenario, the induced epipolar geometry might be unreliable and therefore produce poor line segment matching results. On the other hand, because our approach performs well in junction matching, we can also use the junction correspondences to refine the line segment matching result. 



\section{Conclusion}
\label{sec:con}
In this paper, we proposed a novel junction detector ASJ which exploits the anisotropy of junctions via estimating the endpoints (length) of branches for isotropic scale junctions for indoor images which are dominanted by junctions in a more global manner. We then devised an affine invariant dissimilarity measure to match these anisotropic-scale junctions across different images. We tested our method on a collected indoor images and compared its performance with several current sate-of-the-art methods. The results demonstrated that our approach establishes new state-of-the-art performance on the indoor image dataset. 


\bibliographystyle{IEEEtran}
\bibliography{references}

\begin{thebibliography}{10}
\providecommand{\url}[1]{#1}
\csname url@samestyle\endcsname
\providecommand{\newblock}{\relax}
\providecommand{\bibinfo}[2]{#2}
\providecommand{\BIBentrySTDinterwordspacing}{\spaceskip=0pt\relax}
\providecommand{\BIBentryALTinterwordstretchfactor}{4}
\providecommand{\BIBentryALTinterwordspacing}{\spaceskip=\fontdimen2\font plus
\BIBentryALTinterwordstretchfactor\fontdimen3\font minus
  \fontdimen4\font\relax}
\providecommand{\BIBforeignlanguage}[2]{{%
\expandafter\ifx\csname l@#1\endcsname\relax
\typeout{** WARNING: IEEEtran.bst: No hyphenation pattern has been}%
\typeout{** loaded for the language `#1'. Using the pattern for}%
\typeout{** the default language instead.}%
\else
\language=\csname l@#1\endcsname
\fi
#2}}
\providecommand{\BIBdecl}{\relax}
\BIBdecl

\bibitem{Wu13}
C.~Wu, ``Towards linear-time incremental structure from motion,'' in
  \emph{International Conference on 3D Vision}, 2013, pp. 127--134.

\bibitem{CrandallOSH13}
D.~J. Crandall, A.~Owens, N.~Snavely, and D.~P. Huttenlocher, ``Sfm with mrfs:
  Discrete-continuous optimization for large-scale structure from motion,''
  \emph{IEEE Transactions on Pattern Analysis and Machine Intelligence},
  vol.~35, no.~12, pp. 2841--2853, 2013.

\bibitem{FuhrmannLG14}
S.~Fuhrmann, F.~Langguth, and M.~Goesele, ``{MVE} - {A} multi-view
  reconstruction environment,'' in \emph{Eurographics Workshop on Graphics and
  Cultural Heritage, Darmstadt, Germany}, 2014, pp. 11--18.

\bibitem{openMVG}
P.~Moulon, P.~Monasse, R.~Marlet, and Others, ``Openmvg. an open multiple view
  geometry library.'' \url{https://github.com/openMVG/openMVG}.

\bibitem{WangBWLT10}
B.~Wang, X.~Bai, X.~Wang, W.~Liu, and Z.~Tu, ``Object recognition using
  junctions,'' in \emph{European Conference on Computer Vision}, 2010, pp.
  15--28.

\bibitem{ChiaRLR12}
A.~Y.~S. Chia, D.~Rajan, M.~K. Leung, and S.~Rahardja, ``Object recognition by
  discriminative combinations of line segments, ellipses, and appearance
  features,'' \emph{IEEE Transactions on Pattern Analysis and Machine
  Intelligence}, vol.~34, no.~9, pp. 1758--1772, 2012.

\bibitem{YanWZYC15}
J.~Yan, J.~Wang, H.~Zha, X.~Yang, and S.~M. Chu, ``Multi-view point
  registration via alternating optimization,'' in \emph{AAAI Conference on
  Artificial Intelligence}, 2015, pp. 3834--3840.

\bibitem{ShenLYXWW15}
Y.~Shen, W.~Lin, J.~Yan, M.~Xu, J.~Wu, and J.~Wang, ``Person re-identification
  with correspondence structure learning,'' in \emph{IEEE International
  Conference on Computer Vision}, 2015, pp. 3200--3208.

\bibitem{MikolajczykS02}
K.~Mikolajczyk and C.~Schmid, ``An affine invariant interest point detector,''
  in \emph{European Conference on Computer Vision}, 2002, pp. 128--142.

\bibitem{Lowe04}
D.~G. Lowe, ``Distinctive image features from scale-invariant keypoints,''
  \emph{International Journal of Computer Vision}, vol.~60, no.~2, pp. 91--110,
  2004.

\bibitem{MatasCUP02}
J.~Matas, O.~Chum, M.~Urban, and T.~Pajdla, ``Robust wide baseline stereo from
  maximally stable extremal regions,'' in \emph{British Machine Vision
  Conference}, 2002, pp. 1--10.

\bibitem{TuytelaarsG04}
T.~Tuytelaars and L.~J.~V. Gool, ``Matching widely separated views based on
  affine invariant regions,'' \emph{International Journal of Computer Vision},
  vol.~59, no.~1, pp. 61--85, 2004.

\bibitem{YuM11}
G.~Yu and J.~Morel, ``{ASIFT:} an algorithm for fully affine invariant
  comparison,'' \emph{{IPOL} Journal}, vol.~1, pp. 11--38, 2011.

\bibitem{FanWH12B}
B.~Fan, F.~Wu, and Z.~Hu, ``Robust line matching through line-point
  invariants,'' \emph{Pattern Recognition}, vol.~45, no.~2, pp. 794--805, 2012.

\bibitem{LiYLLZ16}
K.~Li, J.~Yao, X.~Lu, L.~Li, and Z.~Zhang, ``Hierarchical line matching based
  on line-junction-line structure descriptor and local homography estimation,''
  \emph{Neurocomputing}, vol. 184, pp. 207--220, 2016.

\bibitem{GioiJMR10}
R.~G. von Gioi, J.~Jakubowicz, J.~Morel, and G.~Randall, ``{LSD:} {A} fast line
  segment detector with a false detection control,'' \emph{IEEE Transactions on
  Pattern Analysis and Machine Intelligence}, vol.~32, no.~4, pp. 722--732,
  2010.

\bibitem{GioiJMR12}
------, ``{LSD:} a line segment detector,'' \emph{{IPOL} Journal}, vol.~2, pp.
  35--55, 2012.

\bibitem{ShenP00}
X.~Shen and P.~Palmer, ``Uncertainty propagation and the matching of junctions
  as feature groupings,'' \emph{IEEE Transactions on Pattern Analysis and
  Machine Intelligence}, vol.~22, no.~12, pp. 1381--1395, 2000.

\bibitem{WangWH09}
Z.~Wang, F.~Wu, and Z.~Hu, ``{MSLD:} {A} robust descriptor for line matching,''
  \emph{Pattern Recognition}, vol.~42, no.~5, pp. 941--953, 2009.

\bibitem{Marr82}
D.~Marr, ``A computational investigation into the human representation and
  processing of visual information,'' \emph{Vision}, pp. 125--126, 1982.

\bibitem{Adelson00}
E.~H. Adelson, ``Lightness perception and lightness illusions,'' \emph{New
  Cogn. Neurosci}, vol. 339, 2000.

\bibitem{GuoZW07}
C.~Guo, S.~C. Zhu, and Y.~N. Wu, ``Primal sketch: Integrating structure and
  texture,'' \emph{Computer Vision and Image Understanding}, vol. 106, no.~1,
  pp. 5--19, 2007.

\bibitem{WuXZ07}
T.~Wu, G.~Xia, and S.~C. Zhu, ``Compositional boosting for computing
  hierarchical image structures,'' in \emph{CVPR}, 18-23 June 2007.

\bibitem{MaireAFM08}
M.~Maire, P.~Arbelaez, C.~C. Fowlkes, and J.~Malik, ``Using contours to detect
  and localize junctions in natural images,'' in \emph{CVPR}, June 24-26 2008.

\bibitem{Sinzinger08}
E.~D. Sinzinger, ``A model-based approach to junction detection using radial
  energy,'' \emph{Pattern Recognition}, vol.~41, no.~2, pp. 494--505, 2008.

\bibitem{PuspokiU15}
Z.~P{\"{u}}sp{\"{o}}ki and M.~Unser, ``Template-free wavelet-based detection of
  local symmetries,'' \emph{IEEE Transactions on Image Processing}, vol.~24,
  no.~10, pp. 3009--3018, 2015.

\bibitem{PuspokiUVU16}
Z.~P{\"{u}}sp{\"{o}}ki, V.~Uhlmann, C.~Vonesch, and M.~Unser, ``Design of
  steerable wavelets to detect multifold junctions,'' \emph{IEEE Transactions
  on Image Processing}, vol.~25, no.~2, pp. 643--657, 2016.

\bibitem{XiaDG14}
G.~Xia, J.~Delon, and Y.~Gousseau, ``Accurate junction detection and
  characterization in natural images,'' \emph{International Journal of Computer
  Vision}, vol. 106, no.~1, pp. 31--56, 2014.

\bibitem{SrajerSPP14}
F.~Srajer, A.~G. Schwing, M.~Pollefeys, and T.~Pajdla, ``Match box: Indoor
  image matching via box-like scene estimation,'' in \emph{International
  Conference on 3D Vision}, 2014, pp. 705--712.

\bibitem{FurukawaCSS09}
Y.~Furukawa, B.~Curless, S.~M. Seitz, and R.~Szeliski, ``Reconstructing
  building interiors from images,'' in \emph{IEEE Conference on Computer Vision
  and Pattern Recognition}, September 27 - October 4 2009, pp. 80--87.

\bibitem{Forstner86}
W.~F{\"o}rstner, ``A feature based correspondence algorithm for image
  matching,'' \emph{International Archives of Photogrammetry and Remote
  Sensing}, vol.~26, no.~3, pp. 150--166, 1986.

\bibitem{HarrisS88}
C.~Harris and M.~Stephens, ``A combined corner and edge detector,'' in
  \emph{Alvey Vision Conference}, 1988, pp. 147--151.

\bibitem{MikolajczykS04}
K.~Mikolajczyk and C.~Schmid, ``Scale {\&} affine invariant interest point
  detectors,'' \emph{International Journal of Computer Vision}, vol.~60, no.~1,
  pp. 63--86, 2004.

\bibitem{ForstnerDS09}
W.~F{\"{o}}rstner, T.~Dickscheid, and F.~Schindler, ``Detecting interpretable
  and accurate scale-invariant keypoints,'' in \emph{IEEE Conference on
  Computer Vision and Pattern Recognition}, 2009, pp. 2256--2263.

\bibitem{AlvarezM97}
L.~Alvarez and F.~Morales, ``Affine morphological multiscale analysis of
  corners and multiple junctions,'' \emph{International Journal of Computer
  Vision}, vol.~25, no.~2, pp. 95--107, 1997.

\bibitem{VincentL04}
{\'{E}}.~Vincent and R.~Lagani{\`{e}}re, ``Junction matching and fundamental
  matrix recovery in widely separated views,'' in \emph{British Machine Vision
  Conference}, 2004, pp. 1--10.

\bibitem{Ruff87}
B.~P.~D. Ruff, ``A pipelined architecture for the canny edge detector,'' in
  \emph{Alvey Vision Conference, Cambridge, UK}, 1987, pp. 1--4.

\bibitem{RamalingamASLP15}
S.~Ramalingam, M.~Antunes, D.~Snow, G.~H. Lee, and S.~Pillai, ``Line-sweep:
  Cross-ratio for wide-baseline matching and 3d reconstruction,'' in \emph{IEEE
  Conference on Computer Vision and Pattern Recognition}, 2015, pp. 1238--1246.

\bibitem{FischlerB81}
M.~A. Fischler and R.~C. Bolles, ``Random sample consensus: {A} paradigm for
  model fitting with applications to image analysis and automated
  cartography,'' \emph{Communications of the ACM}, vol.~24, no.~6, pp.
  381--395, 1981.

\bibitem{DesolneuxMM07}
A.~Desolneux, L.~Moisan, and J.-M. Morel, \emph{From Gestalt Theory to Image
  Analysis: A Probabilistic Approach}, 1st~ed.\hskip 1em plus 0.5em minus
  0.4em\relax Springer Publishing Company, Incorporated, 2007.

\bibitem{NeubeckG06}
A.~Neubeck and L.~J.~V. Gool, ``Efficient non-maximum suppression,'' in
  \emph{IEEE International Conference on Pattern Recognition}, 2006, pp.
  850--855.

\bibitem{MikolajczykS05}
K.~Mikolajczyk and C.~Schmid, ``A performance evaluation of local
  descriptors,'' \emph{IEEE Transactions on Pattern Analysis and Machine
  Intelligence}, vol.~27, no.~10, pp. 1615--1630, 2005.

\bibitem{MorelY09}
J.~Morel and G.~Yu, ``{ASIFT:} {A} new framework for fully affine invariant
  image comparison,'' \emph{{SIAM} J. Imaging Sciences}, vol.~2, no.~2, pp.
  438--469, 2009.

\bibitem{PerdochCM09}
M.~Perdoch, O.~Chum, and J.~Matas, ``Efficient representation of local geometry
  for large scale object retrieval,'' in \emph{IEEE Conference on Computer
  Vision and Pattern Recognition}, 2009, pp. 9--16.

\end{thebibliography}

\end{document}